\documentclass{article}

\PassOptionsToPackage{numbers, compress}{natbib}
\usepackage[preprint]{neurips_2025}

\usepackage[utf8]{inputenc} % allow utf-8 input
\usepackage[T1]{fontenc}    % use 8-bit T1 fonts
\usepackage{hyperref}       % hyperlinks
\usepackage{url}            % simple URL typesetting
\usepackage{booktabs}       % professional-quality tables
\usepackage{amsfonts}       % blackboard math symbols
\usepackage{nicefrac}       % compact symbols for 1/2, etc.
\usepackage{microtype}      % microtypography
\usepackage{xcolor}         % colors
\usepackage{subfigure}
\usepackage{wrapfig}

\usepackage{bookmark} 
\usepackage{algorithm}
\usepackage{algorithmic}
\usepackage{amsmath}
\usepackage{amssymb}
\usepackage{mathtools}
\usepackage{amsthm}
\usepackage{enumitem} 
\AtBeginDocument{
  \def\NAT@force@numbers{}%
}
\theoremstyle{plain}
\newtheorem{theorem}{Theorem}[section]

\newtheorem{lemma}[theorem]{Lemma}

\theoremstyle{definition}
\newtheorem{definition}[theorem]{Definition}

\theoremstyle{remark}

\usepackage{listings} % 引入 listings 包

\title{Risk-aware Direct Preference Optimization \\ 
under Nested Risk Measure}

\author{
    Lijun Zhang\textsuperscript{1}, Lin Li\textsuperscript{1}, Yajie Qi\textsuperscript{1}, Huizhong Song\textsuperscript{1}, 
    Yaodong Yang\textsuperscript{2}, Jun Wang\textsuperscript{3}, Wei Wei\textsuperscript{1}
    \thanks{1. Key Laboratory of Computational Intelligence and Chinese Information Processing of Ministry of Education, School of Computer and Information Technology, Shanxi University, Taiyuan, Shanxi, China. 
   2. Institute for AI, Peking University, Beijing, China.  
   3. University College London, London, UK. 
   Correspondence to <\texttt{weiwei@sxu.edu.cn}>.}
}

\begin{document}

\maketitle

\begin{abstract}
    When fine-tuning pre-trained Large Language Models (LLMs) to align with human values and intentions, maximizing the estimated reward can lead to superior performance, but it also introduces potential risks due to deviations from the reference model's intended behavior.
    Most existing methods typically introduce KL divergence to constrain deviations between the trained model and the reference model; however, this may not be sufficient in certain applications that require tight risk control.
    In this paper, we introduce Risk-aware Direct Preference Optimization (Ra-DPO), a novel approach that incorporates risk-awareness by employing a class of nested risk measures.
    This approach formulates a constrained risk-aware advantage function maximization problem and then converts the Bradley-Terry model into a token-level representation.
    The objective function maximizes the likelihood of the policy while suppressing the deviation between a trained model and the reference model using a sequential risk ratio, thereby enhancing the model's risk-awareness.
    Experimental results across three open-source datasets: IMDb Dataset, Anthropic HH Dataset, and AlpacaEval, demonstrate the proposed method's superior performance in balancing alignment performance and model drift.
    Our code is opensourced at {\small \url{https://github.com/zlj123-max/Ra-DPO}}.
\end{abstract}

%%%%%%%%%%%%%%%%%%%%%%%%%%%%%%%%%%%%%%%%%%%%%%
%% Section 1 Introduction
%%%%%%%%%%%%%%%%%%%%%%%%%%%%%%%%%%%%%%%%%%%%%%
\section{Introduction} \label{Section: Introduction}
Learning from human feedback, serving as a bridge to align LLMs with human preferences, is crucial for ensuring that the generations are more helpful, factual, and ethical, among other desiderata~\cite{2017_christiano_rlhf, 2022_ouyang_gpt-3, 2023_touvron_llama, 2024_song_pro}.
Alignment methods such as RLHF \cite{2022_ouyang_gpt-3, 2023_touvron_llama} and DPO \cite{2023_rafailov_dpo} have consistently proven more effective than supervised finetuning (SFT) alone.
Notably, DPO, featuring a simple and straightforward training process, directly uses the likelihood of the policy to define an implicit reward fitted to the preference data, which has emerged as a popular alternative since it bypasses explicit reward modeling challenges while delivering competitive performance.
Subsequently, a variety of DPO variants have been proposed, such as f-DPO \cite{2024_wang_fdpo}, IPO \cite{2024_azar_ipo}, RDPO \cite{2024_fisch_rdpo}, and SimPO \cite{2024_meng_simpo}, to enhance performance. 
However, a key limitation of these methods is that they only consider evaluation at the sentence level, ignoring the fact that the generation of these responses occurs sequentially, following an auto-regressive approach.

Recently, a fresh perspective on LLMs alignment has been introduced, specifically a sequential and token-level direct preference optimization known as TDPO \cite{2024_zeng_tdpo}.
This method allows for examining divergence in relation to a reference model on a more granular, token-by-token basis.
Specifically, inspired by Trust Region Policy Optimization (TRPO) \cite{2015_schulman_trpo} from reinforcement learning (RL) field \cite{2017_arulkumaran_rl_survey, 2018_franccois_rl_survey}, TDPO redefines the objective of maximizing restricted rewards in a sequential manner and bridges sentence-level reward to token-level generation through the Bellman equation.
However, since the objective at each step is to maximize the expected return, a risk-neutral criterion, which neglects the characteristics of the reward distribution beyond the mean, TDPO cannot guarantee a low risk of deviation from the reference model during alignment training.
This could be catastrophic for practical applications, as a significant deviation from the reference model typically implies the degradation of superior decision-making and reasoning capabilities.

Fortunately, in the field of RL, risk-sensitive methods \cite{2022_jaimungal_rrrl, 2022_bisi_rosa, 2023_candela_risk-aware-controller} have been proposed that achieve superior performance by introducing various risk measures.
Recently, some researchers introduce this technology to align LLMs with human preferences.
For instance, RA-RLHF \cite{2024_chaudhary_RA-RLHF} introduces a static risk measure into the fine-tuning of RL, while KTO~\cite{2024_ethayarajh_kto} introduces prospect theory \cite{1992_tversky_prospect-theory} to fit human choice behavior when faced with uncertain events.
However, these methods only consider the risk at the sentence level by analyzing the distribution characteristics of the preference data, thereby overlooking the inherently sequential and auto-regressive process of response generation.

In this paper, we focus on the risk in token-level generation when aligning LLMs with human values and intentions.
Specifically, from a risk-sensitive perspective, we investigate a novel direct preference optimization method and provide corresponding theoretical and empirical results.
Our main contributions are summarized as follows.
\begin{itemize}
    \item	We design a new risk-aware, token-level objective function and prove that maximizing this objective leads to policy improvements.
          Furthermore, by deriving the mapping from the risk-aware state-action value function to the optimal policy and establishing the equivalence between the Bradley-Terry model and the Regret Preference Model, we obtain an optimization objective that is solely dependent on the risk-sensitive policy.
    \item	We propose a novel Risk-aware Direct Preference Optimization (Ra-DPO) method.
          The method maintains a natural and simple loss function, specifically, the sum of the DPO loss and the negative sequential risk ratio (see Figure \ref{Fig.1}).
          This loss function maximizes policy likelihood while suppressing deviation from reference model through the sequential risk ratio, thereby enhancing risk-awareness in striking a balance between alignment performance and model drift.
    \item	Experimentally, we evaluate the effectiveness of our proposed method across various text generation tasks and assess its sensitivity to the risk control parameter.
          The experimental results demonstrate that our method can effectively suppress the risk of model drift while enhancing its performance.
\end{itemize}

%%%%%%%%%%%%%%%%%%%%%%%%%%%%%%%%%%%%%%%%%%%%%%
%% Section 2 Preliminaries
%%%%%%%%%%%%%%%%%%%%%%%%%%%%%%%%%%%%%%%%%%%%%%
\section{Preliminaries} \label{Section: Preliminaries}
\subsection{Preference-based Policy Optimization}
Considering a preference-based language model fine-tuning task, let $x$ denote an input prompt (question), and $y $ denote the generated response (answer).
The notation $y_w \succ y_l \mid x$ symbolizes the human preference data, where $y_w$ (win) represents a response that is more preferred by humans compared to $y_l$ (lose).
Both $x$ and $y_w/y_l$ are sequences of tokens.

\textbf{Bradley-Terry Model. }
In preference-based fine-tuning, to align with human preferences, a preference predictor adhering to the Bradley-Terry (BT) \cite{1952_bradley_BT-model} model has been widely employed for pairwise comparisons.
The likelihood of a preference pair is commonly expressed as:
\begin{equation}  \label{Equation: BT model}
    P_{\mathrm{BT}}\left(y_w \succ y_l \mid x\right) = \frac{\exp\left(r^{\ast}\left(x, y_w\right)\right)}{\exp\left(r^{\ast}\left(x, y_w\right)\right) + \exp \left(r^{\ast}\left(x, y_l\right)\right)},
\end{equation}
where $r^{\ast}(x, y_w)$ and $r^{\ast}(x, y_l)$ stand for the reward function at the sentence level from the preferred and dispreferred answers, respectively.

\textbf{Direct Preference Optimization. }
DPO \cite{2023_rafailov_dpo} begins with the following RL objective:
\begin{equation} \label{Equation: RL objective}
        \max_{\pi_{\theta}} \mathbb{E}_{x \sim \mathcal{D}} \left[\mathbb{E}_{y \sim \pi_{\theta}(\cdot \mid x)}\left[r \left(x, y\right) -\beta D_{\mathrm{KL}}\left(\pi_{\theta}(\cdot \mid x) \| \pi_{\mathrm{ref}}(\cdot \mid x)\right)\right]\right],
\end{equation}
where $\mathcal{D}$ represents the human preference dataset, $\beta$ is the coefficient of the reverse KL divergence penalty, $\pi_{\mathrm{ref}}\left(\cdot \mid x\right)$ is the policy of a fixed reference model (typically selected to be the model that has undergone post-supervised fine-tuning), and $\pi_{\theta}\left(\cdot \mid x\right)$ represents the policy of the trained model, initialized with $\pi_{\theta}=\pi_{\mathrm{ref}}$.

By reparameterizing the reward function in Equation~(\ref{Equation: RL objective}), DPO establishes a direct functional mapping between the reward model and the optimal policy:
\begin{equation}    \label{Equation: the reward function reparameterized by using the policy}
    r(x, y)=\beta \log \frac{\pi^{\ast}_{\theta}(y \mid x)}{\pi_{\mathrm{ref}}(y \mid x)}+\beta \log Z(x),
\end{equation}
where $Z(x)$ is the partition function.
Subsequently, Equation~(\ref{Equation: RL objective}) can be reformulated as DPO loss:
\begin{equation}
    \mathcal{L}_{\mathrm{DPO}}\left(\pi_{\theta}; \pi_{\mathrm{ref}}\right)=-\mathbb{E}_{\left(x, y_w, y_l\right) \sim \mathcal{D}}\left[\log\sigma\left(u\left(x, y_w, y_l\right)\right)\right],
\end{equation}
where $u\left(x, y_w, y_l\right) = \beta\log\frac{\pi_{\theta}\left(y_w \mid x\right)}{\pi_{\mathrm{ref}}\left(y_w \mid x\right)} - \beta\log\frac{\pi_{\theta}\left(y_l \mid x\right)}{\pi_{\mathrm{ref}}\left(y_l \mid x\right)}.$

\subsection{Preference-based Markov Decision Process}     \label{Subsection: Preference-based MDP}
A Preference-based Markov Decision Process (Pb-MDP) can be formulated as a modification of the classical MDP: $\mathcal{M} = \langle \mathcal{S}, \mathcal{A}, r, \mathbf{P}, \gamma, T \rangle$,
where $\mathcal{S}$ and $\mathcal{A}$ represent the finite state and action spaces, respectively;
$\mathbf{P}: \mathcal{S} \times \mathcal{A} \rightarrow \mathcal{S}$ is the probabilistic transition function;
$r$ represents the reward function over the entire prompt-response, which is defined as $\left(\mathcal{S}\times \mathcal{A}\right)^{T} \rightarrow \mathcal{R}$;
$\gamma$ is the discount factor,
and $T$ denotes the length of a trajectory or episode.

Specifically, for language generation, the state $s_t = \left [x, y^{<t} \right] \in \mathcal{S}$ usually consists of the prompt and the generated response up to the current step, and action $a_t = y^t \in \mathcal{A}$ corresponds to the next generated token.
Additionally, note that $y^{<1}= [\; ]$ is an empty sequence.
Therefore, we denote $\left[x\right]=\left[x, [\; ]\right]=\left[x, y^{<1}\right]$.
For a given prompt $x$ and the first $t-1$ tokens $y^{<t}$ of the response $y$, the probability distribution of the next token conditioned on $[x, y^{<t}]$ is denoted by $\pi_{\theta}(\cdot \mid [x, y^{<t}])$.

\subsection{Risk Measure}  \label{Subsection: Risk Measure}
It is more desirable to keep risk under control for language generation tasks rather than relying solely on a risk-neutral criterion, which ignores the distributional characteristics of rewards, especially in applications that may have potential broad societal impact.
Therefore, we introduce the risk-sensitive criterion \cite{2014_bauerle_more-RsMDP, 2022_wang_risk-averse-autonomous-systems} to quantify potential hidden risks.
More specifically, we provide an introduction to the risk-sensitive function and nested risk measure as follows.

\textbf{Risk-sensitive Function. }
In this paper, the risk-sensitive function is required to satisfy the following properties for all $Z, Z'\in \mathcal{Z}$:
\emph{Concavity: } $\forall \; \lambda \in \left[0, 1 \right]: \eta\left(\lambda Z + \left(1-\lambda \right) Z^{\prime}\right) \geq \lambda\eta\left(Z \right) + \left(1 - \lambda\right) \eta\left(Z^{\prime}\right)$;
% \emph{Monotonicity: } If $Z \geq Z^{\prime}$, then $\eta(Z) \geq \eta\left(Z^{\prime}\right)$;
\emph{Translation Invariance: } $\forall \; \epsilon \in \mathbb{R}: \eta\left(Z + \epsilon \right) = \eta\left(Z \right) + \epsilon$.
This class captures a broad range of useful objectives, including the popular Conditional Value-at-Risk (CVaR)~\cite{1997_artzner_CVaR, 2000_rockafellar_CVaR, 2015_chow_cvar} and Entropic Risk Measure (ERM)~\cite{2002_follmer_erm, 2023_hau_ERM}.

\textbf{Nested Risk-measures. }
In the context of standard Pb-MDP, nested risk measures \cite{2020_fei_rsvi, 2024_chen_Iter-CVaR, 2024_zhao_ra-pbrl} can be expressed in Bellman equation type as follows:
\begin{equation} \label{Equation: Nested PbRL MDP}
    \begin{cases}
        Q_{\pi}\left(\left[x, y^{<t}\right], y^t\right) = R\left(\left[x, y^{<t}\right], y^t\right) + \operatorname{\Phi}^{\mu}\left(V_{\pi}\left(\left[x, y^{<t+1}\right]\right)\right), \\
        V_{\pi}\left(\left[x, y^{<t}\right]\right) = \mathbb{E}_{\pi} \left[Q_{\pi}\left(\left[x, y^{<t}\right], y^{t}\right)\right],                                     \\
        V_{\pi}\left(\left[x, y^{<T}\right]\right) = R\left(\left[x, y^{<T}\right]\right),
    \end{cases}
\end{equation}
where $\operatorname{\Phi}(\cdot)$ is a nested risk measure function with a risk control parameter $\mu$, $Q_{\pi}\left(\left[x, y^{<t}\right], y^t\right)$ and $V_{\pi}\left(\left[x, y^{<t}\right]\right)$ represent the state-action value and state value under the nested risk measures at timestep $t \in [1, \cdots, T]$, respectively.

Due to space constraints, we provide a detailed survey on risk measures in Appendix \ref{Appendix: a survey of Risk Measure} and the expanded version of the value function definition in Appendix \ref{Appendix: the expanded version of value function definition}.

%%%%%%%%%%%%%%%%%%%%%%%%%%%%%%%%%%%%%%%%%%%%%%     
%% Section 3 Methodology
%%%%%%%%%%%%%%%%%%%%%%%%%%%%%%%%%%%%%%%%%%%%%%
\section{Methodology}   \label{Section: Methodology}
This section proposes a novel language model alignment method named Risk-aware Direct Preference Optimization (Ra-DPO).
Specifically, we first analyze the characteristics of nested risk measures and design a new risk-aware token-level objective function by reformulating the constrained reward maximization problem into a token-level form.
Subsequently, we prove that maximizing the objective function leads to policy improvements.
Then, an optimization objective solely related to the risk-sensitive policy is obtained by deriving the mapping from the risk-aware state-action function to the optimal policy and establishing BT model equivalence with the Regret Preference Model.
Finally, we conduct a formal analysis of this optimization objective in terms of derivatives and derive the loss function for Ra-DPO.

\subsection{Risk-aware Objective Function}       \label{Subsection: Risk-aware Objective Function}
In this subsection, we aim to design a new risk-aware objective function for preference-based language model fine-tuning.
Unfortunately, although the recursive Bellman equation under nested risk measures was introduced in Subsection \ref{Subsection: Risk Measure}, it cannot be directly applied due to the following reasons:
(1) For the Pb-MDP setting, the algorithm can only obtain the reward (an implicit reward fitted to the preference data) over the entire prompt-response and thus cannot compute the target value at each step.
(2) The nested risk-measures incorporate a Bellman-type recursion and are not law-invariant~\cite{2023_hau_entropic-risk}, making them complex and difficult to compute.

To surmount these obstacles, a straightforward approach is to introduce the state augmentation method, that is, to reconstruct an augmented Pb-MDP \cite{2024_zhao_ra-pbrl}, where the state at each timestep includes a prompt $x$ and the first $t-1$ tokens $y^{<t}$ of the response.
This approach has the property that the state at the previous timestep is a subset of the state at the current timestep, i.e., $\left[x, y^{<t-1}\right] \subset \left[x, y^{<t}\right]$.
This approach can reformulate the recursive Bellman equation into a classical Bellman equation while satisfying the standard requirements for transformer-based long-sequence modeling in LLMs.
Therefore, in this paper, we directly define the state as a combination of the prompt and the generated response up to the current step to model the sequential and auto-regressive generation.
Then, the nested risk-aware objective's Bellman equation in Equation (\ref{Equation: Nested PbRL MDP}) can be rewritten as:
\begin{equation} \label{Equation: New PbRL MDP}
    \begin{cases}
        \tilde{Q}_{\pi}\left(\left[x, y^{<t}\right], y^t\right) = \operatorname{\Phi}^{\mu}\left(\tilde{V}_{\pi}\left(y^{t+1} \circ \left(\left[x, y^{<t}\right], y^t\right)\right)\right), \\
        \tilde{V}_{\pi}\left(\left[x, y^{<t}\right]\right) = \mathbb{E}_{\pi} \left[\tilde{Q}_{\pi}\left(\left[x, y^{<t}\right], y^{t}\right)\right],                        \\
        \tilde{V}_{\pi}\left(\left[x, y^{<T}\right]\right) = R\left(\left[x, y^{<T}\right]\right),
    \end{cases}
\end{equation}
where $\tilde{Q}_\pi\left(\left[x, y^{<t}\right], y^t\right)$ and $\tilde{V}_{\pi}\left(\left[x, y^{<t}\right]\right)$ represent the risk-aware state-action value and state value under the policy $\pi$, respectively.
The operator $\circ$ denotes the concatenation of the state and action at timestep $t+1$.

It is noteworthy that there is a significant difference in the calculation of $\tilde{V}_{\pi}\left(\left[x, y^{<t}\right]\right)$ and $V_{\pi}\left(\left[x, y^{<t}\right]\right)$.
According to Lemma~3.6 in \cite{2024_zhao_ra-pbrl}, we can obtain the following lemma, whose proof is provided in Appendix \ref{Appendix: the proof of Equivalent MDP lemma}.
\begin{lemma}    \label{Lemma: Equivalent MDP}
    For a given Pb-MDP, the reward over the entire prompt-response can be decomposed as $r = \sum_{t=1}^T \gamma^{t-1} R\left(\left[x, y^{<t}\right], y^t\right)$, the relationship between the state value function Equation (\ref{Equation: Nested PbRL MDP}) and Equation (\ref{Equation: New PbRL MDP}) is as follows:
    $\tilde{V}_{\pi}\left(\left[x, y^{<t}\right]\right) = V_{\pi}\left(\left[x, y^{<t}\right]\right) + R_{1: t-1},$
    where $R_{1: t-1} = \sum_{h=1}^{t-1} \gamma^{h-1} R\left(\left[x, y^{<h}\right], y^h\right)$ denotes the cumulative reward of the $1 \sim t-1$ steps of the prompt-response, and $V_{\pi}[x]$ and $\tilde{V}_{\pi}[x]$ are equivalent.
\end{lemma}

Subsequently, based on Equation~(\ref{Equation: New PbRL MDP}), we define the risk-aware advantage function as follows.
\begin{definition}     \label{Definition: RA-advantage}
    For a risk-sensitive Pb-MDP that satisfies the Bellman equation in Equation~(\ref{Equation: New PbRL MDP}), the risk-aware advantage function can be defined as:
    \begin{equation}    \label{Equation: RA-advantage}
        \tilde{A}_{\pi} \left(\left[x, y^{<t}\right], z\right) = \tilde{Q}_\pi\left(\left[x, y^{<t}\right], z\right) - \operatorname{\Phi}^{\mu}(\tilde{V}_\pi\left(\left[x, y^{<t}\right]\right)),
    \end{equation}
    where $z \sim \pi_{\theta}\left(\cdot \mid \left[x, y^{<t}\right]\right)$.
\end{definition}
The definition is reasonable: its derivation is provided in Appendix~\ref{Appendix: the derivation of risk-aware advantage function definition}.
Furthermore, based on the definition of risk-aware advantage function in Definition~\ref{Definition: RA-advantage}, we propose a new risk-aware objective function:
\begin{equation}  \label{Equation: Ra-RL objective function}
    \max _{\pi_{\theta}} \mathbb{E}_{x, y^{<t} \sim \mathcal{D}, z \sim \pi_{\theta}\left(\cdot \mid \left[x, y^{<t}\right]\right)} \left[\tilde{A}_{\pi_{\mathrm{ref}}}\left(\left[x, y^{<t}\right], z\right) -\beta D_{\mathrm{KL}}\left(\pi_{\theta}\left(\cdot \mid \left[x, y^{<t}\right]\right) \| \pi_{\mathrm{ref}}\left(\cdot \mid \left[x, y^{<t}\right]\right)\right)\right].
\end{equation}
The objective function maximizes a risk-sensitive advantage function subject to a KL divergence constraint, which accounts for risk during selecting the policy, thereby striking a better balance between alignment performance and model drift.
It is worth emphasizing that maximizing the risk-aware objective function in Equation~(\ref{Equation: Ra-RL objective function}) leads to policy improvements, as stated in the following lemma, whose proof is provided in Appendix \ref{Appendix: the proof of the policy improvement lemma}.
\begin{lemma}  \label{Lemma: policy improvement}
    Given two policies $\pi$ and $\pi^{\prime}$, if for any state $s_t = \left[x, y^{<t}\right], \mathbb{E}_{z \sim \pi^{\prime}}\left[\tilde{A}_\pi\left(\left[x, y^{<t}\right], z\right)\right] \geq 0$, then we can conclude:
    $\mathbb{E}_{x \sim \mathcal{D}}\left[\tilde{V}_{\pi^{\prime}}([x])\right] \geq \mathbb{E}_{x \sim \mathcal{D}}\left[\tilde{V}_\pi([x])\right].$
\end{lemma}

\subsection{Risk-aware Preference Optimization}     \label{Subsection: Risk-aware Token-Level Preference Optimization}
In this subsection, we focus on how to convert the BT model into risk-sensitive token-level representation, which is divided into two steps:
(1) derive the mapping from the risk-aware state-action function to the optimal policy;
(2) establish the equivalence between the BT model and the Regret Preference Model.

Specifically, starting from Equation (\ref{Equation: Ra-RL objective function}), the mapping from the risk-aware state-action function $\tilde{Q}_\pi$ to the optimal policy $\pi_{\theta}^{\ast}$ can be derived as stated in the following lemma.
\begin{lemma}   \label{Lemma: the mapping between the risk-aware state-action function and the optimal policy}
    The constrained problem in Equation (\ref{Equation: Ra-RL objective function}) has the closed-form solution:
    \begin{equation}    \label{Equation: the mapping between the risk-aware state-action function and the optimal policy}
        \pi_{\theta}^{\ast} \left(z \mid \left[x, y^{<t}\right]\right) =\frac{\pi_{\mathrm{ref}}\left(z \mid \left[x, y^{<t}\right]\right) \exp\left(\frac{1}{\beta} \tilde{Q}_{\pi_{\mathrm{ref}}} \left(\left[x, y^{<t}\right], z\right)\right)}{Z\left(\left[x, y^{<t}\right]; \beta\right)},
    \end{equation}
    where $Z\left(\left[x, y^{<t}\right]; \beta\right) = \mathbb{E}_{z \sim \pi_{\mathrm{ref}}(\cdot \mid [x, y^{<t}])}e^{\frac{1}{\beta} \tilde{Q}_{\pi_{\mathrm{ref}}} \left(\left[x, y^{<t}\right], z\right)}$ is the partition function.
\end{lemma}
The proof is provided in Appendix \ref{Appendix: the mapping between the risk-aware state-action function and the optimal policy lemma}.
Then, by rearranging Equation (\ref{Equation: the mapping between the risk-aware state-action function and the optimal policy}), we obtain the expression of the risk-aware state-action function in terms of the policy:
\begin{equation}   \label{Equation: the risk-aware state-value function}
    \tilde{Q}_{\pi_{\mathrm{ref}}} \left(\left[x, y^{<t}\right], z\right) =\beta \log \frac{\pi_\theta^*\left(z \mid\left[x, y^{<t}\right]\right)}{\pi_{\mathrm{ref}}\left(z \mid\left[x, y^{<t}\right]\right)} + \beta \log Z\left(\left[x, y^{<t}\right]; \beta\right).
\end{equation}

Subsequently, by utilizing the reward decomposition formula $r = \sum_{t=1}^T \gamma^{t-1} R\left(\left[x, y^{<t}\right], y^t\right)$ from Lemma \ref{Lemma: Equivalent MDP}, we establish BT model equivalence with the Regret Preference Model as shown in the following lemma, whose proof is provided in Appendix \ref{Appendix: the proof of the equivalence between the BT model and the Regret Preference Model lemma}.
\begin{lemma}   \label{Lemma: the equivalence between the BT model and the Regret Preference Model}
    Given a reward function $r(x, y)$ of the entire prompt-response, based on the relationship between token-wise rewards and the reward function $r(x, y) = \sum_{t=1}^T \gamma^{t-1} R\left(\left[x, y^{<t}\right], y^t\right)$, we can establish the equivalence between the Bradley-Terry model and the Regret Preference Model, i.e.,
    \begin{equation}   \label{Equation: the equivalence between the BT model and the Regret Preference Model}
        P_{\mathrm{BT}}\left(y_1 \succ y_2 \mid x\right) = \sigma\left(\sum_{t=1}^{T_1} \gamma^{t-1} \tilde{A}_\pi\left(\left[x, y_1^{<t}\right], y_1^t\right) -\sum_{t=1}^{T_2} \gamma^{t-1} \tilde{A}_\pi\left(\left[x, y_2^{<t}\right], y_2^t\right)\right),
    \end{equation}
    where $\sigma(z) = 1 /\left(1+\exp (-z)\right)$ is the logistic sigmoid function for any random variable $z$.
\end{lemma}

According to the definition of the risk-aware advantage function in Definition~\ref{Definition: RA-advantage}, we can directly establish the relationship between the optimal solution in Equation~(\ref{Equation: the risk-aware state-value function}) and preference optimization objective in Equation~(\ref{Equation: the equivalence between the BT model and the Regret Preference Model}).
In this way, we reformulate the BT model to be directly tied to the risk-aware optimal policy $\pi_{\theta}^{\ast}$ and the reference policy $\pi_{\mathrm{ref}}$, which is summarized in the following theorem, whose proof is provided in the Appendix~\ref{Appendix: the proof of the Bradley Terry model express the human preference probability theorem}.
\begin{theorem}   \label{Theorem: the Bradley Terry model express the human preference probability}
    Given prompts $x$ and pairwise responses $\left(y_1, y_2\right)$, and the risk-aware objective function in Equation~(\ref{Equation: Ra-RL objective function}), the Bradley-Terry model expresses the human preference probability in terms of the risk-aware optimal policy $\pi_\theta^*$ and reference policy $\pi_{\mathrm{ref}}$:
    \begin{equation}   \label{Equation: the Bradley Terry model express the human preference probability}
        P_{\mathrm{BT}}^{\ast} \left(y_1 \succ y_2 \mid x\right)=\sigma\left(u^{\ast} \left(x, y_1, y_2\right) - \delta^{\ast} \left(x, y_1, y_2\right)\right),
    \end{equation}
    where $u\left(x, y_1, y_2\right)$ represents the difference in implicit rewards defined by the risk-aware policy $\pi_{\theta}^{\ast}$ and the reference policy $\pi_{\mathrm{ref}}$, weighted by $\beta$, represented as:
    \begin{equation}
        u\left(x, y_1, y_2\right) = \beta\log\frac{\pi_{\theta}\left(y_1 \mid x\right)}{\pi_{\mathrm{ref}}\left(y_1 \mid x\right)} - \beta\log\frac{\pi_{\theta}\left(y_2 \mid x\right)}{\pi_{\mathrm{ref}}\left(y_2 \mid x\right)},
    \end{equation}
    and $\delta\left(x, y_1, y_2\right)$ represents the difference in sequential risk ratios between two pairs $\left(x, y_1\right)$ and $\left(x, y_2\right)$, expressed as:
    \begin{equation}
        \delta\left(x, y_1, y_2\right)= \beta D_{\mathrm{SeqRR}} \left(x, y_2; \pi_{\mathrm{ref}} \mid \pi_{\theta}\right) -\beta D_{\mathrm{SeqRR}}\left(x, y_1 ; \pi_{\mathrm{ref}} \mid \pi_{\theta}\right),
    \end{equation}
    where $D_{\mathrm{SeqRR}}\left(x, y; \pi_{\mathrm{ref}} \mid \pi_{\theta}\right) = \sum_{t=1}^T \operatorname{\Phi}^{\mu}_{z \sim \pi_{\mathrm{ref}}} \left(\log \frac{\pi_\mathrm{ref}\left(z \mid\left[x, y^{<t}\right]\right)}{\pi_{\theta}\left(z \mid\left[x, y^{<t}\right]\right)}\right).$
\end{theorem}

\subsection{Loss Function and Formal Analysis}
Drawing on Theorem~\ref{Theorem: the Bradley Terry model express the human preference probability}, we reformulate the BT model into a structure solely relevant to the risk-sensitive policy, which enables us to formulate a likelihood maximization objective for a parametrized policy $\pi_{\theta}$.
The loss function is given by:
\begin{equation}   \label{Equation: Ra-DPO_1 loss}
    \mathcal{L}_{\text{Ra-DPO}_\text{1}} \left(\pi_{\theta}; \pi_{\mathrm{ref}}\right) =-\mathbb{E}_{\left(x, y_w, y_l\right) \sim \mathcal{D}}\left[\log \sigma\left(u\left(x, y_w, y_l\right) - \delta\left(x, y_w, y_l\right)\right)\right].
\end{equation}

In Equation~(\ref{Equation: Ra-DPO_1 loss}), the sequential risk ratio is explicitly introduced into the loss function, which incorporates risk-awareness to balance alignment performance and model drift.
To elucidate the benefits of the proposed method, we conduct the further analysis of the loss function and its corresponding gradient.
For brevity, we use $u$ to denote $u\left(x, y_w, y_l\right)$, and $\delta$ to represent $\delta\left(x, y_w, y_l\right)$.
By simple calculations, we can derive the gradient of the loss function in Equation (\ref{Equation: Ra-DPO_1 loss}) with respect to parameter $\theta$ :
\begin{equation}   \label{Equation: the gradient of Ra-DPO_1 loss}
    \nabla_{\theta} \mathcal{L}_{\text{Ra-DPO}_\text{1}} \left(\pi_{\theta}; \pi_{\mathrm{ref}}\right) =-\mathbb{E}_{\left(x, y_w, y_l\right) \sim \mathcal{D}}\left[(-u+\delta)\left[\nabla_{\theta} u-\nabla_{\theta} \delta\right]\right],
\end{equation}
where $(-u+\delta)$ serves as the weighting factor for the gradient.

From Equation~(\ref{Equation: the gradient of Ra-DPO_1 loss}), we can observe that the first part, $(-u)$, corresponds to the weight factor in the first part of loss function of TDPO.
Its value increases when the language model makes prediction errors relative to human preferences, i.e., $\log\frac{\pi_{\theta}\left(y_l \mid x\right)}{\pi_{\mathrm{ref}}\left(y_l \mid x\right)}> \log \frac{\pi_{\theta} \left(y_w \mid x\right)}{\pi_{\mathrm{ref}}\left(y_w \mid x\right)}$.
The second part, $\delta$, consists of the difference between the sequential risk ratios of the dispreferred and preferred response subsets, which is a distinctive component of our method.
When selecting a convex function (risk-averse), such as CVaR, as the risk measure, our method can automatically control the risk ratio balance.

Furthermore, building upon a common objective shared by our method and TDPO \cite{2024_zeng_tdpo}, i.e., reducing risks stemming from model drift and ensuring training stability, we further provide a second version of our method, $\text{Ra-DPO}_\text{2}$.
The loss function of $\text{Ra-DPO}_\text{2}$ is given by:
\begin{equation}   \label{Equation: Ra-DPO_2 loss}
    \mathcal{L}_{\text{Ra-DPO}_\text{2}} \left(\pi_{\theta}; \pi_{\mathrm{ref}}\right) =-\mathbb{E}_{\left(x, y_w, y_l\right) \sim \mathcal{D}}\left[\log \sigma\left(u\left(x, y_w, y_l\right) - \alpha \delta_2 \left(x, y_w, y_l\right)\right)\right],
\end{equation}
where $\delta_2 \left(x, y_1, y_2\right) =\beta D_{\mathrm{SeqRR}} \left(x, y_2; \pi_{\mathrm{ref}} \mid \pi_{\theta}\right) -\operatorname{sg} \left(\beta D_{\mathrm{SeqRR}}\left(x, y_1 ; \pi_{\mathrm{ref}} \mid \pi_{\theta}\right)\right)$.
\begin{wrapfigure}{r}{8cm}
    \vskip -0.1in
    \centering
    \includegraphics[width=0.95 \linewidth]{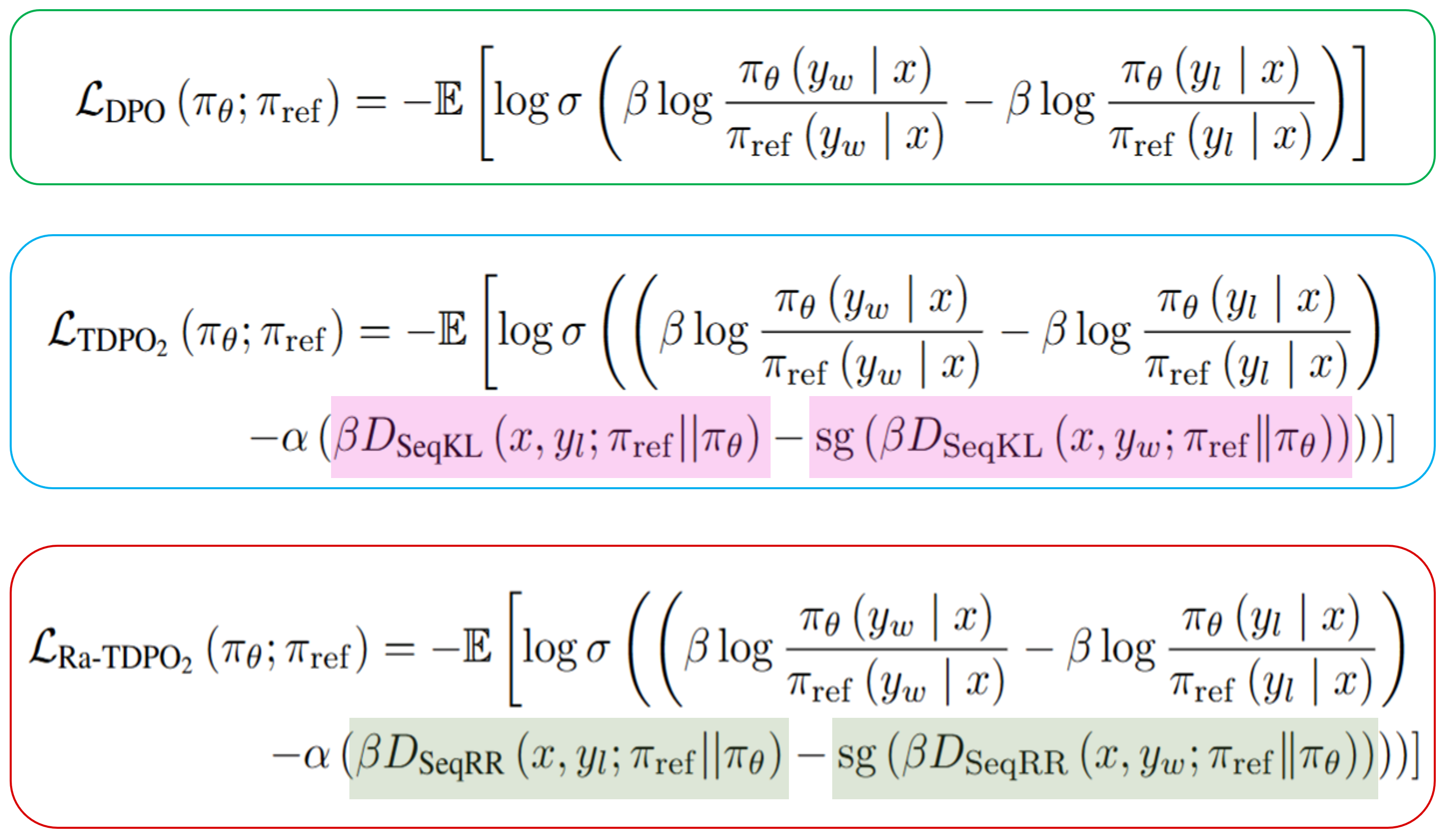}
    \caption{
        Comparison of loss functions for DPO, $\text{TDPO}_\text{2}$ and $\text{Ra-DPO}_\text{2}$ methods. The $\operatorname{sg}$ denotes the stop-gradient operator.
    }
    \label{Fig.1}
    \vskip -0.1in
\end{wrapfigure}
The operator $\operatorname{sg}$ represents the stop-gradient operator, which blocks the propagation of gradients. 
The parameter $\beta$ can control the deviation between $D_{\mathrm{SeqRR}} \left(x, y_2; \pi_{\mathrm{ref}} \mid \pi_{\theta}\right)$ and $\left(\beta D_{\mathrm{SeqRR}}\left(x, y_1 ; \pi_{\mathrm{ref}} \mid \pi_{\theta}\right)\right)$.
$\text{Ra-DPO}_\text{2}$ modifies the loss function of $\text{Ra-DPO}_\text{1}$ by disabling the gradient propagation of $D_{\mathrm{SeqRR}} (x, y_w; \pi_{\mathrm{ref}} \mid \pi_{\theta})$ and treating it as a baseline term for alignment of $D_{\mathrm{SeqRR}} (x, y_l; \pi_{\mathrm{ref}} \mid \pi_{\theta})$.
The aim of the modification is to ensure training stability, rather than to accelerate training speeding.
To summarize, the comparison of the loss functions for DPO, $\text{TDPO}_\text{2}$, and $\text{Ra-DPO}_\text{2}$ is shown in Figure~\ref{Fig.1}.
In addition, we provide a procedure of our method, and provide its pseudocode (Algorithm \ref{Appendix: Ra-DPO algorithm}) in Appendix \ref{Appendix: Algorithm}.

\begin{figure*}[htb]
    \vskip -0.1in
    \centering
    \subfigure[]{
        \includegraphics[width=0.31 \linewidth]{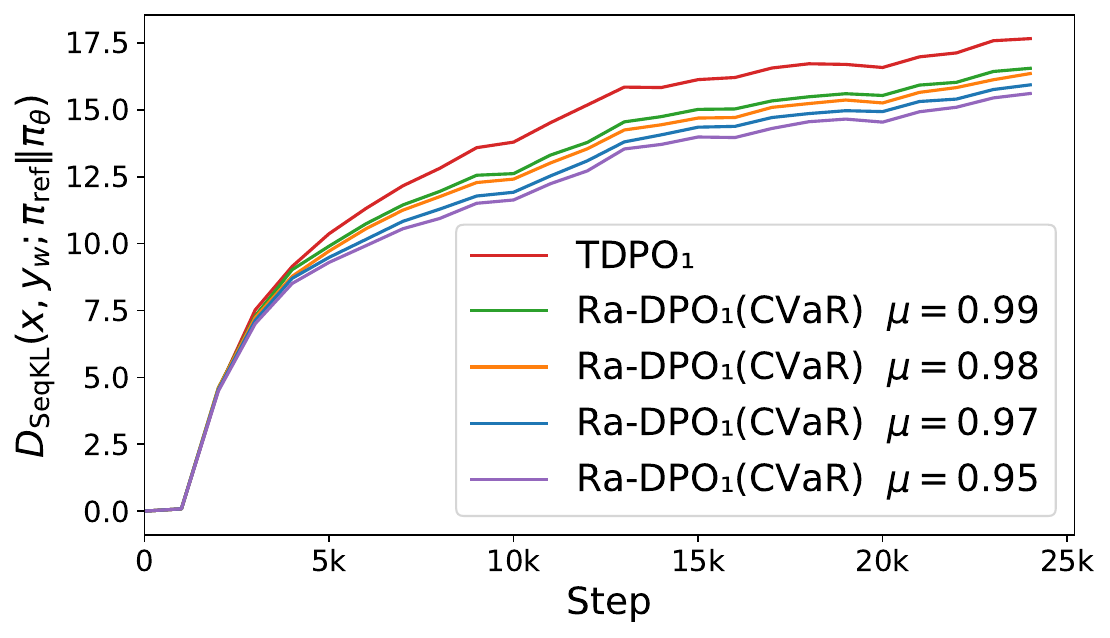}
        \label{subfig2.1}
    }
    \subfigure[]{
        \includegraphics[width=0.31 \linewidth]{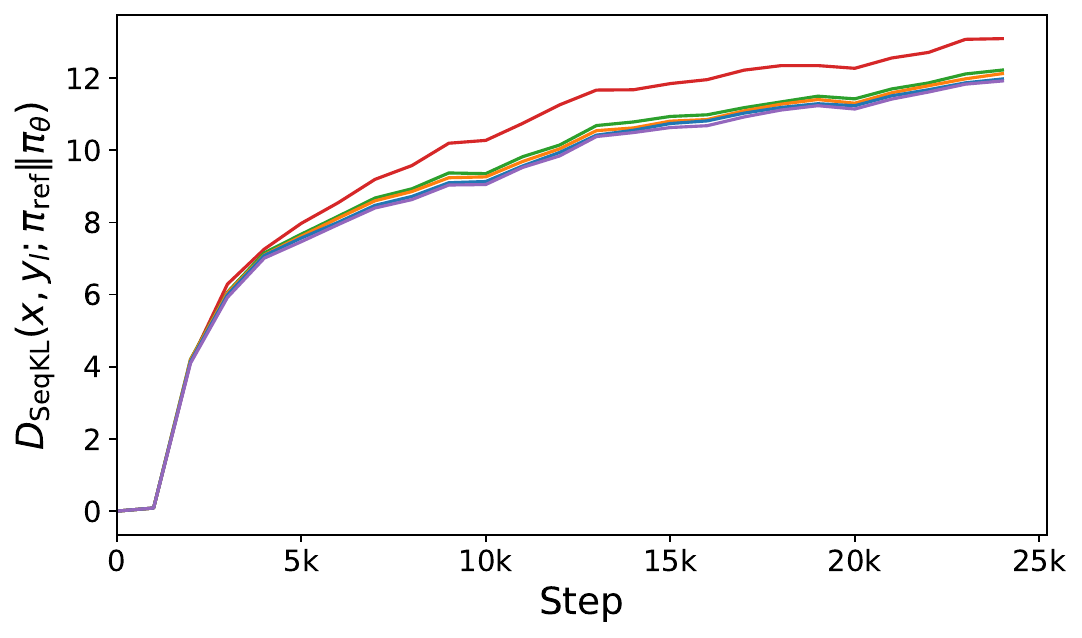}
        \label{subfig2.2}
    }
    \subfigure[]{
        \includegraphics[width=0.31 \linewidth]{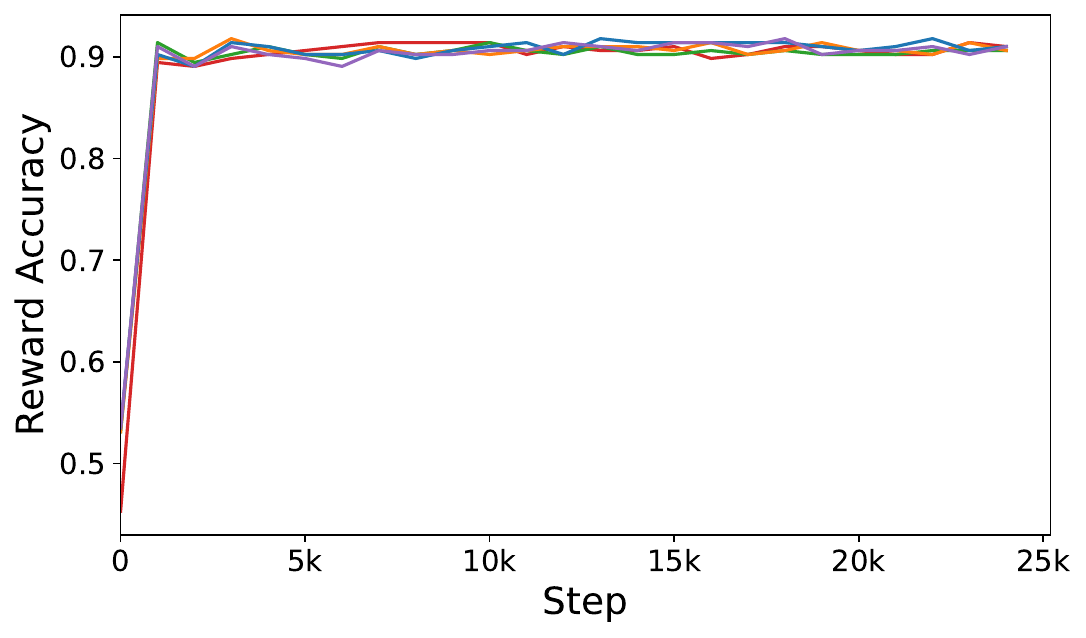}
        \label{subfig2.3}
    }
    \caption{
        The experiment on the IMDb dataset with GPT-2 Large serving as the base model.
        (a) and (b) present the progression of sequential KL divergence (the lower the better) for both preferred and dispreferred responses.
        (c) illustrates the reward accuracy curves (the higher the better).
    }
    \label{Fig.2}
    \vskip -0.1in
\end{figure*}

\begin{figure*}[htb]
    % \vskip -0.1in
    \centering
    \subfigure[]{
        \includegraphics[width=0.31 \linewidth]{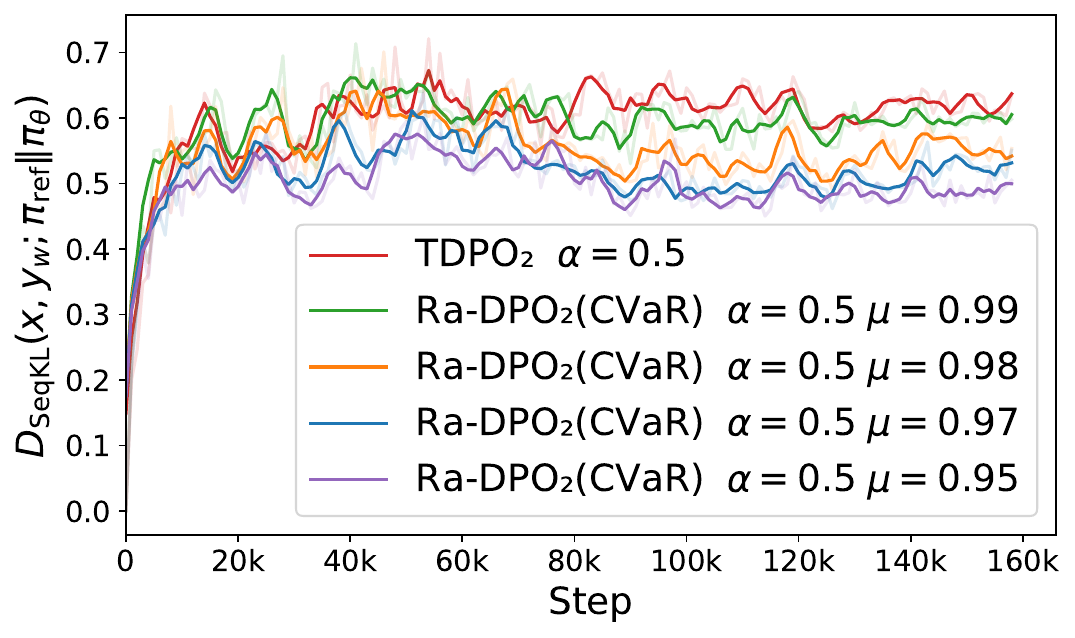}
        \label{subfig3.1}
    }
    \subfigure[]{
        \includegraphics[width=0.31 \linewidth]{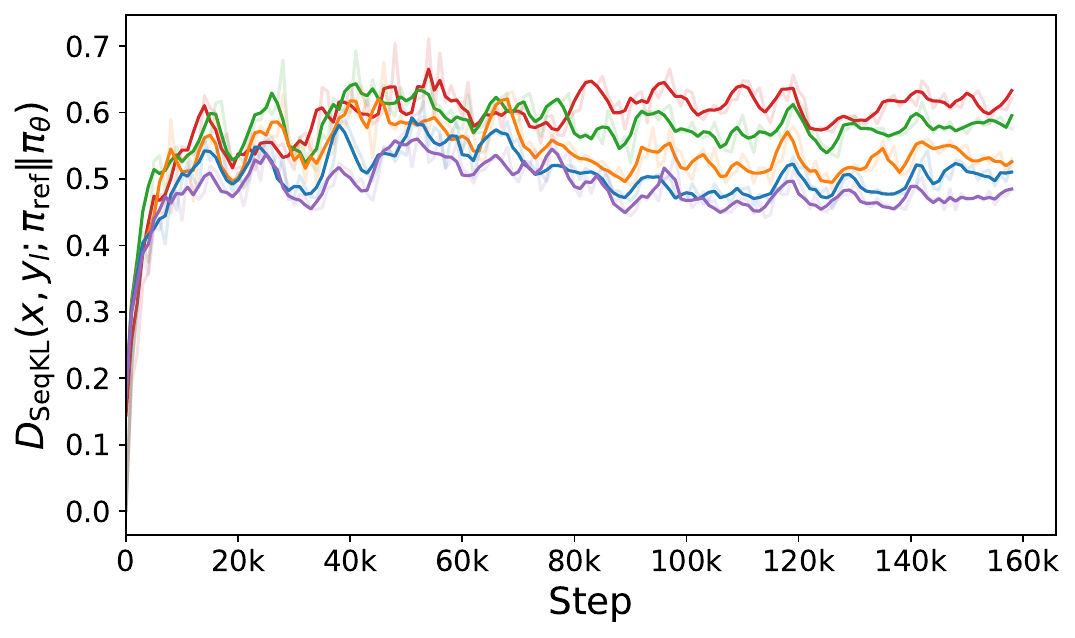}
        \label{subfig3.2}
    }
    \subfigure[]{
        \includegraphics[width=0.31 \linewidth]{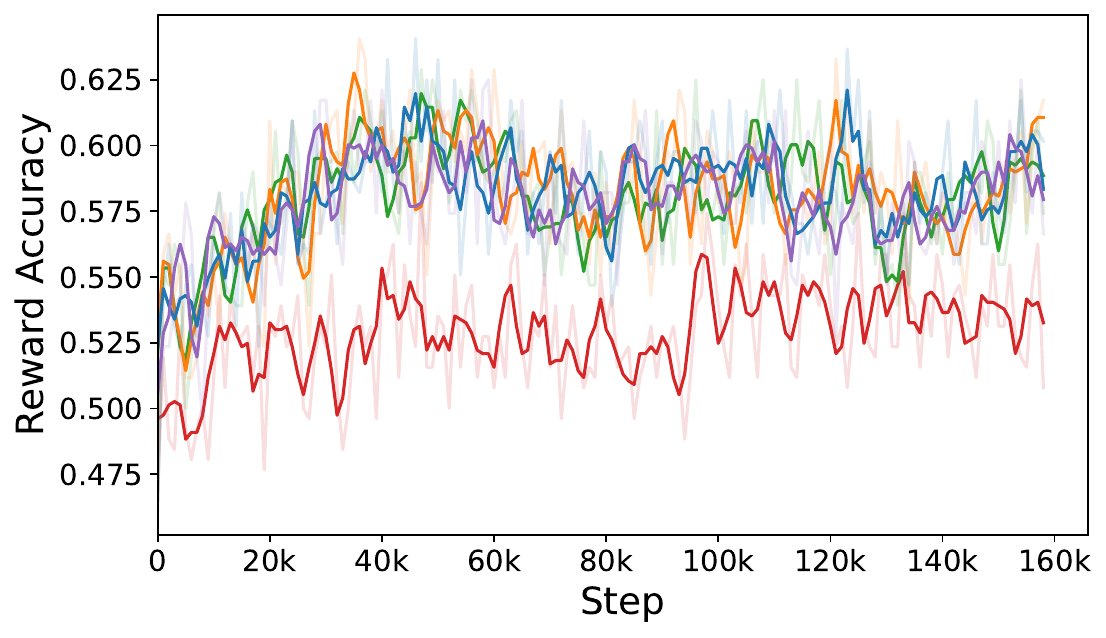}
        \label{subfig3.3}
    }
    \subfigure[]{
        \includegraphics[width=0.31 \linewidth]{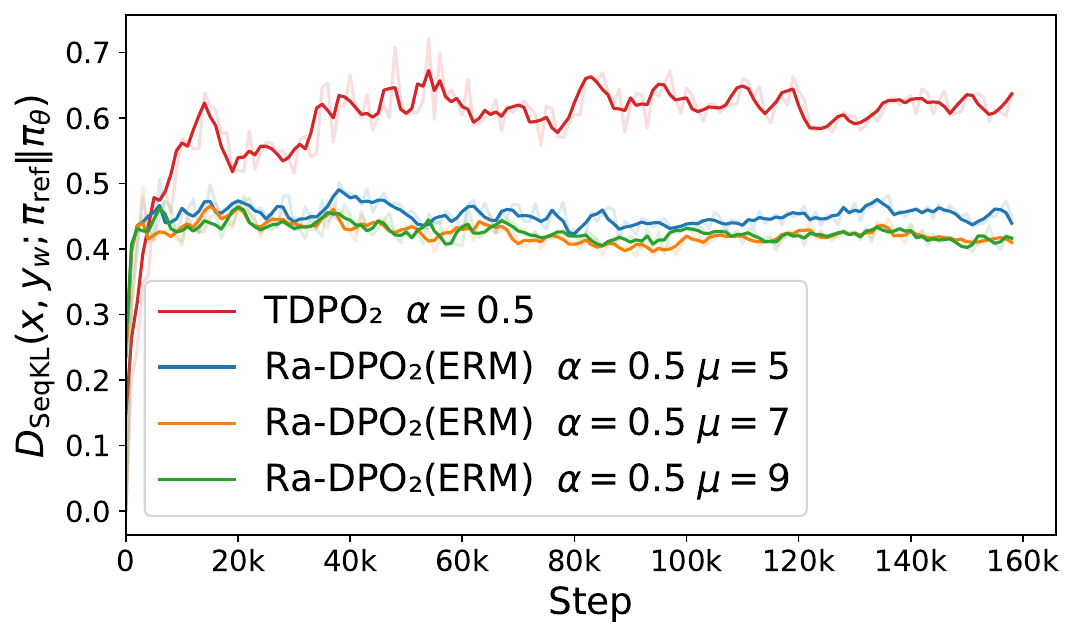}
        \label{subfig3.4}
    }
    \subfigure[]{
        \includegraphics[width=0.31 \linewidth]{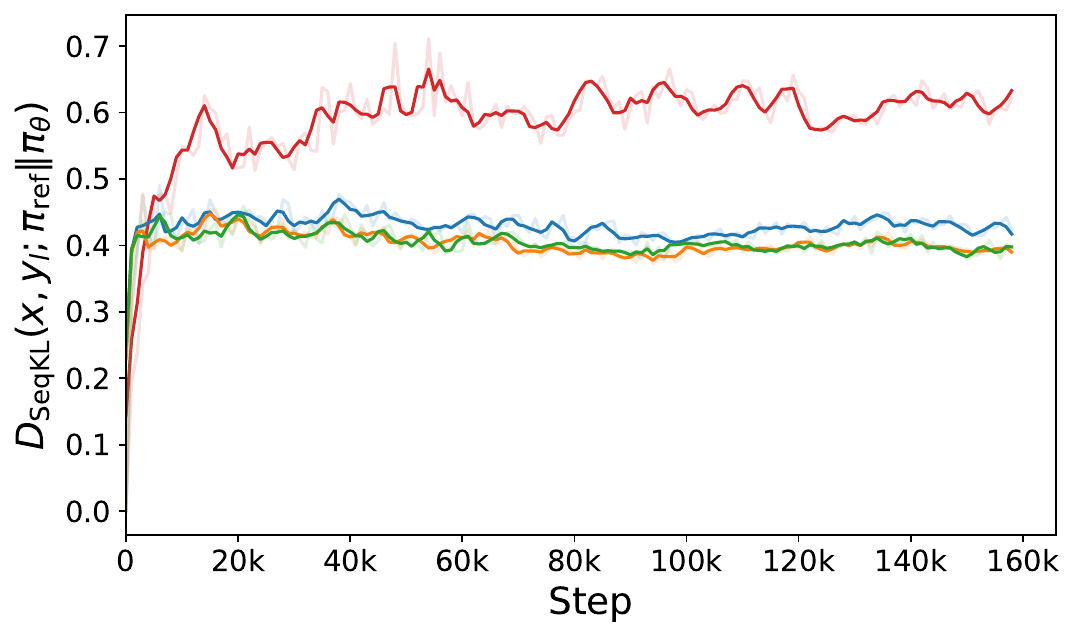}
        \label{subfig3.5}
    }
    \subfigure[]{
        \includegraphics[width=0.31 \linewidth]{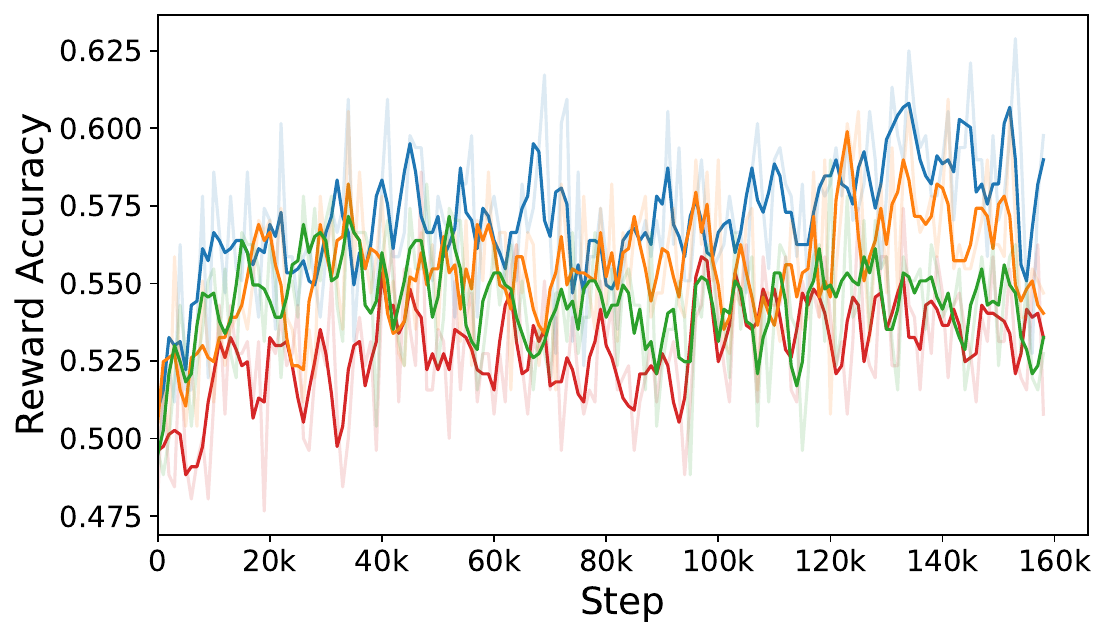}
        \label{subfig3.6}
    }
    \caption{
        The experiment on the Anthropic HH dataset with Pythia-1.4B serving as the base model.
        \textbf{Left} and \textbf{Middle} present the progression of sequential KL divergence (the lower the better) for both preferred and dispreferred responses.
        \textbf{Right} illustrates reward accuracy curves (the higher the better).
    }
    \label{Fig.3}
    \vskip -0.1in
\end{figure*}

%%%%%%%%%%%%%%%%%%%%%%%%%%%%%%%%%%%%%%%%%%%%%%
%% Section 4 Experimental Setup
%%%%%%%%%%%%%%%%%%%%%%%%%%%%%%%%%%%%%%%%%%%%%%
\section{Experiments}   \label{Section: Experiments}
We empirically evaluate our method on several open-source datasets and pre-trained models, aiming to investigate the following questions:
(1) How does the performance of our method compare with that of existing methods, particularly in terms of risk sensitivity when handling challenging text generation tasks?
(2) How does the risk control parameter $\mu$ affect the performance of our method?

To answer these questions, we conducted experiments on IMDb Dataset \cite{2011_maas_imdb}, Anthropic HH Dataset~\cite{2022_bai_Anthropic-HH}, and AlpacaEval \cite{2024_dubois_AlpacaEval} for three different text generation tasks.
Based on the original \emph{KTO implementation}\footnote{Available at \url{https://github.com/ContextualAI/HALOs}}, we trained Ra-DPO and baseline models using the same hyperparameters. 
Specifically, for Ra-DPO, we employed nested risk measures based on CVaR \cite{2000_rockafellar_CVaR} and ERM \cite{2023_hau_ERM}.
We compare our method against the following algorithms:
(1) DPO \cite{2023_rafailov_dpo}, which only considers evaluation at the sentence level;
(2) PPO \cite{2017_schulman_ppo}, an offline PPO variant provided by the original KTO implementation;
(3) $\text{TDPO}_\text{1}$ and $\text{TDPO}_\text{2}$ \cite{2024_zeng_tdpo}, which convert the BT model into token-level representations;
(4) KTO \cite{2024_ethayarajh_kto}, which considers preferences in human decisions that are not aimed at maximizing utility.
Experimental setup and results are reported in Subsections~\ref{Subsection: Experiments on IMDb Dataset}-\ref{Subsection: Experiments on AlpacaEval} and Appendix~\ref{Appendix: Experiments}.

\subsection{Experiments on IMDb Dataset}   \label{Subsection: Experiments on IMDb Dataset}
\textbf{Experimental Setup:}
The IMDb dataset is a controlled semantic generation dataset within the context of movie reviews, serving as a valuable resource for training and evaluating sentiment analysis models.
We employ GPT-2 Large \cite{2019_radford_gpt2} as the base model and use the model checkpoint \emph{insub/gpt2-large-IMDb-fine-tuned}\footnote{\url{https://huggingface.co/insub/gpt2-large-IMDb-fine-tuned}} as the SFT model.
The results of the versions of $\text{Ra-DPO}_\text{1}$ (CVaR) with risk control parameter $\mu \in \{0.99, 0.98, 0.97, 0.95\}$ are shown in Figure~\ref{Fig.2}.

\textbf{Evaluation:}
Figure~\ref{Fig.2} shows that $\text{Ra-DPO}_\text{1}$ can outperform or achieve reward accuracy similar to the advanced $\text{TDPO}$ algorithm while reducing model drift (i.e., lower sequential KL divergence), demonstrating the risk-awareness of $\text{Ra-DPO}_\text{1}$ in balancing alignment performance and model drift.

\begin{figure*}[htb]
    \vskip -0.1in
    \centering
    \subfigure[]{
        \includegraphics[width=0.31 \linewidth]{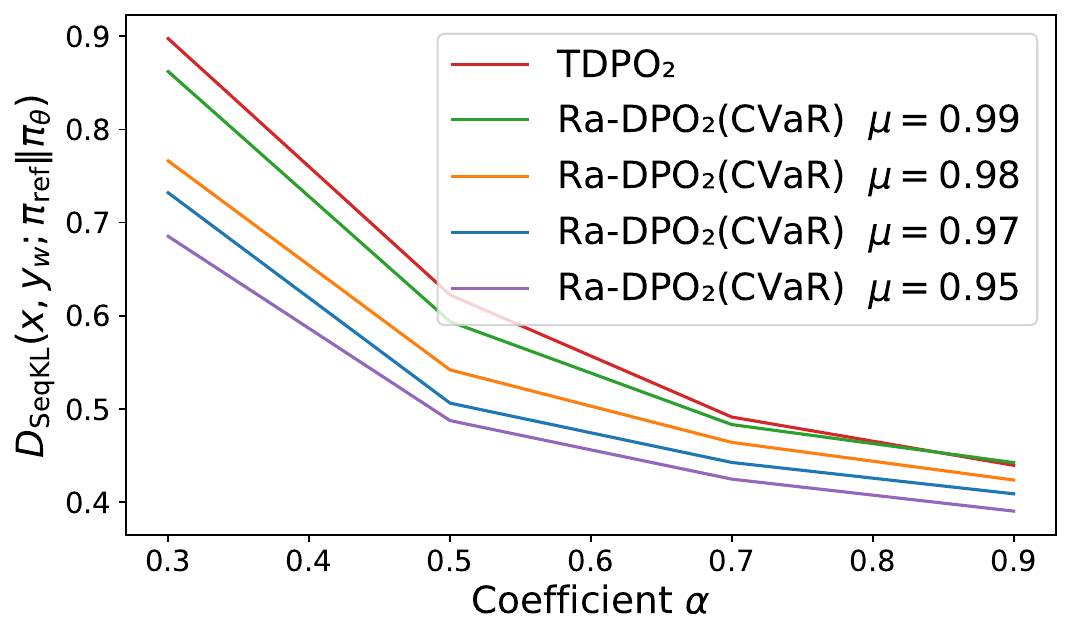}
        \label{subfig4.1}
    }
    \subfigure[]{
        \includegraphics[width=0.31 \linewidth]{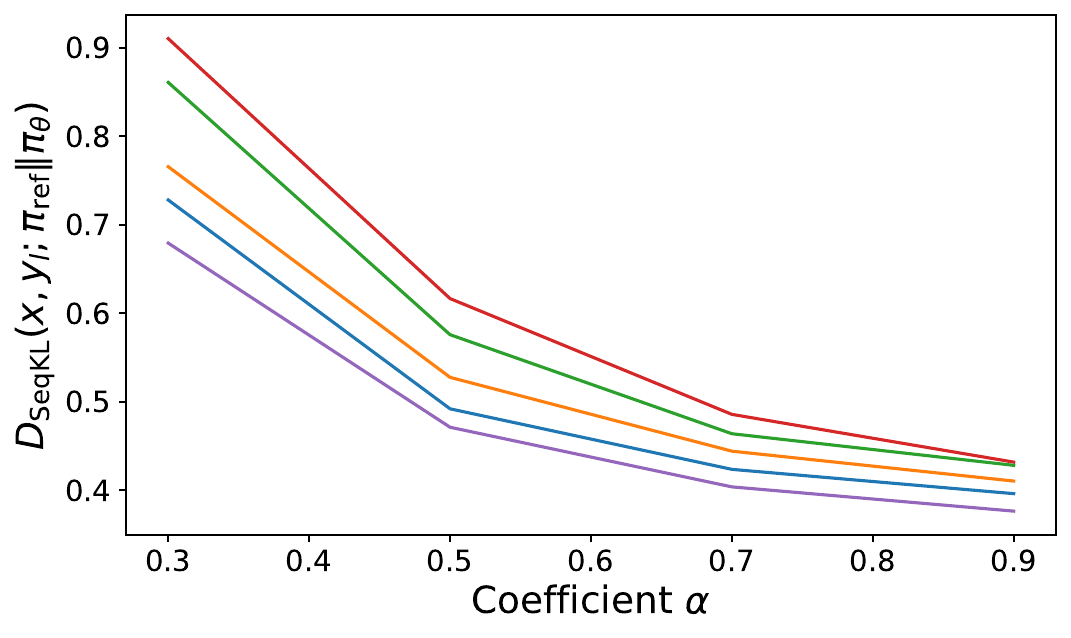}
        \label{subfig4.2}
    }
    \subfigure[]{
        \includegraphics[width=0.31 \linewidth]{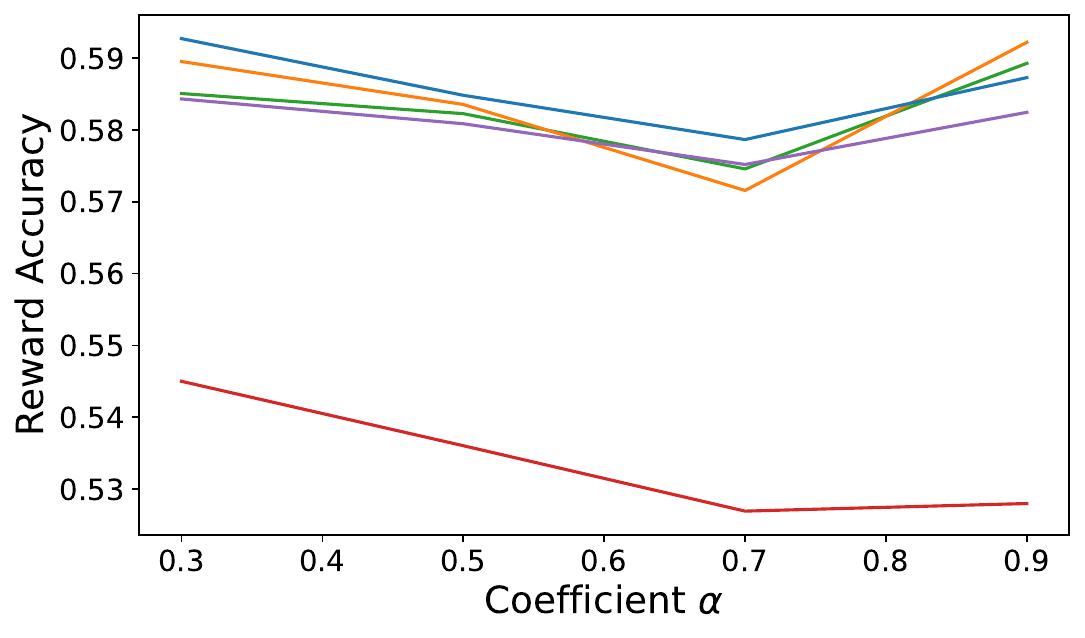}
        \label{subfig4.3}
    }
    \subfigure[]{
        \includegraphics[width=0.31 \linewidth]{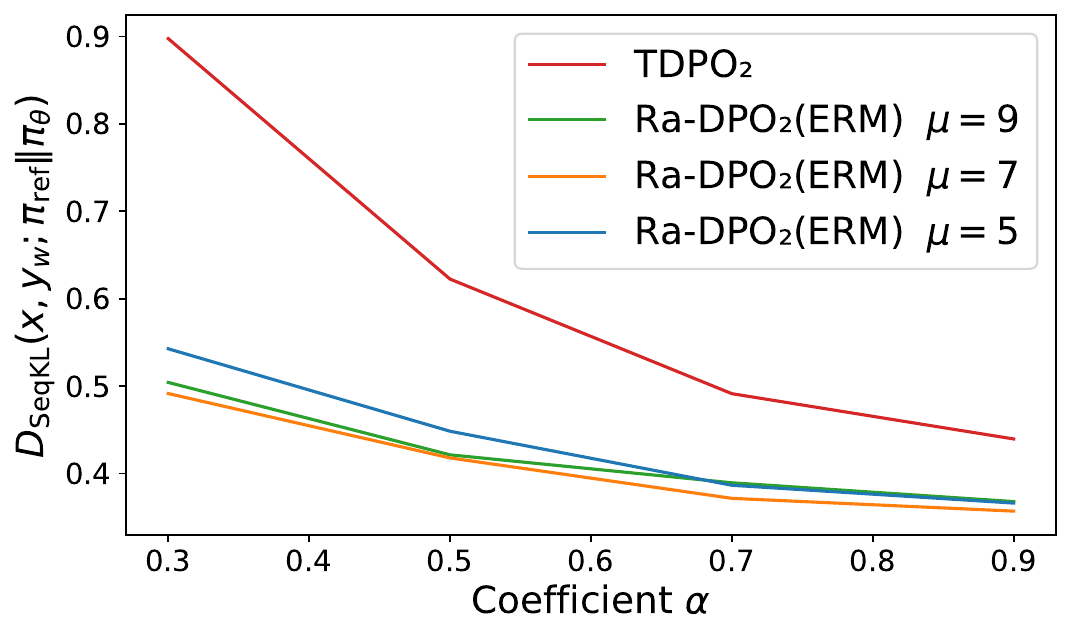}
        \label{subfig4.4}
    }
    \subfigure[]{
        \includegraphics[width=0.31 \linewidth]{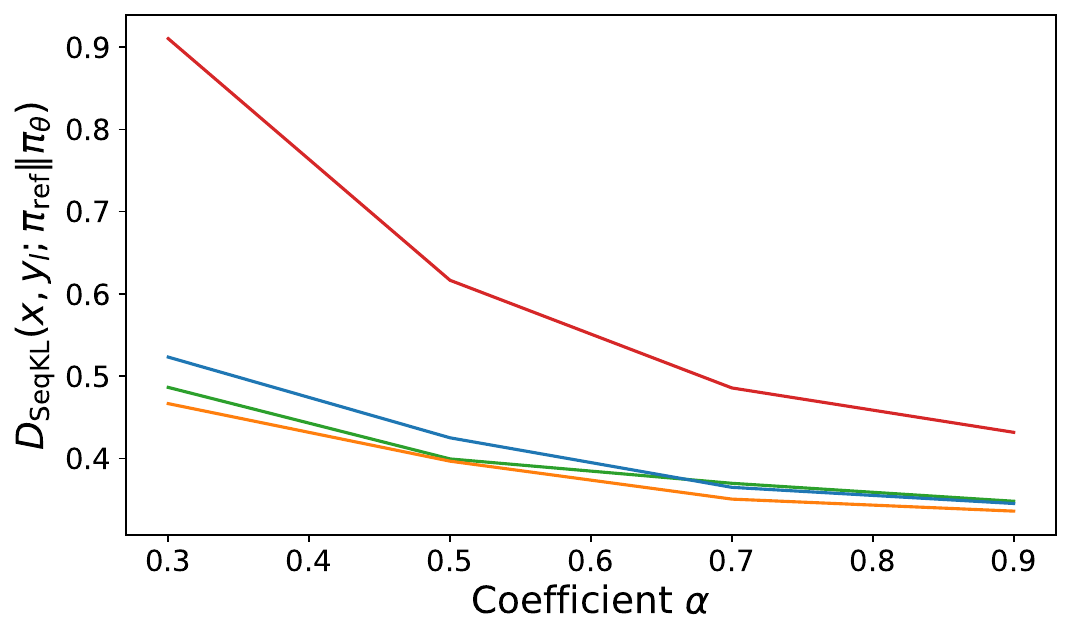}
        \label{subfig4.5}
    }
    \subfigure[]{
        \includegraphics[width=0.31 \linewidth]{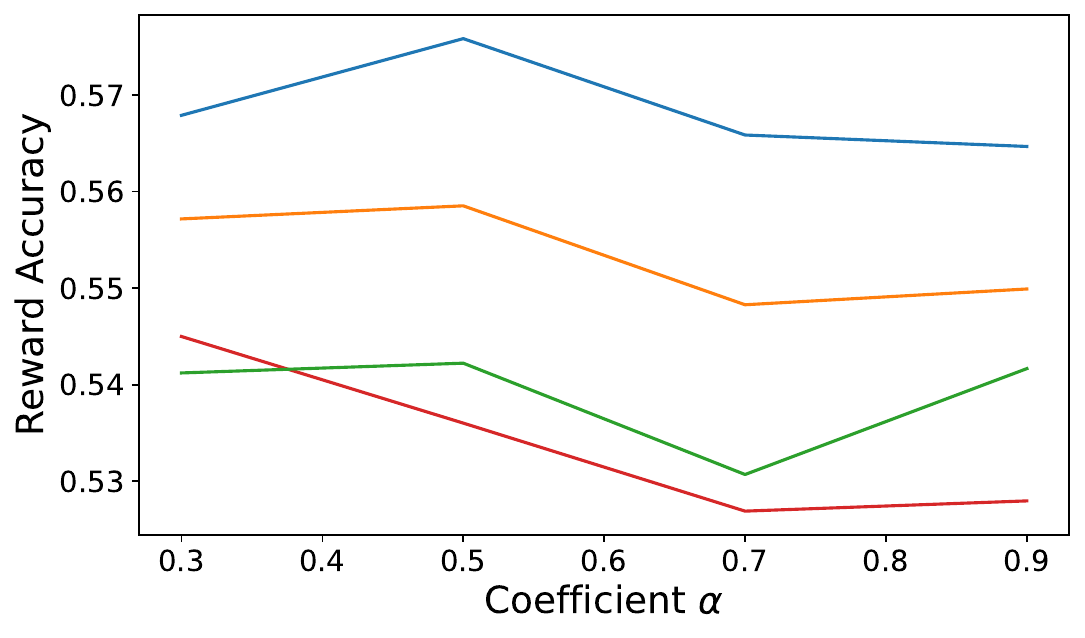}
        \label{subfig4.6}
    }
    \caption{
        The experiment on the Anthropic HH dataset with Pythia-1.4B serving as the base model.
        \textbf{Left} and \textbf{Middle} presents the sequential KL divergence (the lower the better) for preferred and dispreferred responses, while \textbf{Right} presents the reward accuracy curves (the higher the better) under $\alpha = \{0.3, 0.5, 0.7, 0.9\}$.
    }
    \label{Fig.4}
    \vskip -0.1in
\end{figure*}

\subsection{Experiments on Anthropic HH Dataset}   \label{Subsection: Experiments on Anthropic HH Dataset}
\textbf{Experimental Setup: }
Anthropic HH dataset contains 170k dialogues between a human and an automated assistant, where each transcript ends with a pair of responses generated by an LLM along with a preference label denoting the human-preferred response.
We use Pythia-1.4B and Pythia-2.8B~\cite{2023_biderman_pythia} as the base models to test our method on Anthropic HH dataset, respectively.
The reference models are trained by fine-tuning the base models on chosen completions.
The results are depicted in Figure~\ref{Fig.3}, Figure~\ref{Fig.4}, and Appendix~\ref{Appendix: Additional experimental results}.

\textbf{Evaluation: }
Figure~\ref{Fig.3} shows the performance of $\text{TDPO}_\text{2}$ and different versions of $\text{Ra-DPO}_\text{2}$ with respect to the risk control parameter $\mu$ while keeping coefficient $\alpha$ constant at 0.5.
Figure~\ref{Fig.4} presents the statistical results of different algorithms with coefficient $\alpha = \{0.3, 0.5, 0.7, 0.9\}$, and the corresponding curve plots are provided in Appendix~\ref{Appendix: Additional experimental results}.
From Figures~\ref{Fig.3} and \ref{Fig.4}, we can observe that $\text{Ra-DPO}_\text{2}$ almost always achieves superior performance (higher reward accuracy) and maintains minimal model drift (lower sequential KL divergence), under both CVaR-based and ERM-based nested risk measures.
Additionally, Figure~\ref{Fig.4} illustrates that $\text{Ra-DPO}_\text{2}$ is highly effective in suppressing deviation from reference model, particularly when $\alpha$ takes on smaller values.
We hypothesize that this phenomenon may be attributed to the fact that the sequential risk ratio accumulates step-wise risks through nested risk measures, enabling it to remain responsive to significant model biases even when $\delta_2 \left(x, y_1, y_2\right)$ in Equation (\ref{Equation: Ra-DPO_2 loss}) carries a relatively low weight.

\subsection{Experiments on AlpacaEval}   \label{Subsection: Experiments on AlpacaEval}
\begin{wraptable}
    {r}{0.45\textwidth}
    \vskip -0.2in
    \caption{
        The compare between different Algorithms and \emph{gpt4\_1106\_preview}.
    }
    \label{Table 1}
    \begin{center}
        \begin{small}
            % \begin{sc}
                \begin{tabular}{lcccr}
                    \toprule
                    Method                   & Winrate                & Lc winrate             \\
                    \midrule
                    DPO                      & 51.1$\pm$ 1.9          & 44.7$\pm$ 0.4          \\
                    PPO                      & 52.1$\pm$ 1.8          & 51.9$\pm$ 0.5          \\
                    KTO                      & 51.5$\pm$ 1.8          & 50.2$\pm$ 0.6          \\
                    $\text{TDPO}_\text{1}$   & 51.9$\pm$ 1.8          & 53.0$\pm$ 0.6          \\
                    $\text{TDPO}_\text{2}$   & 52.2$\pm$ 1.6          & 52.2$\pm$ 0.5          \\
                    $\text{Ra-DPO}_\text{1}$ & \textbf{53.5$\pm$ 1.8} & 53.9$\pm$ 0.5          \\
                    $\text{Ra-DPO}_\text{2}$ & 52.1$\pm$ 1.8          & \textbf{55.7$\pm$ 0.5} \\
                    \bottomrule
                \end{tabular}
            % \end{sc}
        \end{small}
    \end{center}
    \vskip -0.2in
\end{wraptable}
\textbf{Experimental Setup:}
To comprehensively evaluate $\text{Ra-DPO}_\text{2}$ in terms of generation quality, we conducted pairwise comparisons on AlpacaEval using models trained on the Anthropic HH dataset.
Following the official \emph{AlpacaEval implementation}\footnote{\url{https://github.com/tatsu-lab/alpaca\_eval}}, we sampled responses with a temperature coefficient of $0.7$.
The winrate comparisons based on \emph{oasst\_pythia\_12b}\footnote{\url{https://huggingface.co/OpenAssistant/oasst-sft-4-pythia-12b-epoch-3.5}} are summarized in Table~\ref{Table 1} and Appendix \ref{Appendix: Additional experimental results}.
Both winrate and length-controlled winrate (Lc winrate) are evaluated based on \emph{oasst\_pythia\_12b}.

\textbf{Evaluation:}
Table~\ref{Table 1} reveals that under the two indicators of winrate and length-controlled winrate, most of the implemented algorithms can outperform the common default baseline \emph{gpt4\_1106\_preview} (DPO is more prone to generating long responses).
Among them, $\text{Ra-DPO}_\text{1}$ and $\text{Ra-DPO}_\text{2}$ demonstrate the highest level of performance, especially when it comes to the length-controlled winrate indicator.

%%%%%%%%%%%%%%%%%%%%%%%%%%%%%%%%%%%%%%%%%%%%%%
%% Section 5 Related Work
%%%%%%%%%%%%%%%%%%%%%%%%%%%%%%%%%%%%%%%%%%%%%%
\section{Related Work} \label{Section: Related Work}
\subsection{LLMs Alignment}
With the development of LLMs, numerous researchers have encountered challenges stemming from the misaligned next-token prediction task used in the pre-training stage \cite{2022_bai_Anthropic-HH, 2023_bhardwaj_safety-alignment, 2024_dai_safe-RLHF, 2024_yeh_reliable}, particularly in balancing adherence to human instructions (explicit objectives) with the pursuit of being helpful, honest, and harmless (implicit objectives).
Therefore, a typical post-training stage, referred to as preference optimization, is commonly performed to align pre-trained language models with human intentions, and has become an indispensable aspect in the fine-tuning of LLMs.
Most approaches~\cite{2023_wu_fine-grained-rlhf, 2024_wang_fdpo, 2024_meng_simpo} only utilize KL divergence at the sentence level to limit significant deviations from the reference model.
However, the generation of responses occurs sequentially, following an auto-regressive approach.
Recent works \cite{2024_zeng_tdpo, 2024_ouyang_token-ppo} introduce a fresh perspective, specifically the sequential and token-level direct preference optimization, which allows for examining sequential KL divergence in relation to a reference LLM.
However, due to their neglect of reward distribution characteristics other than the mean, these methods suffer from the trouble of being insensitive to risk.

\subsection{Risk-aware Reinforcement Learning}
RL has made groundbreaking achievements~\cite{2017_arulkumaran_rl_survey, 2022_ouyang_gpt-3, 2024shao_deepseekmath, 2025_guo_deepseek-r1} through approaches such as Q-learning~\cite{2015_mnih_dqn} and policy gradients~\cite{2015_schulman_trpo, 2017_schulman_ppo} in sequential decision tasks, but it also faces challenges when applied in the real world \cite{2022_wang_risk-averse-autonomous-systems, 2022_singh_rl_survey, 2024_gu_saferl_review}.
A primary reason is that the risk-neutral criterion (maximizing the expectation) ignores the characteristics of a reward distribution other than the mean, which may be important for certain systems, especially in applications requiring tight risk control \cite{2020_fei_rsvi, 2022_bisi_rosa}.
In order to tackle this challenge, two types of risk measures have been introduced: nested and static risk measures.
Static risk measures~\cite{2021_fei_static-risk, 2022_bastani_static-risk, 2023_wang_static-risk} are straightforward to interpret, but the resulting optimal policy may not remain Markovian and may become history-dependent.
Nested risk measures~\cite{2022_du_Iterated-CVaR, 2024_chen_Iter-CVaR, 2024_zhao_ra-pbrl} utilize MDPs to ensure risk sensitivity of the value iteration at each step under the current state, resulting in a more conservative approach.
In this paper, we prefer nested risk measures because they recursively adhere to the Bellman equation and allow the MDPs to be reconstructed through state augmentation, enabling them to remain Markovian.

\subsection{Risks in LLMs Alignment}
When aligning LLMs with human preferences, there are many factors that may pose risks, primarily encompassing the following three types:
(1) There may be conflicts among human preferences~\cite{2024_ramesh_grpo}, or human preferences is inherently affected by contextual choice effects \cite{2024_peuter_preference-choice}, thus introducing uncertainty in the objectives when aligning models with human preferences.
(2) Humans do not make decisions by maximizing their expected value for uncertain events; instead, they perceive random variables in a biased but well-defined manner \cite{2024_ethayarajh_kto, 1992_tversky_prospect-theory}.
(3) Many popular methods, such as DPO \cite{2023_rafailov_dpo}, RDPO \cite{2024_fisch_rdpo}, and simPO \cite{2024_meng_simpo}, introduce the new risks during the alignment training process because they only consider the mean of reward or utility, which is risk-neutral and does not capture the distribution characteristics of rewards efficiently.
In this paper, we focus on the third type of risk, which comes from the process during model alignment.

%%%%%%%%%%%%%%%%%%%%%%%%%%%%%%%%%%%%%%%%%%%%%%
%% Section 6 Conclusion
%%%%%%%%%%%%%%%%%%%%%%%%%%%%%%%%%%%%%%%%%%%%%%
\section{Conclusion} \label{Section: Conclusion}
A pressing challenge arises for language generation tasks in the area of risk control, as the models, once trained, are often required to interact directly with humans. In this paper, we propose a novel direct preference optimization method that incorporates risk awareness by introducing nested risk measures into the Bellman equation, to align pre-trained LLMs with human preferences.
Specifically, we design a new risk-aware token-level objective function by reformulating the constrained reward maximization problem into a token-level form and then prove that maximizing this objective function leads to improvements in policy performance.
Then, an optimization objective solely related to the risk-sensitive policy
is obtained by deriving the mapping between the risk-aware state-action function and the optimal policy and establishing BT model equivalence with the Regret Preference Model.
Finally, we conduct a formal analysis of this optimization objective and derive the loss function of Ra-DPO, which has practical implications for language generation tasks.

\newpage
%%%%%%%%%%%%%%%%%%%%%%%%%%%%%%%%%%%%%%%%%%%%%%%%%%%%%%%%%%%%
\bibliographystyle{unsrt}
\bibliography{Ra-TDPO}
\medskip

%%%%%%%%%%%%%%%%%%%%%%%%%%%%%%%%%%%%%%%%%%%%%%%%%%%%%%%%%%%%
\appendix

\newpage
%%%%%%%%%%%%%%%%%%%%%%%%%%%%%%%%%%%%%%%%%%%%%%%%%%%%%%%%%%%%
\section{Supplementary Materials for Section \ref{Section: Preliminaries}}   \label{Appendix: Preliminaries}
\subsection{Risk Measure: A Brief Overview}   \label{Appendix: a survey of Risk Measure}
For quantifying and managing risks, three main paradigms \cite{2014_bauerle_more-RsMDP, 2022_wang_risk-averse-autonomous-systems, 2022_bisi_rosa} have been developed: the risk-neutral paradigm, the worst-case (i.e., robust) paradigm, and the risk-averse paradigm. 
The risk-neutral paradigm aims to find a policy that maximize the expected cumulative reward, but it ignores characteristics of the reward distribution other than the mean, which can be crucial for systems with safety concerns. 
For example, a system may need to operate in a way that mitigates harmful consequences, even in rare and unpredictable situations.
The worst-case paradigm \cite{2005_morimoto_robust-RL, 2018_chen_hamilton} focuses on finding a policy that satisfies the constraints of a specific cost function, generally assuming that the maximum possible cost can quantify bounded adversarial disturbances.
However, since the worst-case approach assumes disturbances are bounded, it may not work well when those bounds are hard to determine.

The risk-averse paradigm \cite{1997_artzner_CVaR, 2014_bauerle_more-RsMDP, 2022_bisi_rosa}, an intermediary paradigm between the risk-neutral and worst-case paradigms, has garnered extensive attention.
It describes individuals or algorithms that prefer outcomes with reduced uncertainty by seeking to optimize risk metrics of the possible cumulative reward, emphasizing its distributional characteristics.
In general, there are mainly two types of risk measures: nested and static risk-aware measures, each possessing distinct advantages and limitations.
Static risk measures \cite{2021_fei_static-risk, 2022_bastani_static-risk, 2023_wang_static-risk} are straightforward to interpret, but the resulting optimal policy may not remain Markovian and may become history-dependent.
On the other hand, nested risk measures \cite{2022_du_Iterated-CVaR, 2024_chen_Iter-CVaR, 2024_zhao_ra-pbrl} utilize MDPs to ensure risk sensitivity of the value iteration at each step under the current state, resulting in a more conservative approach.
We prefer nested risk measures because they recursively adhere to the Bellman equation and allow the MDPs to be reconstructed through state augmentation, thereby enabling them to remain Markovian and ensuring that policy choices depend solely on the current state.

In this paper, we employ a class of nested risk measures, which are variants of the popular CVaR and ERM.
Below, we provide introductions to nested risk measures and the CVaR and ERM risk functions.

Specifically, let $(\mathcal{X}, \mathcal{F})$ be a measurable space. 
A risk measure over $\mathcal{X}$ is a function $\rho : \mathcal{X} \to \mathbb{R}$ that maps uncertain outcomes $X \in \mathcal{X}$ to the real line. 
A risk measure of the total discounted return $G$ can be described as:
\begin{equation}  \label{Equation(appendix): risk measure}
    \min_{\pi \in \Pi} \rho^{\pi}(G),
\end{equation}
where the dependence on $\pi$ emphasizes that the underlying probability measure is induced by the chosen policy. 
The simplest example is $\rho^{\pi} = \mathbb{E}^{\pi}$, for which Equation (\ref{Equation(appendix): risk measure}) reduces to the standard risk-neutral RL problem.

\textbf{Nested Risk-measures: }
Consider a time horizon of length $ T \in \mathbb{T} $. A nested risk measure $ \rho $ of a random return $ G = G_0 + G_1 + \cdots + G_T $ takes the form
\begin{equation}
    \rho(G) = \rho_0 \left( G_0 + \rho_1 \left(G_1 + \cdots + \rho_{T-1} \left(G_{T-1} + \rho_T (G_T) \right) \cdots \right) \right).
\end{equation}
where each $\rho_t$ is a risk functional, i.e., a map from a space of random variables to $\mathbb{R} \bigcup \{\infty\}$.

We now introduce the the CVaR and ERM risk functions.

\textbf{Conditional value-at-risk (CVaR): }  
CVaR with risk-aversion level $\alpha \in (0, 1)$ has been defined as:
$$
\rho_{\mathrm{CVaR}}^{\pi}(G; \alpha) = \min_{\eta \in \mathbb{R}} \left\{ \eta + \frac{1}{1 - \alpha} \mathbb{E}^{\pi} \left[ (G - \eta)^+ \right] \right\},
$$
and it has several advantages over VaR: it quantifies the losses encountered in the tail, it can be expressed as a minimization problem, and is a coherent risk measure \cite{1999_artzner_coherent}.

\textbf{Entropic risk measure (ERM): } 
ERM is a popular method for measuring risk, defined as:
$$
\rho_{\mathrm{ERM}}^{\pi}(G; \beta) := \frac{1}{\beta} \log \mathbb{E}^{\pi}[e^{-\beta G}].
$$
where $\beta > 0$ indicates the degree of risk aversion.
Optimizing ERM is equivalent to optimizing an exponential utility function (EU):
$$\rho_{\mathrm{EU}}^{\pi}(G; \beta) := \mathbb{E}^{\pi}[e^{-\beta G}].$$

In this paper, we formulate the nested risk measure within the Preference-based Markov Decision Process (Pb-MDP) framework, and express it in terms of Bellman equation type.
To simplify notation, we uniformly denote all such nested risk measures collectively by $\operatorname{\Phi}^{\mu}$, a nested risk measure with a risk control parameter $\mu$.

\subsection{The Expanded Version of Value Function Definition}   \label{Appendix: the expanded version of value function definition}
The definition of value function for nested risk measure in Equation (\ref{Equation: Nested PbRL MDP}) can be expanded as
\begin{equation}
    \begin{aligned}
        Q_{\pi}\left(\left[x, y^{<t}\right], y^t\right)
         & =R\left(\left[x, y^{<t}\right], y^t\right) + \operatorname{\Phi}^{\mu}\left(R\left(\left[x, y^{<t+1}\right], \pi\left(\cdot \mid \left[x, y^{<t+1}\right]\right)\right)\right. \\
         & \left. +\operatorname{\Phi}^{\mu} \left(\cdots {\Phi}^{\mu}\left(R\left(\left[x, y^{<T}\right], \pi\left(\cdot \mid \left[x, y^{<T}\right]\right)\right)\right)\right)\right),
    \end{aligned}
\end{equation}
\begin{equation}
    \begin{aligned}
        V_{\pi}\left(\left[x, y^{<t}\right]\right)
         & =R\left(\left[x, y^{<t}\right], \pi\left(\cdot \mid \left[x, y^{<t}\right]\right)\right) + \operatorname{\Phi}^{\mu}\left(R\left(\left[x, y^{<t+1}\right], \pi\left(\cdot \mid \left[x, y^{<t+1}\right]\right)\right)\right. \\
         & \left. +\operatorname{\Phi}^{\mu} \left(\cdots {\Phi}^{\mu}\left(R\left(\left[x, y^{<T}\right], \pi\left(\cdot \mid \left[x, y^{<T}\right]\right)\right)\right)\right)\right).
    \end{aligned}
\end{equation}

Similarly, the definition of the optimal value function, can be expanded as
\begin{equation}
    \begin{aligned}
        Q_{\pi}^{\ast}\left(\left[x, y^{<t}\right], y^t\right)
         & = \max \left\{R\left(\left[x, y^{<t}\right], y^t\right) + \operatorname{\Phi}^{\mu}\left(R\left(\left[x, y^{<t+1}\right], \pi\left(\cdot \mid \left[x, y^{<t+1}\right]\right)\right)\right.  \right.       \\
         & \left.\left. +\operatorname{\Phi}^{\mu} \left(\cdots \operatorname{\Phi}^{\mu}\left(R\left(\left[x, y^{<T}\right], \pi\left(\cdot \mid \left[x, y^{<T}\right]\right)\right)\right)\right)\right) \right\},
    \end{aligned}
\end{equation}
\begin{equation}
    \begin{aligned}
        V_{\pi}^{\ast}\left(\left[x, y^{<t}\right]\right)
         & =\max \left\{R\left(\left[x, y^{<t}\right], \pi\left(\cdot \mid \left[x, y^{<t}\right]\right)\right) + \operatorname{\Phi}^{\mu}\left(R\left(\left[x, y^{<t+1}\right], \pi\left(\cdot \mid \left[x, y^{<t+1}\right]\right)\right)\right. \right. \\
         & \left. \left. +\operatorname{\Phi}^{\mu} \left(\cdots \operatorname{\Phi}^{\mu} \left(R\left(\left[x, y^{<T}\right], \pi\left(\cdot \mid \left[x, y^{<T}\right]\right)\right)\right)\right)\right)\right\}.
    \end{aligned}
\end{equation}

\section{Supplementary Materials for Section \ref{Section: Methodology}}   \label{Appendix: Methodology}
\subsection{The Proof of Lemma~\ref{Lemma: Equivalent MDP}}    \label{Appendix: the proof of Equivalent MDP lemma}
\textbf{Lemma \ref{Lemma: Equivalent MDP} Restated.}    
For a given Pb-MDP, the cumulative reward over the entire prompt-response can be decomposed as $r = \sum_{t=1}^T \gamma^{t-1} R\left(\left[x, y^{<t}\right], y^t\right)$,
the relationship between the state value function Equation~(\ref{Equation: Nested PbRL MDP}) and Equation~(\ref{Equation: New PbRL MDP}) is as follows:
$\tilde{V}_{\pi}\left(\left[x, y^{<t}\right]\right) = V_{\pi}\left(\left[x, y^{<t}\right]\right) + R_{1: t-1},$
where $R_{1: t-1} = \sum_{h=1}^{t-1} \gamma^{h-1} R\left(\left[x, y^{<h}\right], y^h\right)$ denotes the reward of the $1 \sim t-1$ steps of the prompt-response, and $V_{\pi}[x]$ and $\tilde{V}_{\pi}[x]$ are equivalent.

\begin{proof}
    First, according to \cite{2003_givan_equivalence-MDP, 2010_lowd-tree-like-MDP,2024_zhao_ra-pbrl}, we can reformulate the Pb-MDP as a decision tree-like MDP:

    (1) The state transition graph of the Pb-MDP is connected and acyclic;

    (2) Each state in the Pb-MDP corresponds to a unique node in the tree;

    (3) There is a single root node from which every other node is reachable via a unique path;

    (4) The transition probabilities between states follow the Markov property, i.e., the probability of transitioning to any future state depends only on the current state and not on the sequence of events that preceded it.

    Formally, let $S$ be the set of states and $p_{i j}$ be the transition probabilities between states $\mathbf{s}_i$ and $\mathbf{s}_j$.
    For an Pb-MDP with a tree-like structure, the probabilistic transition matrix $P$ is defined such that:
    \begin{equation}
        p_{i j}>0 \text { if there is an edge between } \mathbf{s}_i \text { and } \mathbf{s}_j \text { in the tree, and } p_{i j}=0 \text { otherwise. }
    \end{equation}
    Moreover, for each non-root node $\mathbf{s}_j$, there exists exactly one $\mathbf{s}_i$ such that $p_{i j}>0$, and $\mathbf{s}_i$ is the unique parent of $\mathbf{s}_j$ in the tree structure.

    To differentiate the two value functions, we denote the value from Equation (\ref{Equation: New PbRL MDP}) as $\tilde{V}_\pi\left(\left[x, y^{<t}\right]\right)$ and the value from Equation (\ref{Equation: Nested PbRL MDP}) as $V_\pi\left(\left[x, y^{<t}\right]\right)$. 
    Since the reward of the entire prompt-response can be decomposed as $r = \sum_{t=1}^T \gamma^{t-1} R\left(\left[x, y^{<t}\right], y^t\right)$, we have the following relationship:
    \begin{equation*}
        \tilde{V}_{\pi}\left(\left[x, y^{<t}\right]\right) = V_{\pi}\left(\left[x, y^{<t}\right]\right) + R_{1: t-1},
    \end{equation*}
    where $R_{1: t-1} = \sum_{h=1}^{t-1} \gamma^{h-1} R\left(\left[x, y^{<h}\right], y^h\right)$ denotes the reward of the $1 \sim t-1$ steps of a prompt-response.
    We prove this relationship by mathematical induction as follows.

    \textbf{Initial Case.}
    Using the tree-like Pb-MDP and the initial conditions of the Bellman equation, at the final step $t=T$, we have
    \begin{equation}
        \begin{aligned}
            \tilde{V}_{\pi}\left(\left[x, y^{<T}\right]\right)
             & =V_{\pi}\left(\left[x, y^{<T}\right], \pi\left(\cdot \mid \left[x, y^{<t}\right]\right)\right) + R_{1: T-1} \\
             & =V_{\pi}\left(\left[x, y^{<T}\right]\right) + R_{1: T-1}.
        \end{aligned}
    \end{equation}

    \textbf{Induction Step.}
    We now prove that if $\tilde{V}_{\pi}\left(\left[x, y^{<t+1}\right]\right) = V_{\pi}\left(\left[x, y^{<t+1}\right]\right) + R_{1: t}$ holds, then $\tilde{V}_{\pi}\left(\left[x, y^{<t}\right]\right) = V_{\pi}\left(\left[x, y^{<t}\right]\right) + R_{1: t-1}$ also holds.
    Since this policy $\pi$ on tree-like Pb-MDP is fixed, it has only one path to arrive $t\text{-}th$ state $\left(s_t = \left[x, y^{<t}\right]\right)$, denoted as:
    \begin{equation*}
        \Xi_t\left(s_{T, 1}\right)=\Xi_h\left(s_{T, 2}\right) \quad \forall \, s_{T, 1}, s_{T, 2} \in\left\{s_T \mid S_t\left(s_T\right) = \left[x, y^{<t}\right]\right\}.
    \end{equation*}

    Therefore, $R_{1: t-1}$ is unique.
    \begin{equation}
        \begin{aligned}
            \tilde{V}_{\pi} \left(\left[x, y^{<t}\right]\right)
             & =\operatorname{\Phi}^{\mu}\left(V_{\pi}\left(\left[x, y^{<t+1}\right]\right) + R_{1: t}\right),                                                                                             \\ % \quad \left[x, y^{<t+1}\right] \sim \mathbf{P}\left(\left[x, y^{<t}\right], y^t\right)                                                                                               \\
             & =\operatorname{\Phi}^{\mu}\left(V_{\pi}\left(\left[x, y^{<t+1}\right]\right) + R\left(\left[x, y^{<t}\right], \pi\left(\cdot \mid \left[x, y^{<t}\right]\right)\right) + R_{1: t-1}\right), \\ % \quad \left[x, y^{<t+1}\right] \sim \mathbf{P}\left(\left[x, y^{<t}\right], y^t\right) \\
             & =\operatorname{\Phi}^{\mu}\left(V_{\pi}\left(\left[x, y^{<t+1}\right]\right) + R\left(\left[x, y^{<t}\right], \pi\left(\cdot \mid \left[x, y^{<t}\right]\right)\right)\right) + R_{1: t-1}, \\ % \quad \left[x, y^{<t+1}\right] \sim \mathbf{P}\left(\left[x, y^{<t}\right], y^t\right) \\
             & =V_{\pi}\left(\left[x, y^{<t+1}\right]\right) + R_{1: t-1},
        \end{aligned}
    \end{equation}
    where the third equality holds because the risk measure function $\operatorname{\Phi}$ satisfies translation invariance.
    Then, by applying this conclusion, we observe that when $t=1, \tilde{V}_\pi[x] = V_\pi[x]$ holds.
    Thus, we have proven that for the Pb-MDP, the reward over the entire prompt-response can be decomposed as $r = \sum_{t=1}^T \gamma^{t-1} R\left(\left[x, y^{<t}\right], y^t\right)$, and $V_{\pi}[x]$ in Equation (\ref{Equation: Nested PbRL MDP}) and $\tilde{V}_{\pi}[x]$ in Equation (\ref{Equation: New PbRL MDP}) are equivalent.
\end{proof}

\subsection{The Derivation of Definition \ref{Definition: RA-advantage}}    \label{Appendix: the derivation of risk-aware advantage function definition}
\textbf{Definition \ref{Definition: RA-advantage} Restated.}
For a risk-sensitive Pb-MDP that satisfies the Bellman equation in Equation (\ref{Equation: New PbRL MDP}), the risk-aware advantage function can be defined as
\begin{equation*}
    \tilde{A}_{\pi} \left(\left[x, y^{<t}\right], z\right) = \tilde{Q}_\pi\left(\left[x, y^{<t}\right], z\right) - \operatorname{\Phi}^{\mu}(\tilde{V}_\pi\left(\left[x, y^{<t}\right]\right)),
\end{equation*}
where $z \sim \pi_{\theta}\left(\cdot \mid \left[x, y^{<t}\right]\right)$.

\textbf{The Derivation.}
In terms of designing the objective function at the token level, TDPO \cite{2024_zeng_tdpo} provides us with a valuable insight by introducing the advantage function from the TRPO algorithm in RL field as the target for each step.
Building upon TDPO, we consider the risk associated with language generation at each step and devise a novel risk-sensitive advantage function.
First, based on assumption that $r = \sum_{t=1}^T \gamma^{t-1} R\left(\left[x, y^{<t}\right], y^t\right)$, we can get:
\begin{equation}
    \begin{aligned}
        r & =\sum_{t=1}^T \gamma^{t-1} R\left(\left[x, y^{<t}\right], y^t\right)                                                                                                                                                                                                                                                                                                                                                                                              \\
          & =\sum_{t=1}^T \gamma^{t-1} \left(R\left(\left[x, y^{<t}\right], y^t\right)  + \gamma\operatorname{\Phi}^{\mu}\left(\tilde{V}_{\pi}\left(\left[x, y^{<t+1}\right]\right)\right) - \gamma\operatorname{\Phi}^{\mu} \left(\tilde{V}_{\pi}\left(\left[x, y^{<t+1}\right]\right)\right)\right)    \\                                                                                                                                                                     \\
          & =\operatorname{\Phi}^{\mu} \left(\tilde{V}_{\pi}\left([x]\right)\right) + \sum_{t=1}^T \gamma^{t-1} \left(R\left(\left[x, y^{<t}\right], y^t\right) + \gamma \operatorname{\Phi}^{\mu} \left(\tilde{V}_{\pi}\left(\left[x, y^{<t+1}\right]\right)\right) \right.   \\
          & \qquad \left. -\operatorname{\Phi}^{\mu} \left(\tilde{V}_{\pi}\left(\left[x, y^{<t}\right]\right)\right)\right) - \gamma^T \operatorname{\Phi}^{\mu}\left(\tilde{V}_{\pi}\left(\left[x, y^{<T+1}\right]\right)\right)   \\
          & =\operatorname{\Phi}^{\mu} \left(\tilde{V}_{\pi}\left([x]\right)\right) + \sum_{t=1}^T \gamma^{t-1} \left(\tilde{Q}_{\pi}\left(\left[x, y^{<t}\right], y^{t}\right) - \operatorname{\Phi}^{\mu} \left(\tilde{V}_{\pi}\left(\left[x, y^{<t}\right]\right)\right)\right) - \gamma^T \operatorname{\Phi}^{\mu} \left(\tilde{V}_{\pi}\left(\left[x, y^{<T+1}\right]\right)\right).
    \end{aligned}
\end{equation}
Next, note that $y^T = \text{EOS}$ denotes the end of the text sequence.
Therefore,
\begin{equation}
    V_\pi\left(\left[x, y^{<T+1}\right]\right) = \mathbb{E}_\pi\left[\sum_{k=0}^{\infty} \gamma^k R\left(\left[x, y^{<T+1+k}\right], y^{T+1+k}\right) \mid s_t = \left[x, y^{<T+1}\right]\right] = 0.
\end{equation}

Furthermore, we have
\begin{equation}
    r = \operatorname{\Phi}^{\mu}\left(\tilde{V}_{\pi}\left([x]\right)\right) + \sum_{t=1}^T \gamma^{t-1} \left(\tilde{Q}_{\pi}\left(\left[x, y^{<t}\right], y^{t}\right) - \operatorname{\Phi}^{\mu}\left(\tilde{V}_{\pi}\left(\left[x, y^{<t}\right]\right)\right)\right).
\end{equation}
So, we definite the risk-aware advantage function as $\tilde{A}_{\pi}\left(\left[x, y^{<t}\right], z\right) = \tilde{Q}_{\pi}\left(\left[x, y^{<t}\right], z\right)-\operatorname{\Phi}^{\mu}\left(\tilde{V}_{\pi}\left(\left[x, y^{<t}\right]\right)\right)$, where $z \sim \pi_{\theta}\left(\cdot \mid \left[x, y^{<t}\right]\right)$.

\subsection{The Proof of Lemma \ref{Lemma: policy improvement}}    \label{Appendix: the proof of the policy improvement lemma}
\textbf{Lemma \ref{Lemma: policy improvement} Restated.}
Given two policies $\pi$ and $\pi^{\prime}$, if for any state $s_t = \left[x, y^{<t}\right], \mathbb{E}_{z \sim \pi^{\prime}}\left[\tilde{A}_{\pi}\left(\left[x, y^{<t}\right], z\right)\right] \geq 0$ holds, then we can conclude:
\begin{equation*}
    \mathbb{E}_{x \sim \mathcal{D}}\left[\tilde{V}_{\pi^{\prime}}([x])\right] \geq \mathbb{E}_{x \sim \mathcal{D}}\left[\tilde{V}_{\pi}([x])\right].
\end{equation*}

\begin{proof}
    Let $\tau := (x, y^1, y^2, \ldots)$ denote a trajectory, where the expectation $\mathbb{E}_{\tau \mid \pi^{\prime}}[\cdot]$ is taken over trajectories generated by policy $\pi^{\prime}$. We then have
    \begin{equation}
        \begin{aligned}
              & \mathbb{E}_{x \sim \mathcal{D}}\left[\tilde{V}_{\pi^{\prime}}([x])\right] - \mathbb{E}_{x \sim \mathcal{D}}\left[\tilde{V}_{\pi}([x])\right]                                                                                                                                                                                                \\
            = & \mathbb{E}_{\tau \mid \pi^{\prime}} \left[\sum_{t=1}^{\infty} \gamma^{t-1} \left(R\left(\left[x, y^{<t}\right], y^t\right) + \gamma \operatorname{\Phi}^{\mu} \left(\tilde{V}_{\pi} \left(\left[x, y^{<t+1}\right]\right)\right)\right) - \tilde{V}_{\pi}([x]) \right]                                                                      \\
            = & \mathbb{E}_{\tau \mid \pi^{\prime}} \left[\sum_{t=1}^{\infty} \gamma^{t-1} \left(R\left(\left[x, y^{<t}\right], y^t\right) + \gamma \operatorname{\Phi}^{\mu} \left(\tilde{V}_{\pi} \left(\left[x, y^{<t+1}\right]\right)\right) - \operatorname{\Phi}^{\mu} \left(\tilde{V}_{\pi} \left(\left[x, y^{<t}\right]\right)\right)\right)\right] \\
            = & \mathbb{E}_{\tau \mid \pi^{\prime}} \left[\sum_{t=1}^{\infty} \gamma^{t-1} \left(\tilde{A}_{\pi} \left(\left[x, y^{<t}\right], y^t\right)\right)\right]                                                                                                                                                                                     \\
            = & \mathbb{E}_{\tau \mid \pi^{\prime}} \left[\sum_{t=1}^{\infty} \gamma^{t-1} \left(\mathbb{E}_{y^t \sim \pi^{\prime}}\left[\tilde{A}_{\pi}\left(\left[x, y^{<t}\right], y^t\right)\right)\right]\right].
        \end{aligned}
    \end{equation}

    Since for any state $s_t=\left[x, y^{<t}\right], \mathbb{E}_{z \sim \pi^{\prime}}\left[\tilde{A}_{\pi} \left(\left[x, y^{<t}\right], z\right)\right] \geq 0$, so we can obtain
    \begin{equation*}
        \mathbb{E}_{x \sim \mathcal{D}}\left[\tilde{V}_{\pi^{\prime}} ([x])\right]-\mathbb{E}_{x \sim \mathcal{D}}\left[\tilde{V}_{\pi} ([x])\right] \geq 0.
    \end{equation*}
    This completes the proof of Lemma~\ref{Lemma: policy improvement}.
\end{proof}

\subsection{The Proof of Lemma \ref{Lemma: the mapping between the risk-aware state-action function and the optimal policy}}    \label{Appendix: the mapping between the risk-aware state-action function and the optimal policy lemma}
\textbf{Lemma \ref{Lemma: the mapping between the risk-aware state-action function and the optimal policy} Restated.}
The constrained problem in Equation (\ref{Equation: Ra-RL objective function}) has the closed-form solution:
\begin{equation*}
    \pi_{\theta}^{\ast} \left(z \mid \left[x, y^{<t}\right]\right) = \frac{\pi_{\mathrm{ref}}\left(z \mid\left[x, y^{<t}\right]\right) \exp \left(\frac{1}{\beta} \tilde{Q}_{\pi_{\mathrm{ref}}} \left(\left[x, y^{<t}\right], z\right)\right)}{Z\left(\left[x, y^{<t}\right]; \beta\right)},
\end{equation*}
where $Z\left(\left[x, y^{<t}\right]; \beta\right) = \mathbb{E}_{z \sim \pi_{\mathrm{ref}}(\cdot \mid[x, y^{<t}])} e^{\frac{1}{\beta} \tilde{Q}_{\pi_{\mathrm{ref}}} \left(\left[x, y^{<t}\right], z\right)}$ is the partition function.

\begin{proof}
    \begin{equation}
        \begin{aligned}
              & \max_{\pi_{\theta}} \mathbb{E}_{z \sim \pi_{\theta}\left(\cdot \mid \left[x, y^{<t}\right]\right)} \tilde{A}_{\pi_{\mathrm{ref}}} \left(\left[x, y^{<t}\right], z\right) - \beta D_{\mathrm{KL}}\left(\pi_{\theta}\left(\cdot \mid \left[x, y^{<t}\right]\right) \| \pi_{\mathrm{ref}}\left(\cdot \mid \left[x, y^{<t}\right]\right)\right)                                                                                   \\
            = & \max_{\pi_{\theta}} \mathbb{E}_{z \sim \pi_{\theta}\left(\cdot \mid \left[x, y^{<t}\right]\right)} \left(\left(\tilde{Q}_{\pi_{\mathrm{ref}}} \left(\left[x, y^{<t}\right], z\right) - \tilde{V}_{\pi_{\mathrm{ref}}} \left(\left[x, y^{<t}\right]\right)\right) + \beta \log \left(\frac{\pi_{\mathrm{ref}}\left(z \mid \left[x, y^{<t}\right]\right)}{\pi_{\theta}\left(z \mid \left[x, y^{<t}\right]\right)}\right)\right) \\
            = & \max_{\pi_{\theta}} \beta \mathbb{E}_{z \sim \pi_{\theta}\left(\cdot \mid \left[x, y^{<t}\right]\right)} \log \left(\frac{\pi_{\mathrm{ref}}\left(z \mid \left[x, y^{<t}\right]\right) e^{\frac{1}{\beta} \tilde{Q}_{\pi_{\mathrm{ref}}} \left(\left[x, y^{<t}\right], z\right)}}{\pi_{\theta}\left(z \mid \left[x, y^{<t}\right]\right)}\right) - \tilde{V}_{\pi_{\mathrm{ref}}} \left(\left[x, y^{<t}\right]\right)         \\
            = & \max_{\pi_{\theta}} \beta \mathbb{E}_{z \sim \pi_{\theta}\left(\cdot \mid \left[x, y^{<t}\right]\right)} \log \left(\frac{\pi_{\mathrm{ref}}\left(z \mid \left[x, y^{<t}\right]\right) e^{\frac{1}{\beta} \tilde{Q}_{\pi_{\mathrm{ref}}} \left(\left[x, y^{<t}\right], z\right)}}{Z\left(\left[x, y^{<t}\right]; \beta\right) \pi_{\theta}\left(z \mid \left[x, y^{<t}\right]\right)}\right)                                  \\
              & - \tilde{V}_{\pi_{\mathrm{ref}}} \left(\left[x, y^{<t}\right]\right) + \beta \log Z\left(\left[x, y^{<t}\right]; \beta\right)                                                                                                                                                                                                                                                                                                 \\
            = & \max_{\pi_{\theta}} -\beta D_{\mathrm{KL}}\left(\pi_{\theta}\left(z \mid \left[x, y^{<t}\right]\right) \| \frac{\pi_{\mathrm{ref}}\left(z \mid \left[x, y^{<t}\right]\right) e^{\frac{1}{\beta} \tilde{Q}_{\pi_{\mathrm{ref}}} \left(\left[x, y^{<t}\right], z\right)}}{Z\left(\left[x, y^{<t}\right]; \beta\right)}\right)                                                                                                   \\
              & - \tilde{V}_{\pi_{\mathrm{ref}}} \left(\left[x, y^{<t}\right]\right) + \beta \log Z\left(\left[x, y^{<t}\right]; \beta\right),
        \end{aligned}
    \end{equation}
    where $Z\left(\left[x, y^{<t}\right]; \beta\right)$ is the partition function:
    \begin{equation}
        Z\left(\left[x, y^{<t}\right]; \beta\right)=\mathbb{E}_{z \sim \pi_{\mathrm{ref}}\left(\cdot \mid\left[x, y^{<t}\right]\right)} \exp \left(\frac{1}{\beta} \tilde{Q}_{\pi_{\mathrm{ref}}} \left(\left[x, y^{<t}\right], z\right)\right).
    \end{equation}

    Then, we can derive the relationship between the optimal policy and the state-action function:
    \begin{equation*}
        \pi_{\theta}^{\ast} \left(z \mid \left[x, y^{<t}\right]\right)=\frac{\pi_{\mathrm{ref}}\left(z \mid\left[x, y^{<t}\right]\right) \exp \left(\frac{1}{\beta} \tilde{Q}_{\pi_{\mathrm{ref}}} \left(\left[x, y^{<t}\right], z\right)\right)}{Z\left(\left[x, y^{<t}\right]; \beta\right)}.
    \end{equation*}
    This completes the proof of Lemma~\ref{Lemma: the mapping between the risk-aware state-action function and the optimal policy}.
\end{proof}

\subsection{The Proof of Lemma \ref{Lemma: the equivalence between the BT model and the Regret Preference Model}}    \label{Appendix: the proof of the equivalence between the BT model and the Regret Preference Model lemma}
\textbf{Lemma \ref{Lemma: the equivalence between the BT model and the Regret Preference Model} Restated.}
Given a reward function $r(x, y)$ over the entire prompt-response, based on the relationship between token-wise rewards and the reward function $r(x, y) = \sum_{t=1}^T \gamma^{t-1} R\left(\left[x, y^{<t}\right], y^t\right)$, we can establish the equivalence between the Bradley-Terry model and the Regret Preference Model, i.e.,
\begin{equation*}   
    P_{\mathrm{BT}}\left(y_1 \succ y_2 \mid x\right) = \sigma\left(\sum_{t=1}^{T_1} \gamma^{t-1} \tilde{A}_\pi\left(\left[x, y_1^{<t}\right], y_1^t\right) -\sum_{t=1}^{T_2} \gamma^{t-1} \tilde{A}_\pi\left(\left[x, y_2^{<t}\right], y_2^t\right)\right),
\end{equation*}
where $\sigma(z) = 1 /\left(1+\exp (-z)\right)$ is the logistic sigmoid function for any random variable $z$.

\begin{proof}
    Recalling to the BT model in Equation (\ref{Equation: BT model})
    \begin{equation}
        P_{\mathrm{BT}}\left(y_1 \succ y_2 \mid x\right) = \frac{\exp\left(r \left(x, y_1\right)\right)}{\exp\left(r \left(x, y_1\right)\right) + \exp\left(r \left(x, y_2\right)\right)},
    \end{equation}
    and the equivalence between prompt-response reward and the risk-aware advantage function:
    \begin{equation*}
        \begin{aligned}
            r = & \operatorname{\Phi}^{\mu}\left(\tilde{V}_{\pi}\left([x]\right)\right) + \sum_{t=1}^T \gamma^{t-1} \left(\tilde{Q}_{\pi}\left(\left[x, y^{<t}\right], y^{t}\right) - \operatorname{\Phi}^{\mu} \left(\tilde{V}_{\pi}\left(\left[x, y^{<t}\right]\right)\right)\right) \\
            =   & \operatorname{\Phi}^{\mu}\left(\tilde{V}_{\pi}\left([x]\right)\right) + \sum_{t=1}^T \gamma^{t-1} \tilde{A}_{\pi} \left(\left[x, y^{<t}\right], y^t\right).
        \end{aligned}
    \end{equation*}
    Then, we have
    \begin{equation*}
        P_{\mathrm{BT}}\left(y_1 \succ y_2 \mid x\right) = \sigma\left(\sum_{t=1}^{T_1} \gamma^{t-1} \tilde{A}_{\pi} \left(\left[x, y_1^{<t}\right], y_1^t\right) -\sum_{t=1}^{T_2} \gamma^{t-1} \tilde{A}_{\pi}\left(\left[x, y_2^{<t}\right], y_2^t\right)\right).
    \end{equation*}
    This completes the proof of Lemma \ref{Lemma: the equivalence between the BT model and the Regret Preference Model}. 
\end{proof}

\subsection{The Proof of Theorem \ref{Theorem: the Bradley Terry model express the human preference probability}}    \label{Appendix: the proof of the Bradley Terry model express the human preference probability theorem}
\textbf{Theorem \ref{Theorem: the Bradley Terry model express the human preference probability} Restated.}
Given prompts $x$ and pairwise responses $\left(y_1, y_2\right)$, and the risk-aware objective function in Equation (\ref{Equation: Ra-RL objective function}), the Bradley-Terry model expresses the human preference probability in terms of the risk-aware optimal policy $\pi_\theta^*$ and reference policy $\pi_{\mathrm{ref}}$:
\begin{equation*}
    P_{\mathrm{BT}}^{\ast} \left(y_1 \succ y_2 \mid x\right)=\sigma\left(u^{\ast} \left(x, y_1, y_2\right) - \delta^{\ast} \left(x, y_1, y_2\right)\right),
\end{equation*}
where $u\left(x, y_1, y_2\right)$ represents the difference in implicit rewards defined by the risk-aware policy $\pi_{\theta}^{\ast}$ and the reference policy $\pi_{\mathrm{ref}}$, weighted by $\beta$, represented as
\begin{equation*}
    u\left(x, y_1, y_2\right) = \beta\log\frac{\pi_{\theta}\left(y_1 \mid x\right)}{\pi_{\mathrm{ref}}\left(y_1 \mid x\right)} - \beta\log\frac{\pi_{\theta}\left(y_2 \mid x\right)}{\pi_{\mathrm{ref}}\left(y_2 \mid x\right)},
\end{equation*}
and $\delta\left(x, y_1, y_2\right)$ represents the difference in sequential risk ratio between two pairs $\left(x, y_1\right)$ and $\left(x, y_2\right)$, expressed as
\begin{equation*}
    \delta\left(x, y_1, y_2\right)=\beta D_{\mathrm{SeqRR}} \left(x, y_2; \pi_{\mathrm{ref}} \mid \pi_{\theta}\right) -\beta D_{\mathrm{SeqRR}}\left(x, y_1 ; \pi_{\mathrm{ref}} \mid \pi_{\theta}\right).
\end{equation*}

\begin{proof}
    According to the Lemma \ref{Lemma: the mapping between the risk-aware state-action function and the optimal policy}, we have
    \begin{equation}   \label{Equation(appendix): the mapping between the risk-aware state-action function and the optimal policy}
        \pi_\theta^{\ast}\left(z \mid\left[x, y^{<t}\right]\right)=\frac{\pi_{\mathrm{ref}}\left(z \mid\left[x, y^{<t}\right]\right) \exp \left(\frac{1}{\beta} \tilde{Q}_{\pi_{\mathrm{ref}}}\left(\left[x, y^{<t}\right], z\right)\right)}{Z\left(\left[x, y^{<t}\right] ; \beta\right)},
    \end{equation}
    where $Z\left(\left[x, y^{<t}\right] ; \beta\right)=\mathbb{E}_{z \sim \pi_{\mathrm{ref}}\left(\cdot \mid\left[x, y^{<t}\right]\right)} e^{\frac{1}{\beta} \tilde{Q}_{\pi_{\mathrm{ref}}}\left(\left[x, y^{<t}\right], z\right)}$ is the partition function.
    Rearrange Equation (\ref{Equation(appendix): the mapping between the risk-aware state-action function and the optimal policy}), we obtain
    \begin{equation}   \label{Equation(appendix): the rearrange the mapping between the risk-aware state-action function and the optimal policy}
        \tilde{Q}_{\pi_{\mathrm{ref}}}\left(\left[x, y^{<t}\right], z\right)=\beta \log \frac{\pi_\theta^*\left(z \mid\left[x, y^{<t}\right]\right)}{\pi_{\mathrm{ref}}\left(z \mid\left[x, y^{<t}\right]\right)}+\beta \log Z\left(\left[x, y^{<t}\right]; \beta\right).
    \end{equation}

    From Lemma \ref{Lemma: the equivalence between the BT model and the Regret Preference Model}, we can get
    \begin{equation}  \label{Equation(appendix): BT model}
        P_{\mathrm{BT}}\left(y_1 \succ y_2 \mid x\right)=\sigma\left(\sum_{t=1}^{T_1}\left(\gamma^{t-1} \tilde{A}_\pi\left(\left[x, y_1^{<t}\right], y_1^t\right)\right) - \sum_{t=1}^{T_2}\left(\gamma^{t-1} \tilde{A}_\pi\left(\left[x, y_2^{<t}\right], y_2^t\right)\right)\right).
    \end{equation}

    By leveraging Equation (\ref{Equation(appendix): the rearrange the mapping between the risk-aware state-action function and the optimal policy}), we can derive
    \begin{equation}
        \begin{aligned}
              & \sum_{t=1}^T \gamma^{t-1}\tilde{A}_{\pi_{\mathrm{ref}}}\left(\left[x, y^{<t}\right], y^t\right)                                                                                                                                                                                                                                                                                                                                                                                             \\
            = & \sum_{t=1}^T \gamma^{t-1}\left(Q_{\pi_{\mathrm{ref}}}\left(\left[x, y^{<t}\right], y^t\right) - \operatorname{\Phi}^{\mu} \left(\tilde{V}_{\pi_{\mathrm{ref}}}\left(\left[x, y^{<t}\right]\right)\right)\right)                                                                                                                                                                                                                                                                             \\
            = & \sum_{t=1}^T \gamma^{t-1}\left(\tilde{Q}_{\pi_{\mathrm{ref}}}\left(\left[x, y^{<t}\right], y^t\right) - \operatorname{\Phi}^{\mu} \left(\tilde{Q}_{\pi_{\mathrm{ref}}}\left(\left[x, y^{<t}\right], z\right)\right)\right)                                                                                                                                                                                                                                                                  \\
            = & \sum_{t=1}^T \gamma^{t-1}\left(\beta \log \frac{\pi_{\theta}^{\ast} \left(y^t \mid \left[x, y^{<t}\right]\right)}{\pi^{\mathrm{ref}}\left(y^t \mid \left[x, y^{<t}\right]\right)} + \beta \log Z\left(\left[x, y^{<t}\right]; \beta\right) \right.\\
              & \qquad \left. -\operatorname{\Phi}^{\mu} \left(\beta \log \frac{\pi_\theta^*\left(z \mid \left[x, y^{<t}\right]\right)}{\pi^{\mathrm{ref}}\left(z \mid \left[x, y^{<t}\right]\right)} + \beta \log Z\left(\left[x, y^{<t}\right]; \beta\right)\right)\right).
        \end{aligned}
    \end{equation}

    Note that
    \begin{equation*}
        \mathbb{E}_{z \sim \pi_{\mathrm{ref}}}\left[\beta \log Z\left(\left[x, y^{<t}\right] ; \beta\right)\right]=\beta \log Z\left(\left[x, y^{<t}\right] ; \beta\right).
    \end{equation*}

    Therefore,
    \begin{equation}
        \begin{aligned}
              & \sum_{t=1}^T \gamma^{t-1} \tilde{A}_{\pi_{\mathrm{ref}}}\left(\left[x, y^{<t}\right], y^t\right)                                                                                                                                                                                                                                                                                                              \\
            = & \beta \sum_{t=1}^T \gamma^{t-1}\left(\log \frac{\pi_\theta^{\ast}\left(y^t \mid \left[x, y^{<t}\right]\right)}{\pi_{\mathrm{ref}}\left(y^t \mid \left[x, y^{<t}\right]\right)} - \operatorname{\Phi}^{\mu}_{z \sim \pi_{\mathrm{ref}}}\left(\log \frac{\pi_\theta^*\left(z \mid\left[x, y^{<t}\right]\right)}{\pi_{\mathrm{ref}}\left(z \mid\left[x, y^{<t}\right]\right)}\right)\right)                      \\
            = & \beta \sum_{t=1}^T \gamma^{t-1} \log \frac{\pi_\theta^{\ast}\left(y^t \mid \left[x, y^{<t}\right]\right)}{\pi_{\mathrm{ref}}\left(y^t \mid \left[x, y^{<t}\right]\right)} + \beta \sum_{t=1}^T \gamma^{t-1} \operatorname{\Phi}^{\mu}_{z \sim \pi_{\mathrm{ref}}}\left(\log \frac{\pi_\mathrm{ref}\left(z \mid\left[x, y^{<t}\right]\right)}{\pi_{\theta^*}\left(z \mid\left[x, y^{<t}\right]\right)}\right).
        \end{aligned}
    \end{equation}

    When substituting $\gamma=1$ into the expression, we obtain a more concise form:
    \begin{equation}      \label{Equation(appendix): a concise form of risk-aware advantage function}
        \begin{aligned}
            \sum_{t=1}^T \tilde{A}_{\pi_{\mathrm{ref}}}\left(\left[x, y^{<t}\right], y^t\right)
            = & \beta \sum_{t=1}^T \log \frac{\pi_\theta^\ast \left(y^t \mid \left[x, y^{<t}\right]\right)}{\pi_{\mathrm{ref}}\left(y^t \mid\left[x, y^{<t}\right]\right)} + \beta \sum_{t=1}^T \operatorname{\Phi}^{\mu}_{z \sim \pi_{\mathrm{ref}}}\left(\log \frac{\pi_\mathrm{ref}\left(z \mid\left[x, y^{<t}\right]\right)}{\pi_{\theta^*}\left(z \mid\left[x, y^{<t}\right]\right)}\right) \\
            = & \beta\left(\log \frac{\pi_\theta^\ast \left(y \mid x\right)}{\pi_{\mathrm{ref}}\left(y \mid x\right)} + D_{\mathrm{SeqRR}}\left(x, y; \pi_{\mathrm{ref}} \mid \pi_\theta^\ast \right)\right),
        \end{aligned}
    \end{equation}
    where
    $
        D_{\mathrm{SeqRR}}\left(x, y; \pi_{\mathrm{ref}} \mid \pi_{\theta}\right) = \sum_{t=1}^T \operatorname{\Phi}^{\mu}_{z \sim \pi_{\mathrm{ref}}} \left(\log \frac{\pi_\mathrm{ref}\left(z \mid\left[x, y^{<t}\right]\right)}{\pi_{\theta}\left(z \mid\left[x, y^{<t}\right]\right)}\right).
    $

    Then, we let
    \begin{equation}
        u\left(x, y_1, y_2\right)=\beta \log \frac{\pi_\theta\left(y_1 \mid x\right)}{\pi_{\mathrm{ref}}\left(y_1 \mid x\right)}-\beta \log \frac{\pi_\theta\left(y_2 \mid x\right)}{\pi_{\mathrm{ref}}\left(y_2 \mid x\right)},
    \end{equation}
    \begin{equation}    \label{Equation(appendix): D_SeqRR}
        \delta\left(x, y_1, y_2\right)=\beta D_{\mathrm{SeqRR}}\left(x, y_2 ; \pi_{\mathrm{ref}} \mid \pi_\theta\right)-\beta D_{\mathrm{SeqRR}}\left(x, y_1 ; \pi_{\mathrm{ref}} \mid \pi_\theta\right).
    \end{equation}

    Substituting Equation (\ref{Equation(appendix): a concise form of risk-aware advantage function}) into Equation (\ref{Equation(appendix): BT model}), we arrive at
    $$P_{\mathrm{BT}}^{\ast}\left(y_1 \succ y_2 \mid x\right)=\sigma\left(u^{\ast}\left(x, y_1, y_2\right)-\delta^{\ast}\left(x, y_1, y_2\right)\right).$$
    This completes the proof Theorem \ref{Theorem: the Bradley Terry model express the human preference probability}.  
\end{proof}

\subsection{Algorithm}    \label{Appendix: Algorithm}
In this subsection, we provide the main pseudocode for Risk-aware Direct Preference Optimization (Ra-DPO), as outlined in Algorithm \ref{Appendix: Ra-DPO algorithm}.
\begin{algorithm}[h!]
    \caption{Risk-aware Direct Preference Optimization (Ra-DPO)}
    \label{Appendix: Ra-DPO algorithm}
    \begin{algorithmic}
        \STATE {\bfseries Input:} Reference model $\pi_{\mathrm{ref}}$, Policy model $\pi_{\theta}$, Coefficient $\alpha$, $\beta$, Risk control parameter $\mu$, Learning rate $\eta$
        \STATE {\bfseries Input:} Dataset $\mathcal{D}=\left\{\left(x, y_w, y_l\right)^i\right\}_{i=1}^N$  of size $N$, Method $\mathcal{M}$
        \STATE {\bfseries Initialize:} $\pi_{\theta} \leftarrow \pi_{\mathrm{ref}}$
        \FOR {each epoch }
        \STATE Sample mini-batch $\mathcal{D}_m=\left\{\left(x, y_w, y_l\right)^m\right\}_{m=1}^M$ from $\mathcal{D}$
        \STATE Predict the probabilities $\pi_{\theta}\left(y_w \mid x\right)$ and $\pi_{\theta}\left(y_l \mid x\right)$ for $\left(x, y_w, y_l\right)$ in the mini-batch $\mathcal{D}_m$ using the policy model
        \STATE Predict the probabilities $\pi_{\mathrm{ref}}\left(y_w \mid x\right)$ and $\pi_{\mathrm{ref}}\left(y_l \mid x\right)$ for $\left(x, y_w, y_l\right)$ in the mini-batch  $\mathcal{D}_m$ using the reference model
        \STATE Calculate the function  $u\left(x, y_w, y_l\right)=\beta \log \frac{\pi_{\theta}\left(y_w \mid x\right)}{\pi_{\mathrm{ref}}\left(y_w \mid x\right)}-\beta \log \frac{\pi_{\theta}\left(y_l \mid x\right)}{\pi_{\mathrm{ref}}\left(y_l \mid x\right)}$
        \STATE Compute the sequential risk ratio $D_{\mathrm{SeqRR}}\left(x, y_w ; \pi_{\mathrm{ref}} \mid \pi_{\theta}\right)$ for $\left(x, y_w\right)$ in the mini-batch $\mathcal{D}_m$
        \STATE Compute the sequential risk ratio $D_{\mathrm{SeqRR}}\left(x, y_l ; \pi_{\mathrm{ref}} \mid \pi_{\theta}\right)$ for $\left(x, y_l\right)$ in the mini-batch $\mathcal{D}_m$
        \IF {Method $\mathcal{M}$  is $\text{Ra-DPO}_\text{1}$}
        \STATE Calculate $\delta\left(x, y_w, y_l\right)=\beta D_{\mathrm{SeqRR}}\left(x, y_l; \pi_{\mathrm{ref}} \mid \pi_{\theta}\right)-\beta D_{\mathrm{SeqRR}}\left(x, y_w ; \pi_{\mathrm{ref}} \mid \pi_{\theta}\right)$
        \STATE $\theta \leftarrow {\theta} + \eta \nabla_{\theta} \mathbb{E}_{\left(x, y_w, y_l\right) \sim \mathcal{D}_m}\left[\log \sigma\left(u\left(x, y_w, y_l\right)-\delta\left(x, y_w, y_l\right)\right)\right]$
        \ELSE [Method $\mathcal{M}$ is $\text{Ra-DPO}_\text{2}$]
        \STATE Calculate $\delta_2\left(x, y_w, y_l\right) = \beta D_{\mathrm{SeqRR}}\left(x, y_l; \pi_{\mathrm{ref}} \mid \pi_{\theta}\right)- \operatorname{sg}\left(\beta D_{\mathrm{SeqRR}}\left(x, y_w; \pi_{\mathrm{ref}} \mid \pi_{\theta}\right)\right)$
        \STATE $\theta \leftarrow {\theta} + \eta \nabla_{\theta} \mathbb{E}_{\left(x, y_w, y_l\right) \sim \mathcal{D}_m}\left[\log \sigma\left(u\left(x, y_w, y_l\right)-\alpha \delta_2\left(x, y_w, y_l\right)\right)\right]$
        \ENDIF
        \ENDFOR
    \end{algorithmic}
\end{algorithm}

\section{Supplementary Materials for Section \ref{Section: Experiments}}    \label{Appendix: Experiments}
\subsection{Experiments compute resources}     \label{Appendix: Experiments compute resources}
All reported results of our algorithm and baseline algorithms are trained using 4 $\times$ A100 GPUs, each with 40GB of memory.

\subsection{Assets}     \label{Appendix: Experiments assets}
We have compiled the datasets, models, and benchmark codes used in this paper and express our gratitude to all relevant sources.

\textbf{Dataset: }
\begin{itemize}[left=4pt]
    \item IMDb Dataset \cite{2011_maas_imdb}: {\small \url{https://huggingface.co/datasets/stanfordnlp/imdb}}
    \item Anthropic HH Dataset \cite{2022_bai_Anthropic-HH}: {\small \url{https://huggingface.co/datasets/Anthropic/hh-rlhf}}
    \item AlpacaEval \cite{2024_dubois_AlpacaEval}: {\small \url{https://huggingface.co/datasets/tatsu-lab/alpaca_eval}}
\end{itemize}

\textbf{Model: }
\begin{itemize}[left=4pt]
    \item GPT-2 Large \cite{2019_radford_gpt2}: {\small \url{https://huggingface.co/openai-community/gpt2-large}}
    \item Gpt2-large-imdb-fine-tuned: {\small \url{https://huggingface.co/insub/gpt2-large-IMDb-fine-tuned}}
    \item Pythia-1.4B \cite{2023_biderman_pythia}: {\small \url{https://huggingface.co/EleutherAI/pythia-1.4b}}
    \item Pythia-2.8B \cite{2023_biderman_pythia}: {\small \url{https://huggingface.co/EleutherAI/pythia-2.8b}} 
    \item Oasst-sft-4-pythia-12b-epoch-3.5: {\small \url{https://huggingface.co/OpenAssistant/oasst-sft-4-pythia-12b-epoch-3.5}
    }
\end{itemize}

\textbf{Code: }
\begin{itemize}[left=4pt]
    \item We trained Ra-DPO and the baseline models based on the original KTO implementation {\small \url{https://github.com/ContextualAI/HALOs}}, and our code can be found at {\small \url{https://github.com/zlj123-max/Ra-DPO}}.
\end{itemize}

\subsection{Experimental Details}     \label{Appendix: Experimental details}
In our experiments, we followed the original KTO implementation for the main parameter settings, and both Ra-DPO and the baseline models used the same hyperparameters, as detailed in Table \ref{Appendix: Table 1}-\ref{Appendix: Table 2}.

\begin{table}[htb]
    \caption{
        Hyperparameters in loss functions for different algorithms.
    }
    \label{Appendix: Table 1}
    \begin{center}
        \begin{small}
            % \begin{sc}
                \begin{tabular}{p{1.3cm} p{0.8cm} p{2.6cm} p{4cm} }
                    \toprule
                    Method                    & $\beta$       & $\alpha$                      & $\mu$                                 \\
                    \midrule
                    DPO                       & 0.1           & -                             & -                                                         \\
                    PPO                       & -             & -                             & -                                                         \\
                    KTO                       & 0.1           & -                             & -                                                         \\
                    $\text{TDPO}_\text{1}$    & 0.1           & -                             & -                                                         \\
                    $\text{TDPO}_\text{2}$    & 0.1           & $\{0.3, 0.5, 0.7, 0.9\}$      & -                                                         \\
                    $\text{Ra-DPO}_\text{1}$  & 0.1           & -                             & -                                                         \\
                    $\text{Ra-DPO}_\text{2}$  & 0.1           & $\{0.3, 0.5, 0.7, 0.9\}$      & CVaR: $\{0.99, 0.98, 0.97, 0.95\}$  ERM: $\{9, 7, 5\}$                \\
                    \bottomrule
                \end{tabular}
            % \end{sc}
        \end{small}
    \end{center}
\end{table}

\begin{table}[htb]
    \caption{
        Hyperparameters in network training.
    }
    \label{Appendix: Table 2}
    \begin{center}
        \begin{small}
            % \begin{sc}
                \begin{tabular}{lc}
                    \toprule
                    Parameter                       & value                \\
                    \midrule
                    max length                      & 512                  \\
                    max prompt length               & 256                  \\
                    gradient accumulation steps     & 4                    \\
                    learning rate                   & $5\times 10^{-6}$       \\
                    optimizer                       & AdamW                \\
                    \bottomrule
                \end{tabular}
            % \end{sc}
        \end{small}
    \end{center}
\end{table}

\subsection{Additional Experimental Results}     \label{Appendix: Additional experimental results}
\begin{figure}[htb]
    \vskip 0.1in
    \centering
    \subfigure{
        \includegraphics[width=0.99 \linewidth]{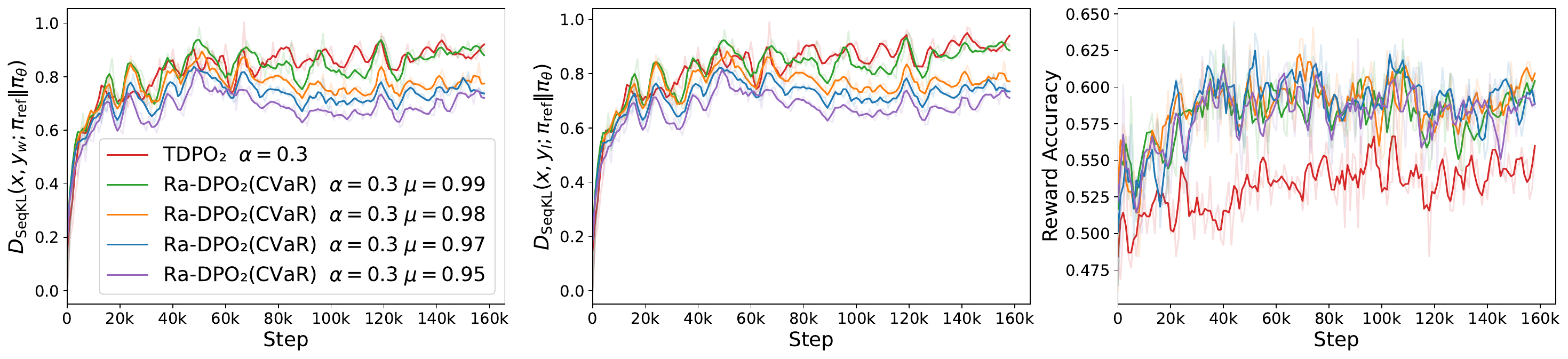}
    }
    \quad
    \subfigure{
        \includegraphics[width=0.99 \linewidth]{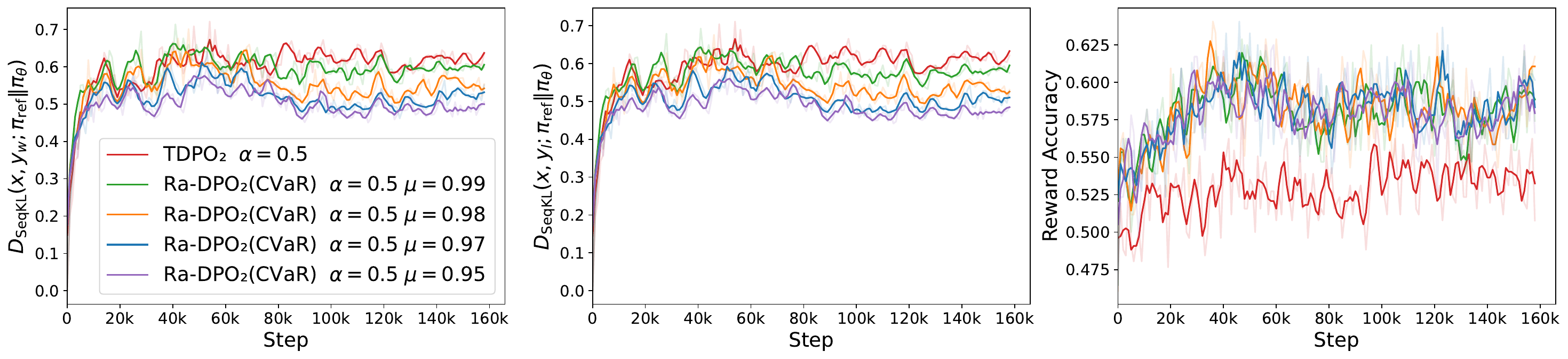}
    }
    \quad
    \subfigure{
        \includegraphics[width=0.99 \linewidth]{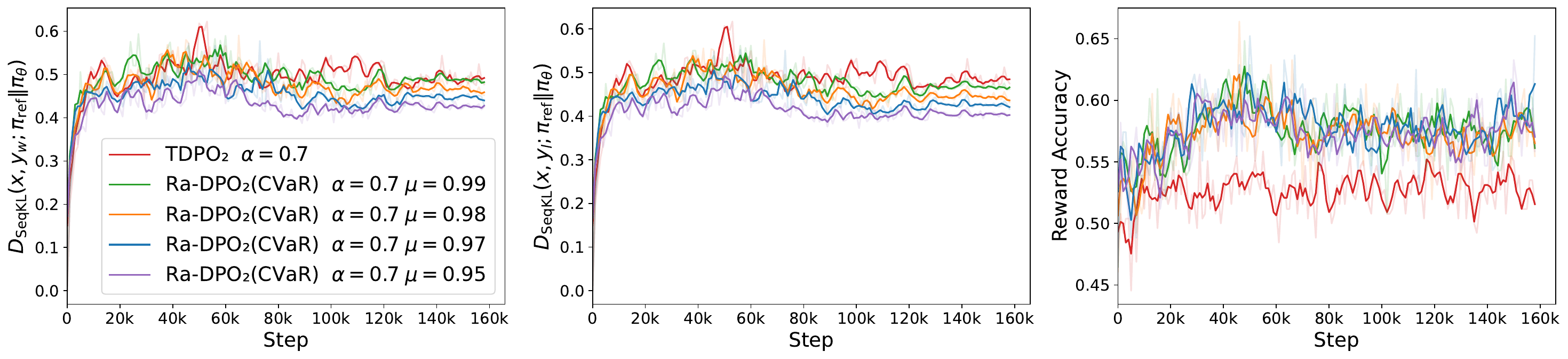}
    }
    \quad
    \subfigure{
        \includegraphics[width=0.99 \linewidth]{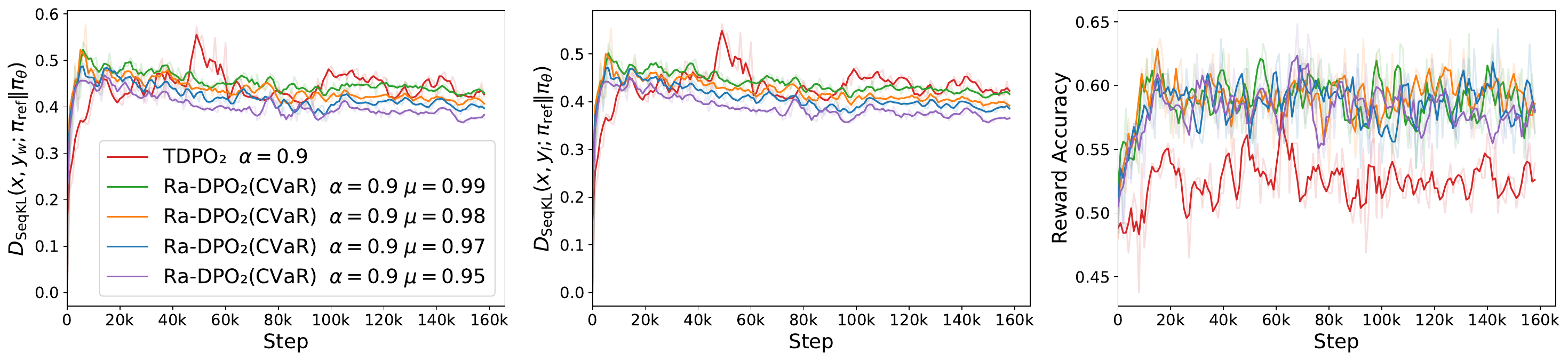}
    }
    \caption{
        The experiment on the Anthropic HH dataset with Pythia-1.4B serving as the base model.
        \textbf{Left} and \textbf{Middle} present the progression of sequential KL divergence (the lower the better) for both preferred and dispreferred responses.
        \textbf{Right} illustrates reward accuracy curves (the higher the better).
    }
    \label{Appendix: Fig.1}
    \vskip 0.1in
\end{figure}

\begin{figure}[h!]
    \vskip 0.1in
    \centering
    \subfigure{
        \includegraphics[width=0.315 \linewidth]{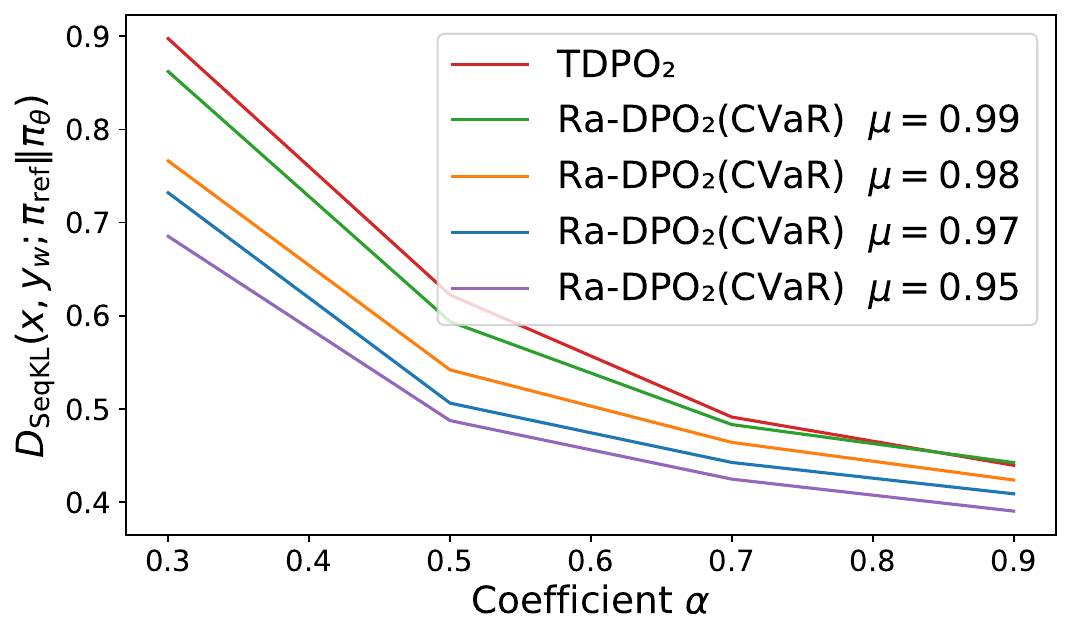}
    }
    \subfigure{
        \includegraphics[width=0.315 \linewidth]{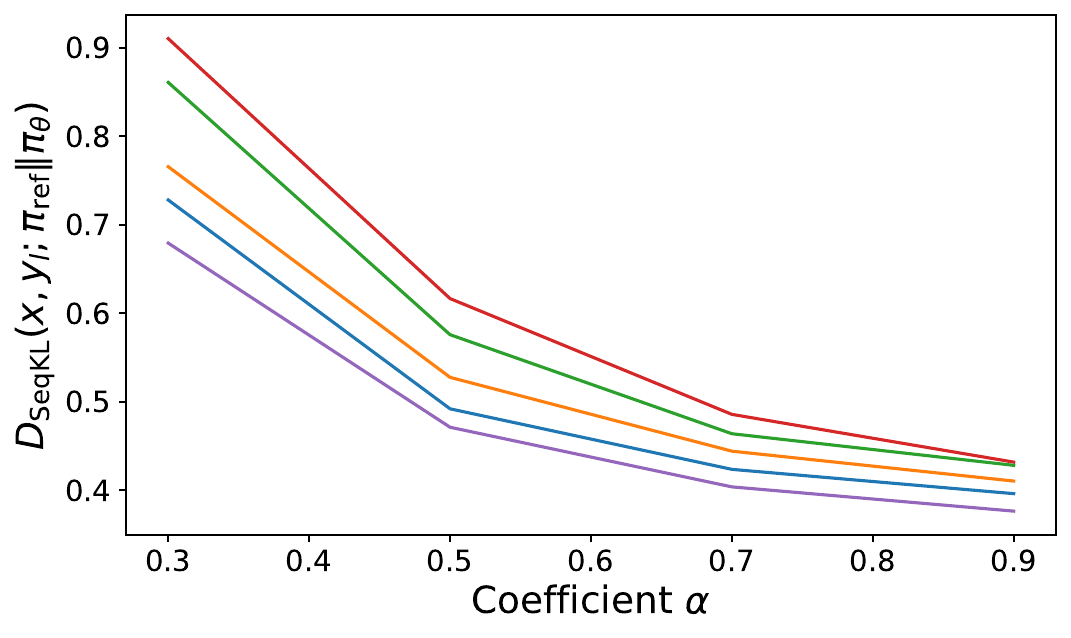}
    }
    \subfigure{
        \includegraphics[width=0.315 \linewidth]{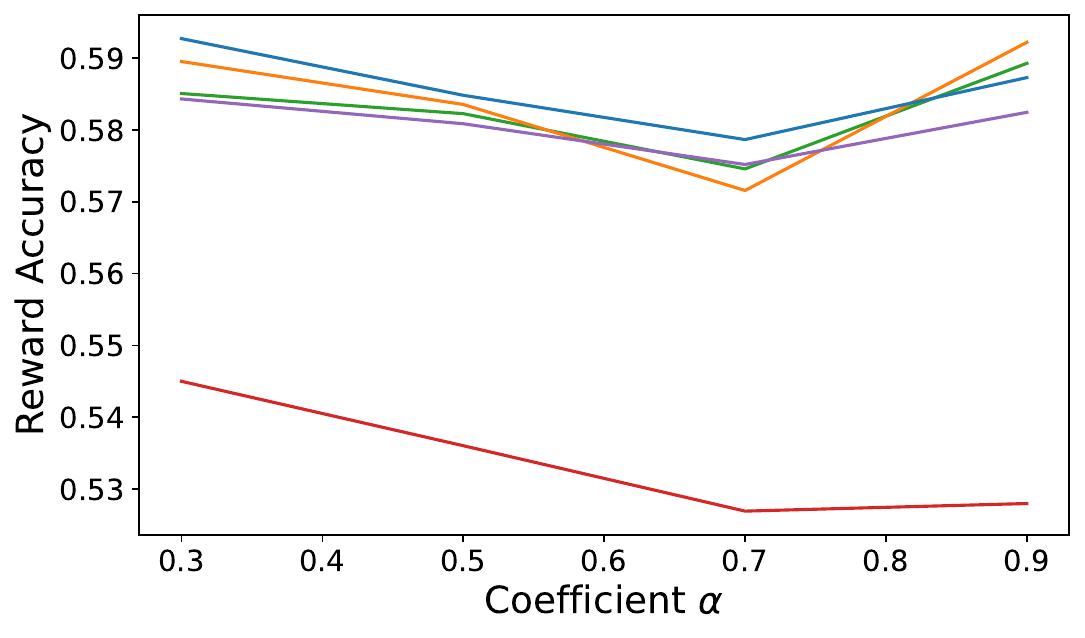}
    }
    \caption{
        The experiment on the Anthropic HH dataset with Pythia-1.4B serving as the base model.
        \textbf{Left} and \textbf{Middle} presents the sequential KL divergence (the lower the better) for preferred and dispreferred responses, while \textbf{Right} presents the reward accuracy curves (the higher the better) under $\alpha = \{0.3, 0.5, 0.7, 0.9\}$.
     }
    \label{Appendix: Fig.2}
    \vskip 0.1in
\end{figure}

\begin{figure}[htb]
    \vskip 0.1in
    \centering
    \subfigure{
        \includegraphics[width=0.99 \linewidth]{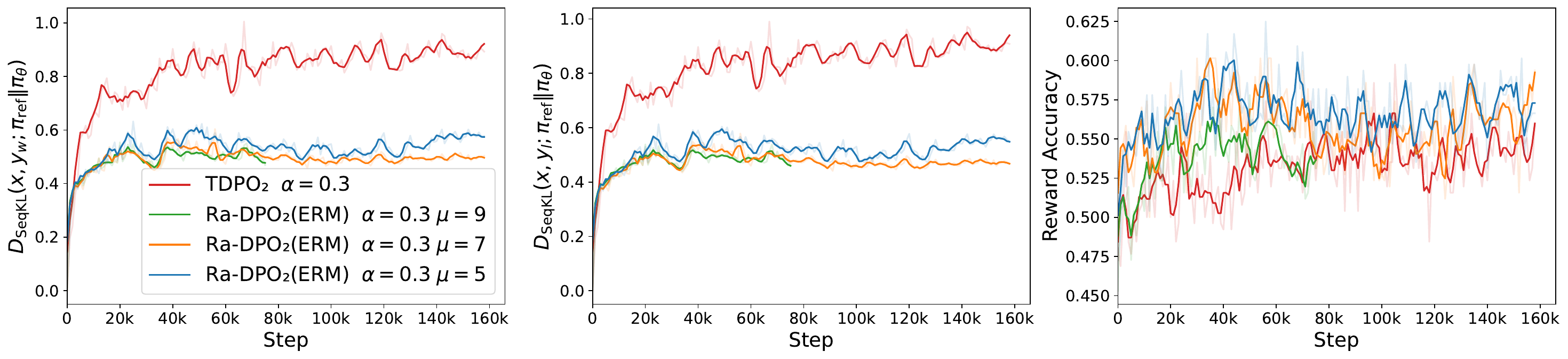}
    }
    \quad
    \subfigure{
        \includegraphics[width=0.99 \linewidth]{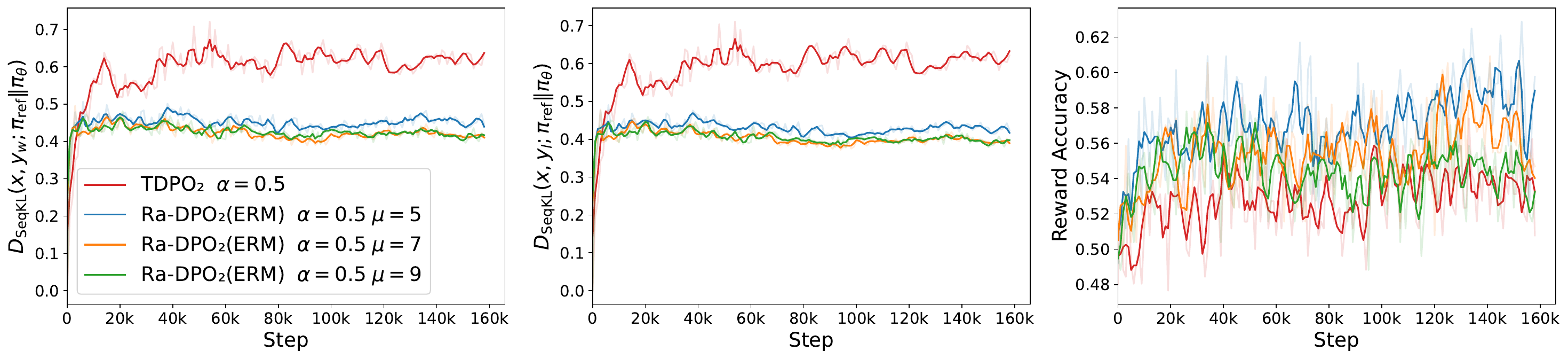}
    }
    \quad
    \subfigure{
        \includegraphics[width=0.99 \linewidth]{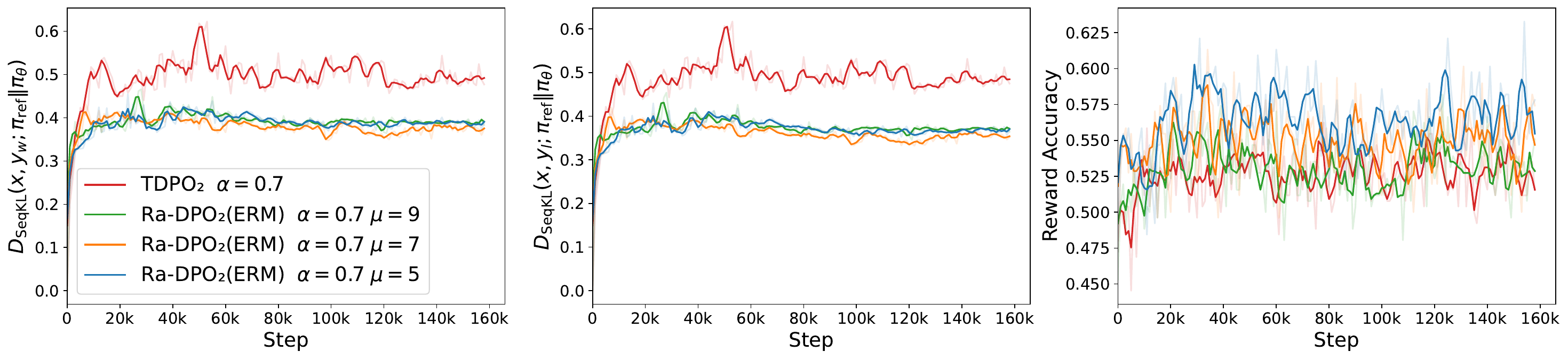}
    }
    \quad
    \subfigure{
        \includegraphics[width=0.99 \linewidth]{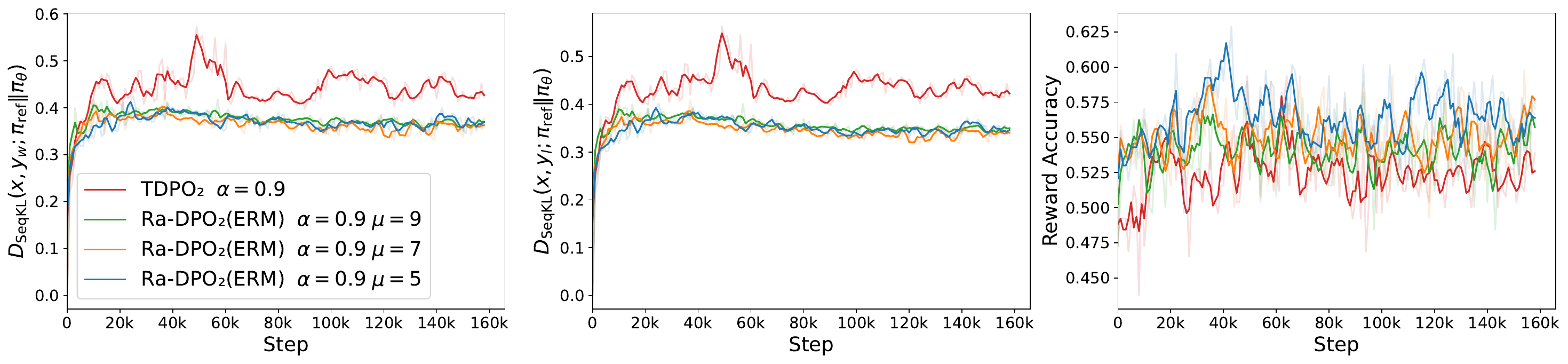}
    }
    \caption{
        The experiment on the Anthropic HH dataset with Pythia-1.4B serving as the base model.
        \textbf{Left} and \textbf{Middle} present the progression of sequential KL divergence (the lower the better) for both preferred and dispreferred responses.
        \textbf{Right} illustrates reward accuracy curves (the higher the better).
    }
    \label{Appendix: Fig.3}
    \vskip 0.1in
\end{figure}

\begin{figure}[h!]
    \vskip 0.1in
    \centering
    \subfigure{
        \includegraphics[width=0.315 \linewidth]{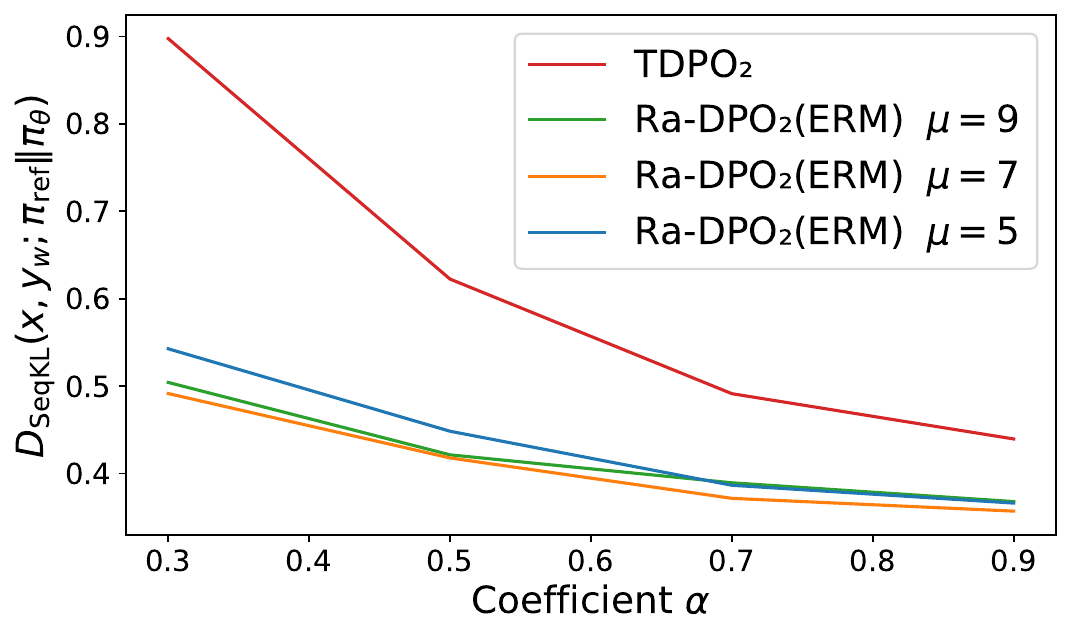}
    }
    \subfigure{
        \includegraphics[width=0.315 \linewidth]{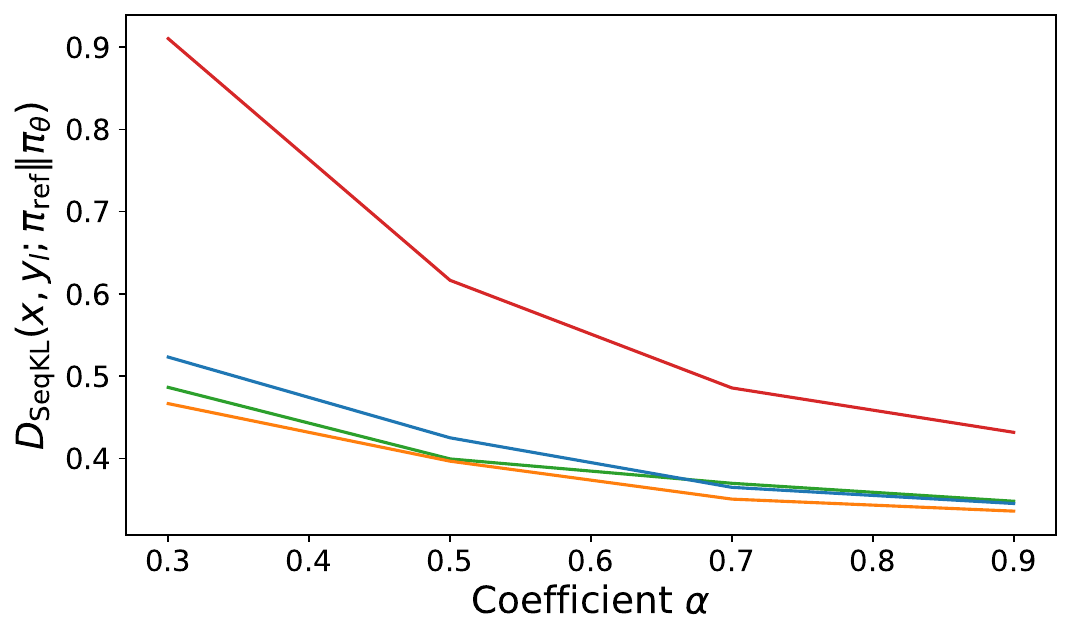}
    }
    \subfigure{
        \includegraphics[width=0.315 \linewidth]{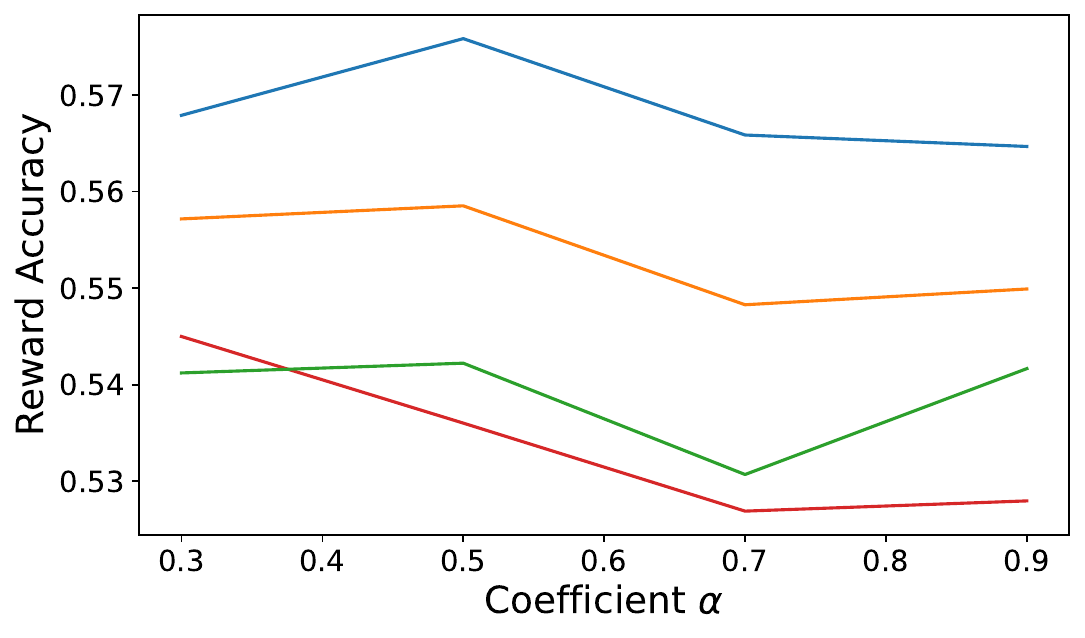}
    }
    \caption{
        The experiment on the Anthropic HH dataset with Pythia-1.4B serving as the base model.
        \textbf{Left} and \textbf{Middle} presents the sequential KL divergence (the lower the better) for preferred and dispreferred responses, while \textbf{Right} presents the reward accuracy curves (the higher the better) under $\alpha = \{0.3, 0.5, 0.7, 0.9\}$.
     }
    \label{Appendix: Fig.4}
    \vskip 0.1in
\end{figure}

\begin{figure}[htb]
    \vskip 0.1in
    \centering
    \subfigure{
        \includegraphics[width=0.99 \linewidth]{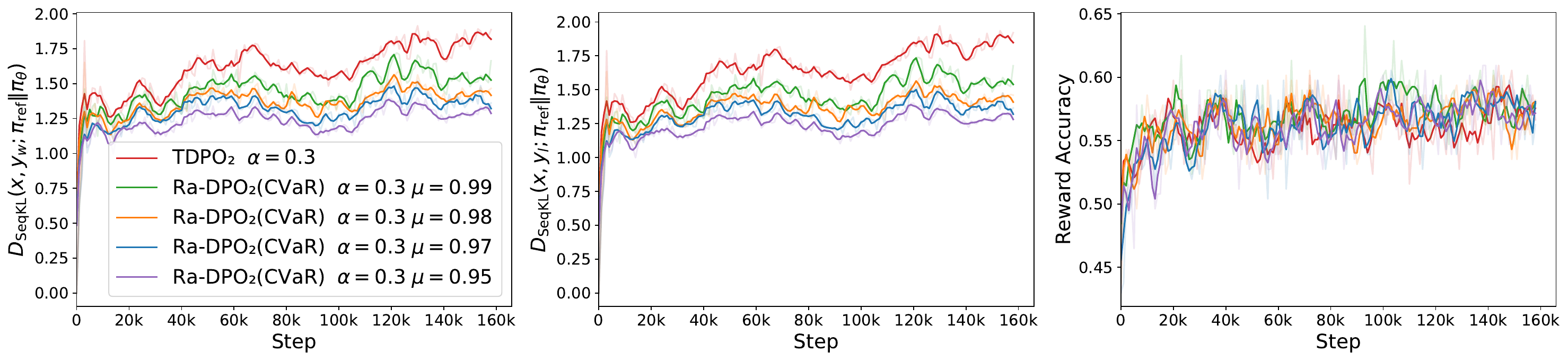}
    }
    \quad
    \subfigure{
        \includegraphics[width=0.99 \linewidth]{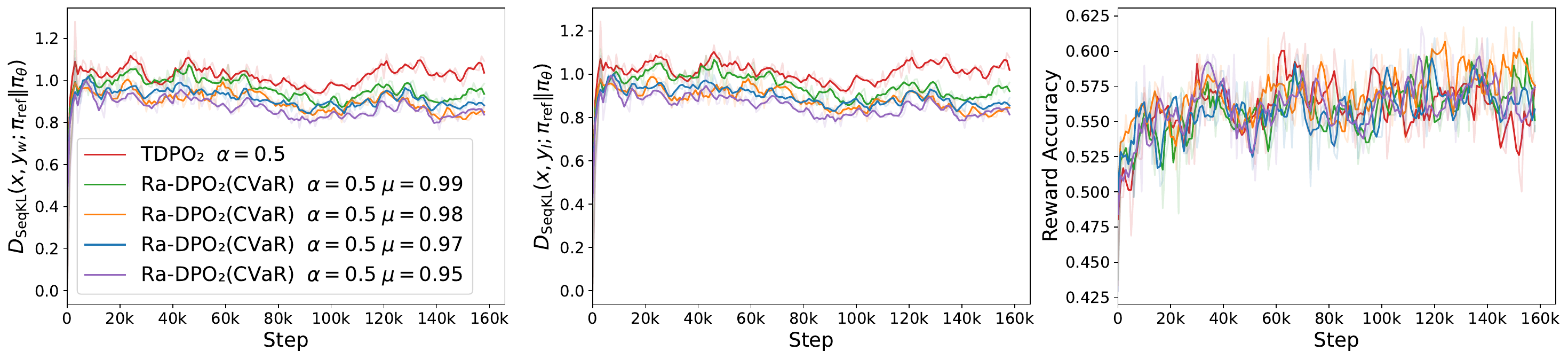}
    }
    \quad
    \subfigure{
        \includegraphics[width=0.99 \linewidth]{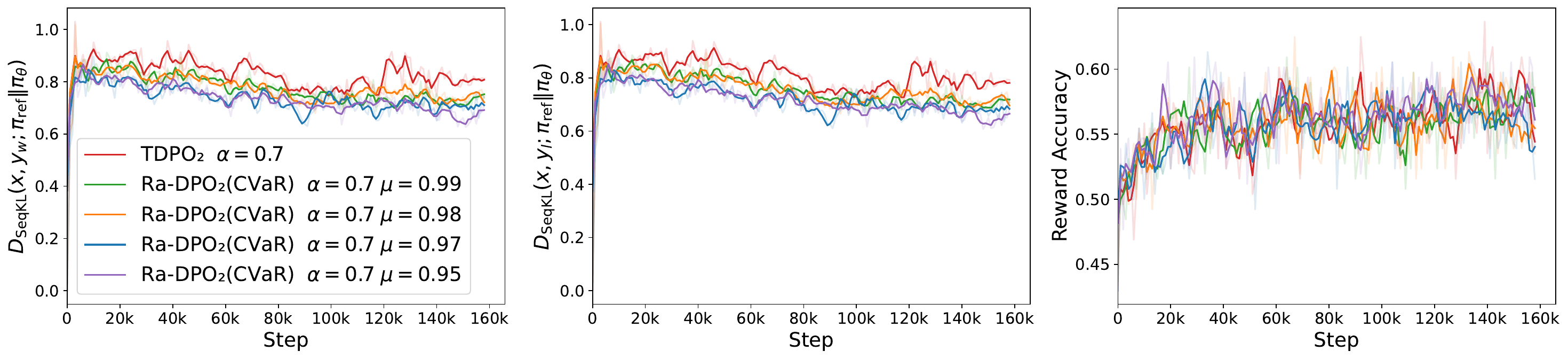}
    }
    \quad
    \subfigure{
        \includegraphics[width=0.99 \linewidth]{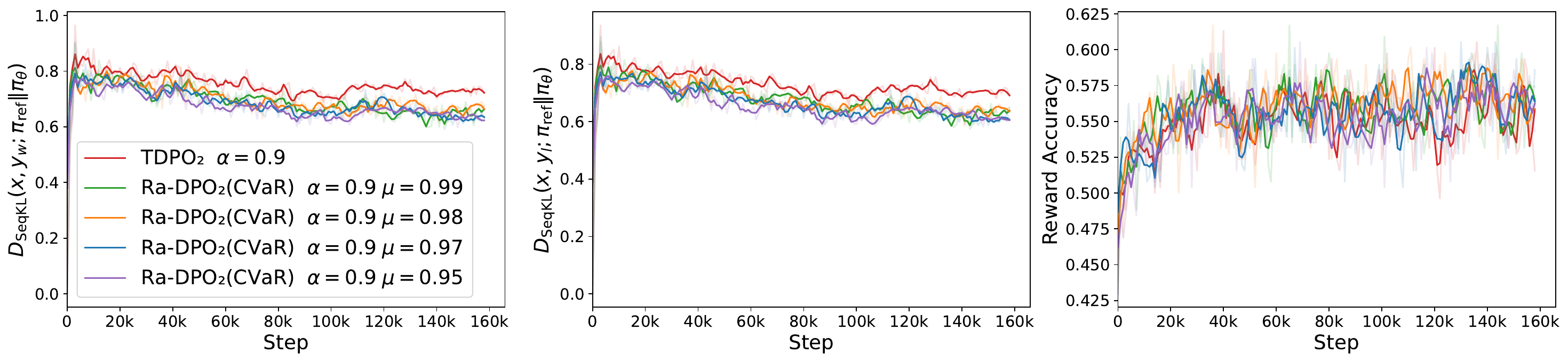}
    }
    \caption{
        The experiment on the Anthropic HH dataset with Pythia-2.8B serving as the base model.
        \textbf{Left} and \textbf{Middle} present the progression of sequential KL divergence (the lower the better) for both preferred and dispreferred responses.
        \textbf{Right} illustrates reward accuracy curves (the higher the better).
    }
    \label{Appendix: Fig.5}
    \vskip 0.1in
\end{figure}

\begin{figure}[h!]
    \vskip 0.1in
    \centering
    \subfigure{
        \includegraphics[width=0.315 \linewidth]{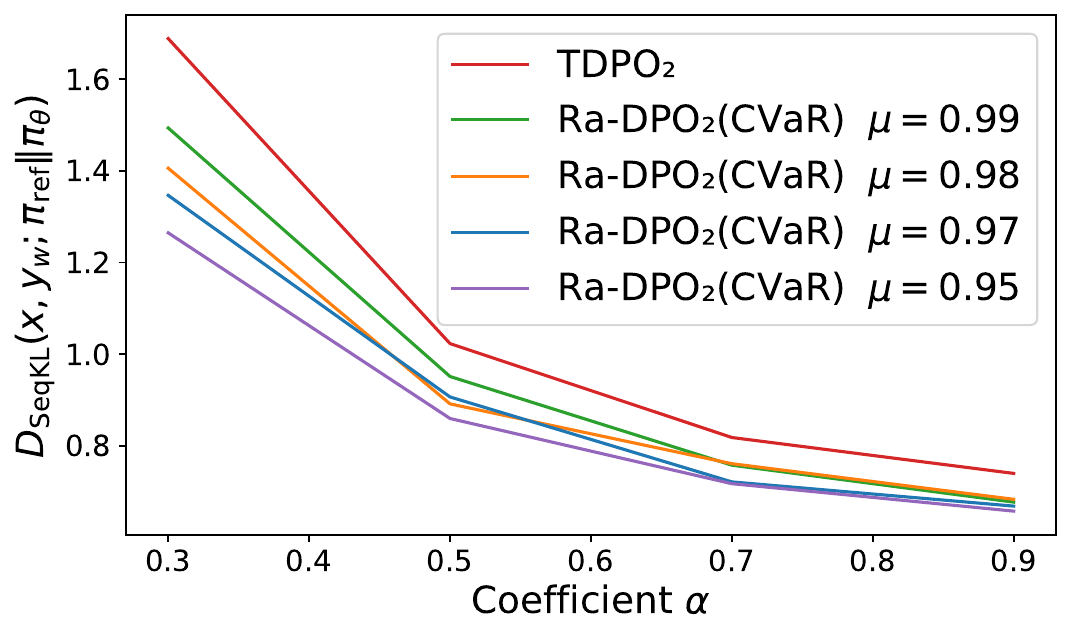}
    }
    \subfigure{
        \includegraphics[width=0.315 \linewidth]{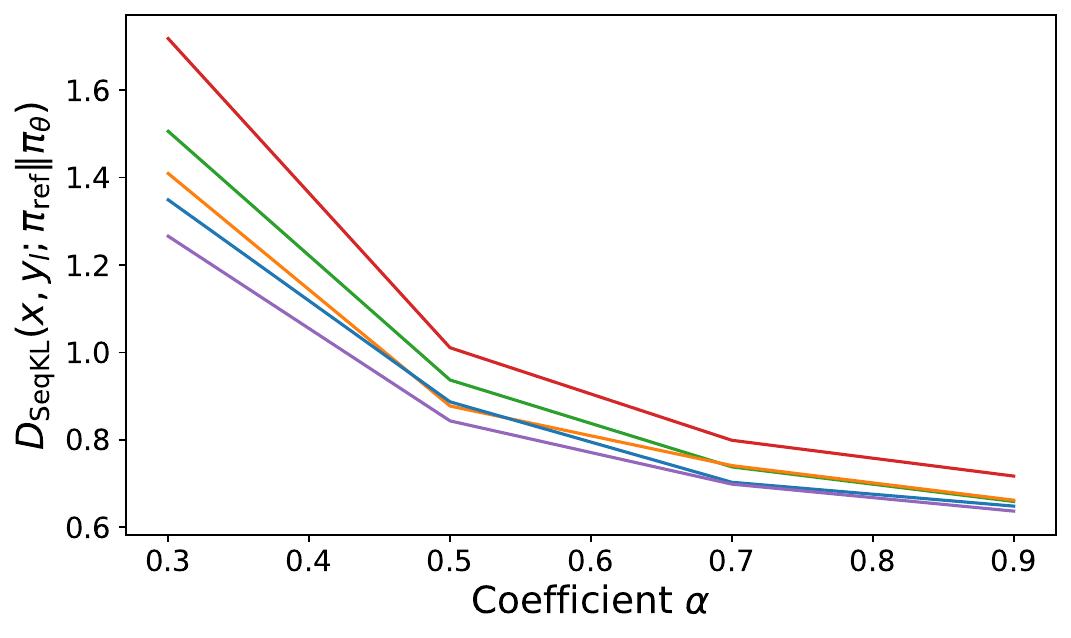}
    }
    \subfigure{
        \includegraphics[width=0.315 \linewidth]{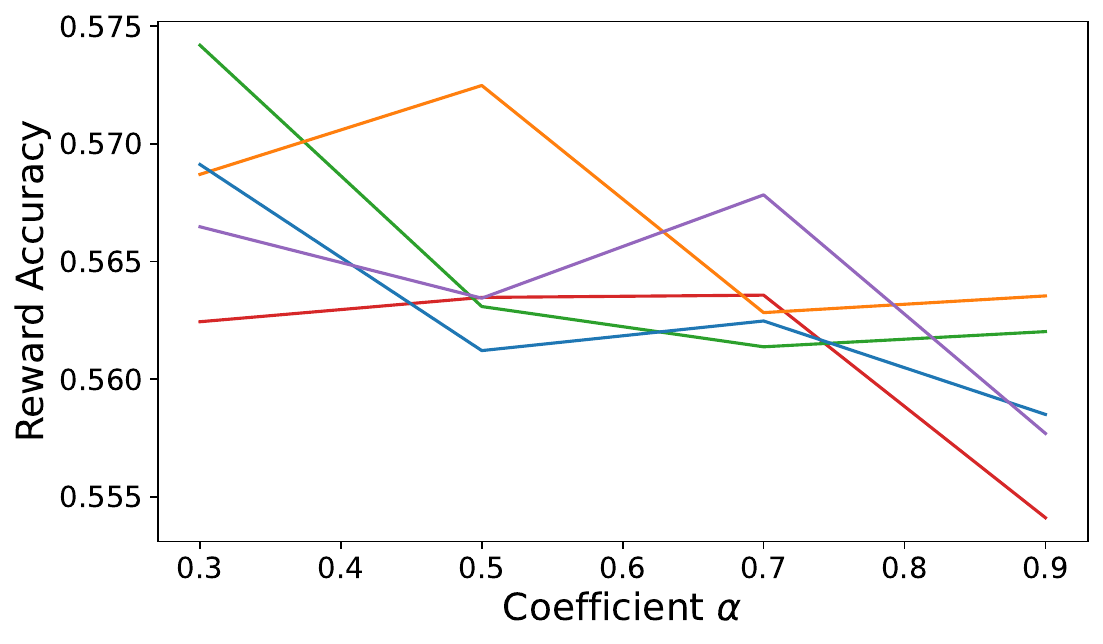}
    }
    \caption{
        The experiment on the Anthropic HH dataset with Pythia-2.8B serving as the base model.
        \textbf{Left} and \textbf{Middle} presents the sequential KL divergence (the lower the better) for preferred and dispreferred responses, while \textbf{Right} presents the reward accuracy curves (the higher the better) under $\alpha = \{0.3, 0.5, 0.7, 0.9\}$.
    }
    \label{Appendix: Fig.6}
    \vskip 0.1in
\end{figure}

\begin{figure}[h!]
    \vskip 0.1in
    \centering
    \subfigure{
        \includegraphics[width=0.99 \linewidth]{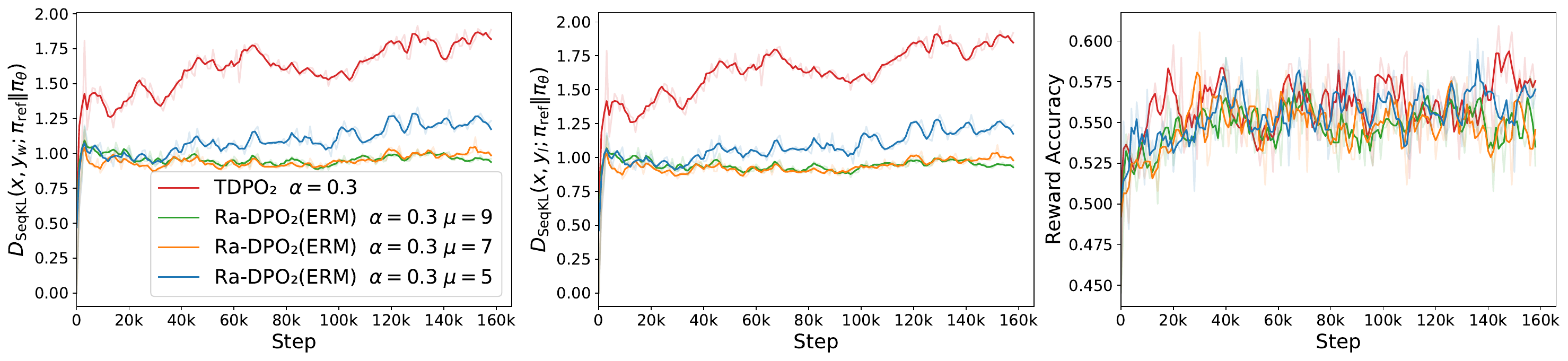}
    }
    \quad
    \subfigure{
        \includegraphics[width=0.99 \linewidth]{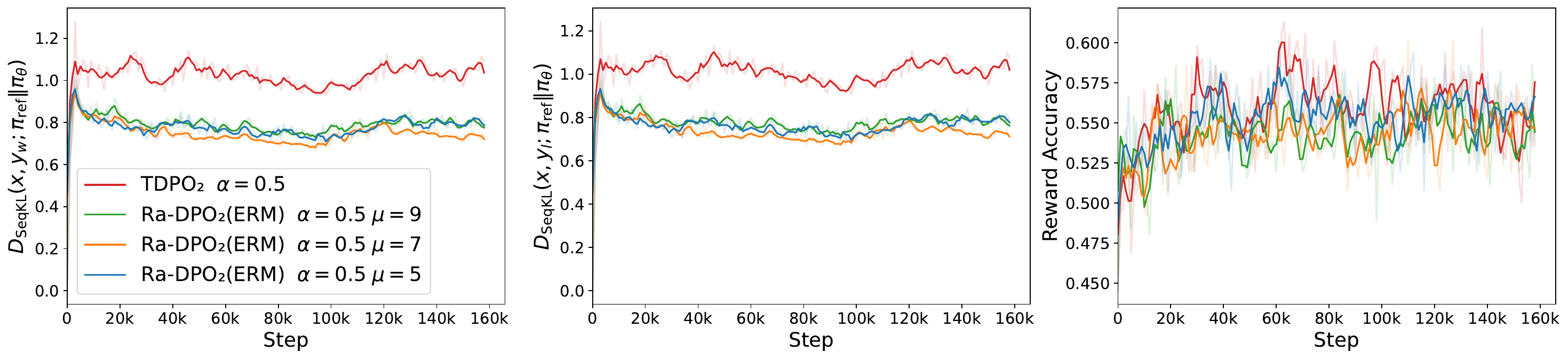}
    }
    \quad
    \subfigure{
        \includegraphics[width=0.99 \linewidth]{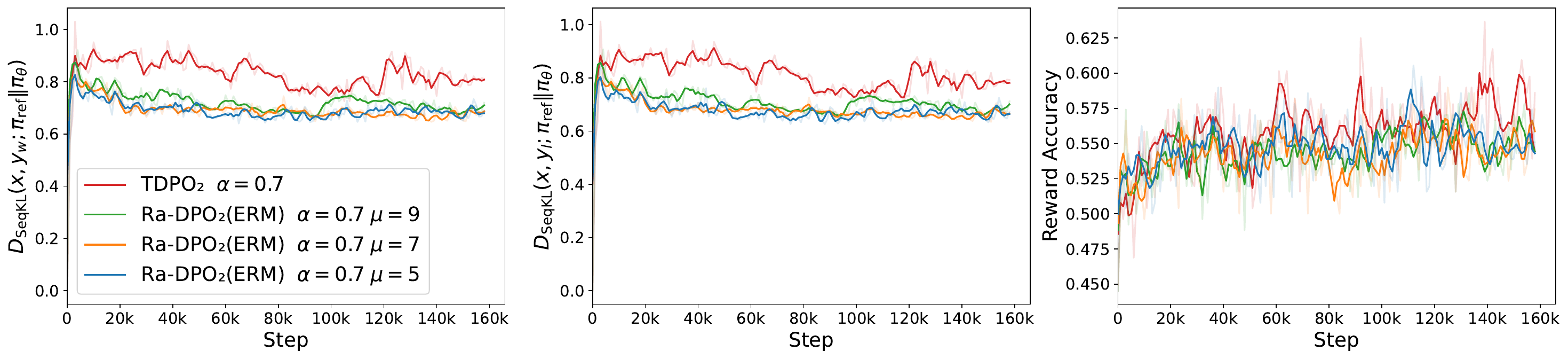}
    }
    \quad
    \subfigure{
        \includegraphics[width=0.99 \linewidth]{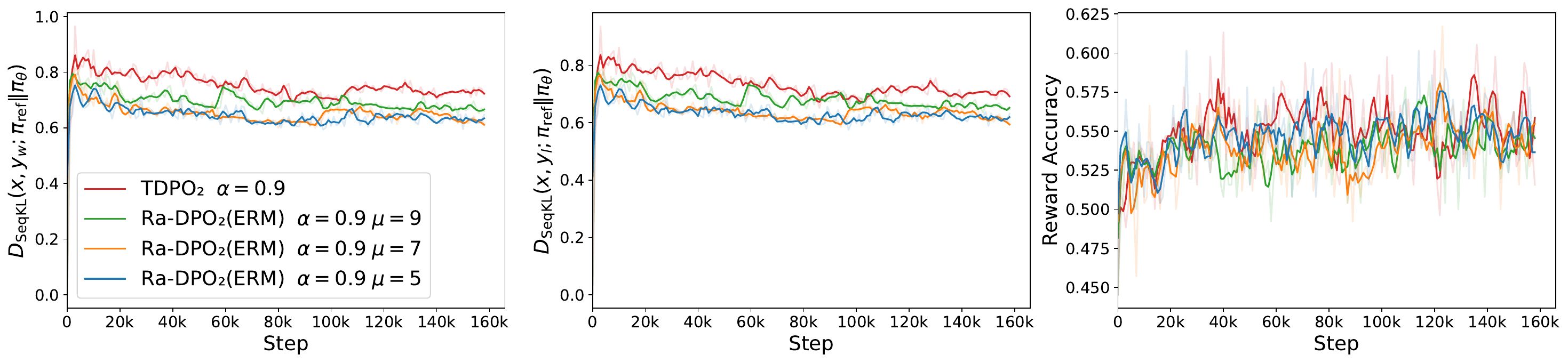}
    }
    \caption{
        The experiment on the Anthropic HH dataset with Pythia-2.8B serving as the base model.
        \textbf{Left} and \textbf{Middle} present the progression of sequential KL divergence (the lower the better) for both preferred and dispreferred responses.
        \textbf{Right} illustrates reward accuracy curves (the higher the better).
    }
    \label{Appendix: Fig.7}
    \vskip 0.1in
\end{figure}

\begin{figure}[h!]
    \vskip 0.1in
    \centering
    \subfigure{
        \includegraphics[width=0.315 \linewidth]{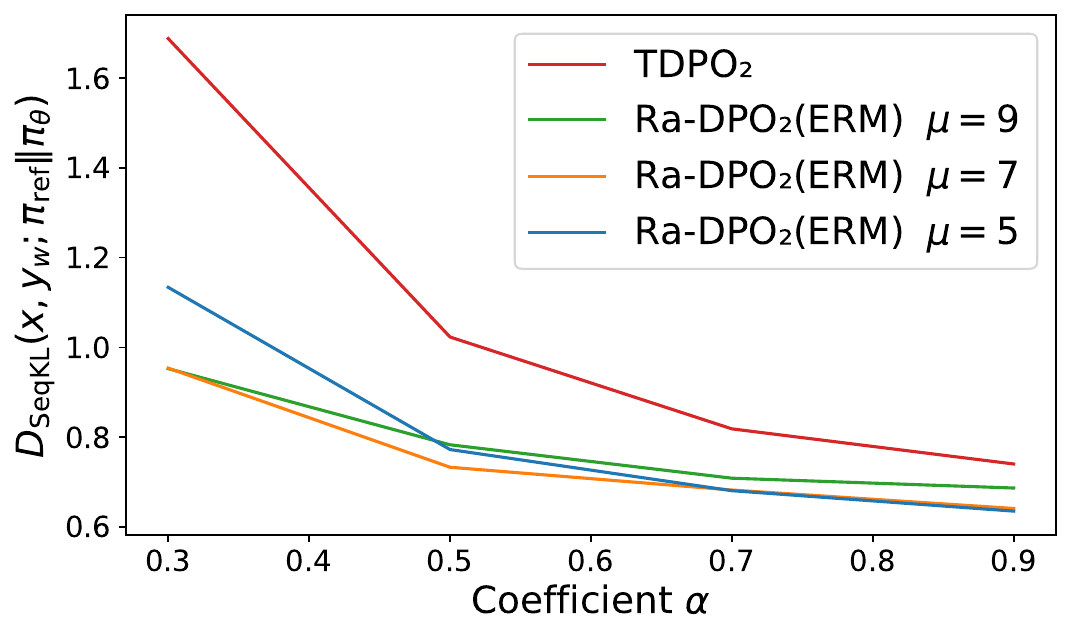}
    }
    \subfigure{
        \includegraphics[width=0.315 \linewidth]{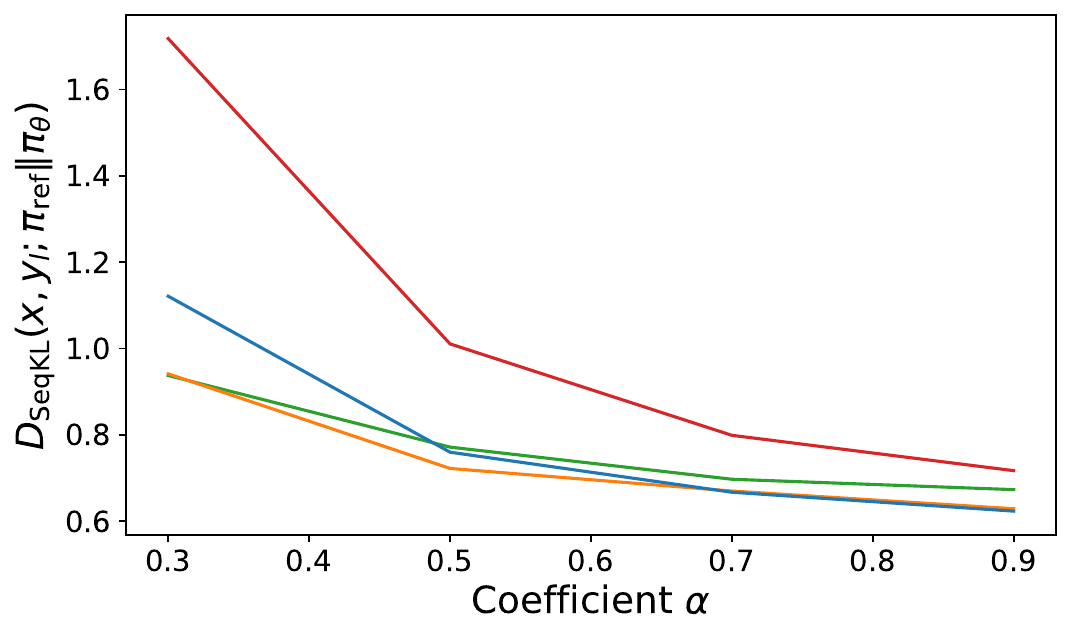}
    }
    \subfigure{
        \includegraphics[width=0.315 \linewidth]{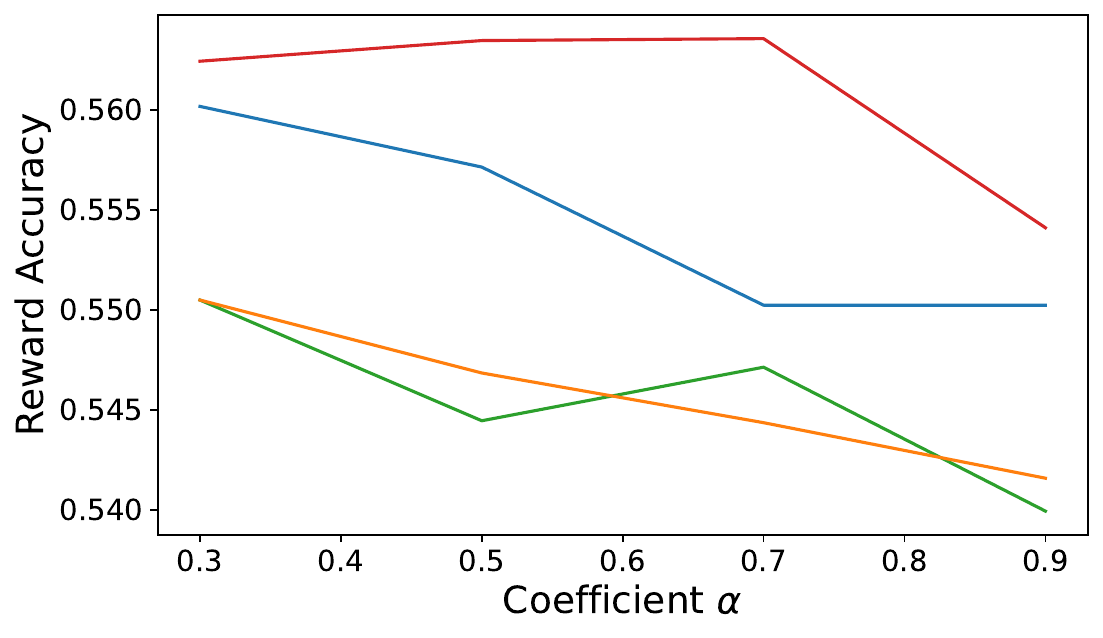}
    }
    \caption{
        The experiment on the Anthropic HH dataset with Pythia-2.8B serving as the base model.
        \textbf{Left} and \textbf{Middle} presents the sequential KL divergence (the lower the better) for preferred and dispreferred responses, while \textbf{Right} presents the reward accuracy curves (the higher the better) under $\alpha = \{0.3, 0.5, 0.7, 0.9\}$.
       }
    \label{Appendix: Fig.8}
    \vskip 0.1in
\end{figure}

\begin{figure}[h!]
    \vskip 0.1in
    \centering

    \subfigure{
        \includegraphics[width=0.99 \linewidth]{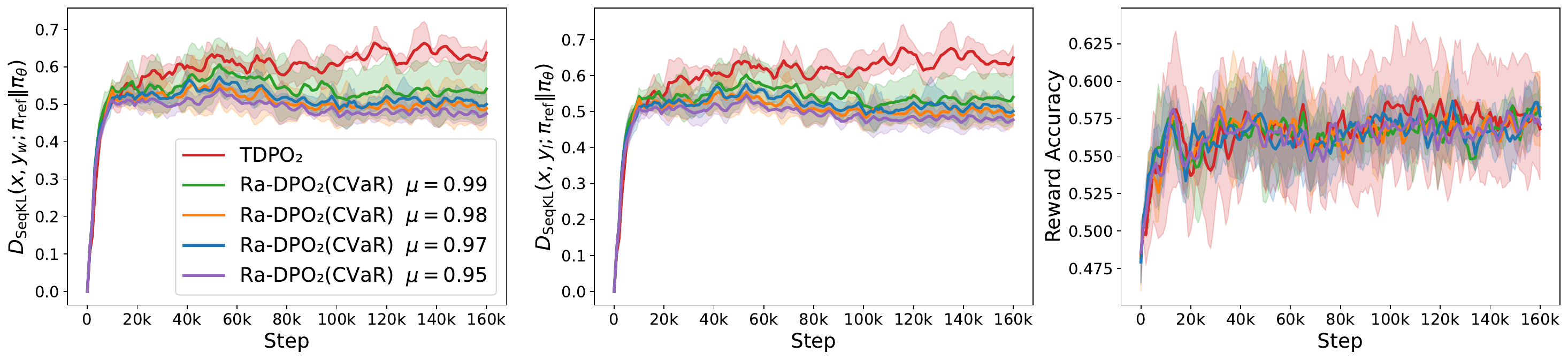}
    }
    \quad
    \subfigure{
        \includegraphics[width=0.99 \linewidth]{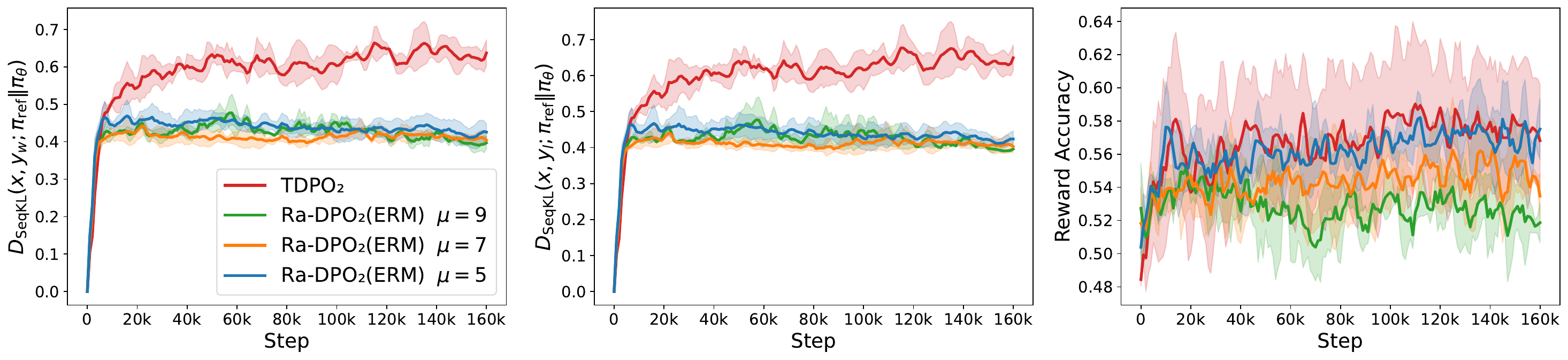}
    }
    \caption{The experiment on the Anthropic HH dataset with Pythia-1.4B serving as the base model.
    % We implemented $\text{TDPO}_\text{2}$, and different versions of $\text{Ra-DPO}_\text{2}$ with respect to the parameters $\alpha$ and $\mu$.
    \textbf{Left} and \textbf{Middle} present the progression of sequential KL divergence (the lower the better) for both preferred and dispreferred responses.
    \textbf{Right} illustrates reward accuracy curves (the higher the better).
    For all algorithms, we report the average performance (solid line) across three random seeds, with the shaded region representing one standard deviation around the mean.
    % The results presented are averaged over three random seeds.
    }
    \label{Appendix: Fig.9}
\end{figure}

\begin{figure}[htb]
    \vskip 0.2in
    \centering
    \includegraphics[width=0.5 \linewidth]{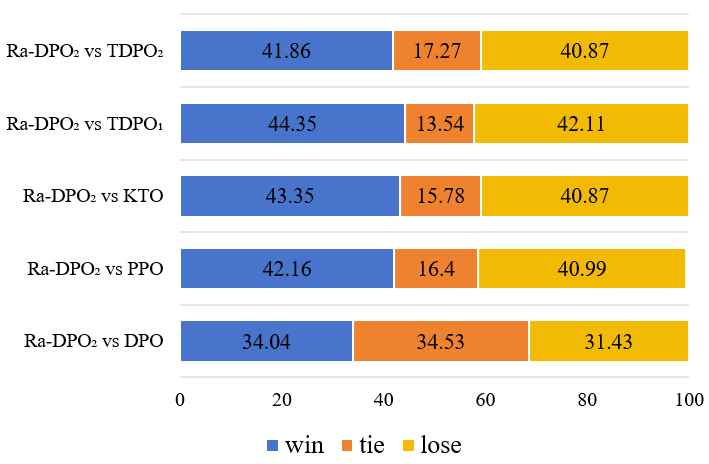}
    \caption{AlpacaEval comparison between DPO, PPO, $\text{TDPO}_\text{1}$, $\text{TDPO}_\text{2}$, and $\text{Ra-DPO}_\text{2}$ methods. The win, tie, and lose rates are evaluated based on \emph{oasst-pythia-12b}.
    }
    \label{Appendix: Fig.10}
\end{figure}

Here, we provide some additional experimental results, which are illustrated in Figures \ref{Appendix: Fig.1}-\ref{Appendix: Fig.10}.

\begin{itemize}
    \item Figures \ref{Appendix: Fig.1}-\ref{Appendix: Fig.4} illustrate the experiment on the Anthropic HH dataset with Pythia-1.4B serving as the base model.
    We implemented $\text{TDPO}_\text{2}$, and different versions of $\text{Ra-DPO}_\text{2}$ with respect to the parameters $\alpha$ and $\mu$.

    \item Figures \ref{Appendix: Fig.5}-\ref{Appendix: Fig.8} show corresponding results using Pythia-2.8B as the base model.
    The same set of algorithms was evaluated under varying $\alpha$ and $\mu$ configurations.

    \item Figure \ref{Appendix: Fig.9} illustrates the experiment on the Anthropic HH dataset with Pythia-1.4B serving as the base model.
    Let $\alpha = 0.5$, we implemented $\text{TDPO}_\text{2}$, and different versions of $\text{Ra-DPO}_\text{2}$ with respect to the risk control parameter $\mu$.
    In the figure, for all algorithms, we report the average performance (solid line) across three random seeds, with the shaded region representing one standard deviation around the mean.
    We aim to highlight the statistically significant improvements achieved by the proposed method, although training large-scale models entails substantial computational costs in terms of time and resources. 
    The figure illustrates that, under both the CVaR-based nested risk measure and the ERM-based risk measure with $\mu = 5$,  the proposed algorithm achieves reward accuracy comparable to that of the baseline method but with greater stability, while maintaining a consistently low sequential KL divergence.

    \item  Figure \ref{Appendix: Fig.10} illustrates the comparison between DPO, PPO, $\text{TDPO}_\text{1}$, $\text{TDPO}_\text{2}$, and $\text{Ra-DPO}_\text{2}$ methods through AlpacaEval. 
    It presents a straightforward result: Compared to the baseline algorithms, $\text{Ra-DPO}_\text{2}$ achieves a high winrate, demonstrating superior performance in assisting LLMs to generate high-quality responses.
\end{itemize}

\section{The Discussion of Limitations and Impacts}   \label{Appendix: The discussion of limitations and impacts}
\subsection{Limitations} \label{Appendix: Limitations}
This paper focuses on the risks associated with token-level generation when aligning LLMs with human values and intentions.
Our main contributions include theoretical analysis (see Section~\ref{Section: Methodology} and Appendix~\ref{Appendix: Methodology}), practical algorithm (see Appendix~\ref{Appendix: Algorithm}) and simulation verification (see Section~\ref{Section: Experiments} and Appendix~\ref{Appendix: Experiments}). 
The main results in the paper characterize the proposed method's performance in terms of reward accuracy and sequential KL divergence. 
Below, we discuss the limitations of the proposed method from both theoretical and experimental viewpoints.

\textbf{Theoretical Viewpoint: } 
Our theoretical results are based on a class of risk-sensitive functionals that satisfy concavity, monotonicity and translation invariance for any random variables $Z, Z'\in \mathcal{Z}$.
Concavity implies risk aversion; translation invariance is an important condition for the validity of Lemma~\ref{Lemma: Equivalent MDP}.
Our conclusions may not be valid when such assumptions do not hold. 
Fortunately, this class captures a broad range of useful objectives, including the popular CVaR \cite{1997_artzner_CVaR} and ERM \cite{2023_hau_ERM}.

\textbf{Experimental Viewpoint: } 
Our experiments show that Ra-DPO (our method) achieves higher reward accuracy while maintaining minimal model drift, as indicated by lower sequential KL divergence.
One limitation is that Ra-DPO may not be fully effective for tasks such as harmful content moderation and toxicity detection. 
This is because our primary goal is to reduce the risk of impaired decision-making and reasoning capabilities due to model deviation from the reference model during LLM alignment.
However, it is worth noting that the problem we are addressing is both widespread and of significant importance.
Furthermore, we recommend a safe or low-risk approach that incorporates risk-awareness within the Safe RLHF \cite{2024_dai_safe-RLHF} or SACPO \cite{2024_wachi_sacpo} framework.
These approaches explicitly or implicitly model both cost and reward functions while accounting for cost distributions. 
It may require more computational resources due to the need to train additional models. 
However, our method can serve as a solid foundation for such potential approaches. 
We intend to explore this direction further in future research.

\subsection{Impact Statement} \label{Appendix: Impact Statement}
This paper presents work aimed at making LLMs more helpful.
Specifically, we focus on how to reduce the risk of impaired decision-making and reasoning capabilities due to model deviation from the reference model during LLM alignment.
Our work has many positive societal impacts, such as providing a theoretical foundation for risk-aware language generation task, none of which we feel must be specifically highlighted.
There are no negative societal impacts on our work.

%%%%%%%%%%%%%%%%%%%%%%%%%%%%%%%%%%%%%%%%%%%%%%%%%%%%%%%%%%%%%%%%%%%%%%%%%%%%%%%
%%%%%%%%%%%%%%%%%%%%%%%%%%%%%%%%%%%%%%%%%%%%%%%%%%%%%%%%%%%%%%%%%%%%%%%%%%%%%%%

\end{document}